\documentclass[10pt,technote,peerreviewca,onecolumn]{IEEEtran} 

\usepackage{graphicx}
\usepackage{epstopdf}
\usepackage{float}
\usepackage{subfig}
\usepackage{multirow}

\usepackage{cite}
\usepackage{amsmath, amssymb, mathbbol}
\usepackage{mathtools}
 
\renewcommand{\QED}{\QEDopen}

\begin{document}
\title{Understanding Encoder-Decoder Structures in Machine Learning Using Information Measures}
\author{Jorge F. Silva, Victor Faraggi, Camilo Ramirez, Alvaro Ega\~{n}a y Eduardo Pavez
\thanks{J. F. Silva, V. Faraggi and C. Ramirez  are with the Information and Decision System Group (IDS), Department of Electrical Engineering,  University of Chile, Av. Tupper 2007 Santiago, Room 508, Chile, 
 (emails: josilva@ing.uchile.cl, victor.faraggi@ug.uchile.cl, camilo.ramirez@ug.uchile.cl).}
\thanks{A. Ega\~{n}a  is with the Advanced Laboratory for Geostatistical Supercomputing (ALGES),  University of Chile (email: aegana@alges.cl).}
\thanks{E. Pavez  is with the Department of Electrical and Computer Engineering,  University of Southern California (email: pavezcar@usc.edu).}
}

\maketitle

\begin{abstract}
We present new results 
to model and understand the role of encoder-decoder design in machine learning (ML) from an information-theoretic angle. We use two main information concepts, information sufficiency (IS) and mutual information loss (MIL), to represent predictive structures in machine learning. Our first main result provides a functional expression that characterizes the class of probabilistic models consistent with an IS encoder-decoder latent predictive structure. This result formally justifies the encoder-decoder forward stages many modern ML architectures adopt to learn latent (compressed) representations for classification.  To illustrate IS as a realistic and relevant model assumption, we revisit some known ML concepts and present some interesting new examples: invariant, robust, sparse, and digital models. Furthermore, our IS characterization allows us to tackle the fundamental question of how much performance (predictive expressiveness) could be lost, using the cross entropy risk, when a given encoder-decoder architecture is adopted in a learning setting. Here, our second main result shows that a mutual information loss quantifies the lack of expressiveness attributed to the choice of a (biased) encoder-decoder ML design. Finally, we address the problem of universal cross-entropy learning with an encoder-decoder design where necessary and sufficiency conditions are established to meet this requirement. In all these results, Shannon's information measures offer new interpretations and explanations for representation learning.
\end{abstract}


 \newtheorem{corollary}{\bf COROLLARY}
\newtheorem{theorem}{\bf THEOREM}
\newtheorem{lemma}{\bf LEMMA}
\newtheorem{proposition}{\bf PROPOSITION}
\newtheorem{definition}{\bf Definition}
\newtheorem{remark}{\bf Remark}

\section{Introduction}
\label{sec_intro}
In many machine learning (ML) tasks, the observation (or input) $X$ lives in a continuous (multivariate) high dimensional space while the class (target) variable $Y$ is discrete. Given this discrepancy at the space level, it is common to assume that there are many latent factors (random innovation components) that produce $X$ but do not affect $Y$ and, consequently, this redundant information is not needed for learning a good classifier \cite{bengio_2013}. Therefore, representation learning (RL) addresses the compression task of finding a lossy transformation of an observation $X$ (or encoder) that is highly informative and ideally sufficient for predicting $Y$. A large body of work addresses the design of lossy representations (compressors) from data. Many of these methods rely on the use of information-theoretic measures to quantify the predictive relationship between $X$ and $Y$ \cite{achille_2018,amjad_2019,alemi_2017,kingma_2014,tishby_1999,strouse_2017,tegmark_2020}. This general idea of redundancy on $X$ to explain $Y$ translates intuitively into a notion of probabilistic structure \cite{Bloem_2019,xu_2012,silva_mlsp_2007,silva_2012b} that is one of the key justifications for the adoption of encoder-decoder strategies in ML.

The formal characterization of probabilistic structures and the study of the repercussions of these model assumptions 
in the design of algorithms and architectures is an essential area of theoretical research in ML \cite{xu_2012,Bloem_2019,Dubois_2021,silva_2022_aistat,devroye1996,silva_2012b}. This theoretical understanding has been used to explain some design choices of neural network architectures \cite{Bloem_2019,zaheer_2017} and has been adopted to design data compressors for prediction \cite{Dubois_2021}. 

On formalizing probabilistic structures in learning and connecting it with the idea of sufficient representations for inference (classification), we highlight the  paper by Bloem-Reddy and Teh \cite{Bloem_2019}. This seminal work studies probabilistic models (where a model is a joint distribution $\mu_{X,Y}$ between $X$ and $Y$) that are invariant to the action of a compact group of measurable transformations $\mathcal{G}$.  
Classification tasks invariant to operations such as permutation, rotation, translations, and scale are commonly considered for the design of ML algorithms in image and computer vision problems \cite{bengio_2013, zaheer_2017}. 
We highlight two important results in \cite{Bloem_2019}. First, they show that under the assumption of invariance of the joint model $\mu_{X,Y}$ to the action of any operation $g(\cdot)$ on $X$  within a class $\mathcal{G}$,  
there is a lossy function $\eta_{\mathcal{G}}(\cdot)$ (determined by $\mathcal{G}$) acting on $X$ that makes $X$ and $Y$ conditionally independent. To the best of our knowledge, this is the first result that makes a connection between the probabilistic structure of a model $\mu_{X,Y}$ (invariance to transformations)
and the existence of a lossy encoder (compressor) that is sufficient in the strong sense of statistical independence \cite{dawid_1979}. Another important result  \cite[Theorem 7]{Bloem_2019} states that for the class of invariant models $\mu_{X,Y}$ w.r.t. a group $\mathcal{G}$, the joint model $\mu_{X,Y}$ has a functional characterization of the form $Y= f(W,\eta_{\mathcal{G}}(X))$, where we recognize the role of an encoder $\eta_{\mathcal{G}}(\cdot)$ 
and a function $f(\cdot, \cdot)$ driven by a random noise $W$  that is independent of $X$.

\subsection{Contributions}
The contribution of this work is to extend the theory of representation learning (RL) introduced by Bloem-Reddy and Teh \cite{Bloem_2019}. Our angle is to study the encoder's role (the compressor element in RL) from an information-theoretic perspective. We have a series of new results organized in three domains: characterization of probabilistic structure in ML, mismatch encoder-decoder analysis in ML, and universal cross-entropy learning with an encoder-decoder design.

On the first part of this work, we extend the theory in \cite{Bloem_2019} about characterizing probabilistic structure in ML by exclusively looking at the predictive component of $\mu_{X,Y}$ and not imposing any condition on the marginal distribution of the input $X$. This novel direction is supported by the observation that in classification, i.e., our main operational problem, the element that is sufficient for optimal decision is the predictive part of $\mu_{X,Y}$ (i.e., $\mu_{Y|X}$ ) \cite{devroye1996,duda1983,michie_1994}. With that in mind, we propose the adoption of information sufficiency (IS) 
to model the latent structure of a model $\mu_{X,Y}$ in the predictive direction $X \rightarrow \eta(X)  \rightarrow Y$ using for that the {\em Shannon mutual information} (MI) \cite{shannon_1948}. 
Our first main result (Theorem \ref{th_representation})  
determines the precise condition on a model to meet the following functional (encoder-decoder) predictive expression $Y= f(W,\eta(X))$. More precisely, Theorem \ref{th_representation} shows that this functional expression is met if, and only if, a model belongs to a specific IS-structured class. As expected, this new IS latent structure only imposes a condition on the predictive component of $\mu_{X,Y}$, and it provides the flexibility to expand the probabilistic analysis introduced in \cite{Bloem_2019} significantly. Indeed, this IS structure is shown to be instrumental in extending the functional expression in \cite[Theorem  7]{Bloem_2019} for a larger class of ML models (see Ths.\ref{lm_invariant_IS} and \ref{lemma_IS_robust}, and related Corollaries).          
%
On the relevance of our IS characterization for ML,  Theorem \ref{th_representation} shows that IS expresses a model predictive structure by the distinctive role played by an encoder (a lossy compressor) and a decoder (a soft mapping). This result explains and justifies the adoption of some specific ML architectures with an encoder-decoder structure and offers a formal justification of the encoder-decoder inference stages adopted by many ML algorithms  \cite{bengio_2013,kingma_2014,Dubois_2021,achille_2018,alemi_2017}. In particular, we determine some encoder-decoder designs entirely consistent with probabilistic models with a given IS-embedded structure (see Ths. \ref{lm_invariant_IS} and \ref{lemma_IS_robust}, and related Corollaries). 

To give further significance to the proposed IS structure in ML, we ask how much performance could be lost (using the cross entropy risk) when a given encoder-decoder architecture is adopted  in a learning (model selection) task. 
Our second main result in Theorem \ref{th_main_IS_mismatch} shows that if the encoder is not IS for the task, the performance degradation (or lack of expressiveness) attributed to this biased encoder-decoder design is determined by a mutual information loss (MIL).  More specifically, Theorem \ref{th_main_IS_mismatch} offers a probabilistic angle to evaluate the distinctive role played by the encoder and decoder in the design.  On the encoder side, which is the central element in representation learning, we show in Theorem \ref{th_if_projection_aanlogy} that the lack of expressiveness induced by an encoder $\eta(\cdot)$ --- that is not IS --- is equivalent to the approximation error induced by projecting the true model over a specific class of IS models characterized in Theorem \ref{th_representation}. 
Relevantly, this  result (Theorem \ref{th_if_projection_aanlogy}) provides an information projection (IP) interpretation to explain the effect of encoder-decoder design in ML. Finally, we apply this IP result as a tool to evaluate the expressive effect of a multilayer (deep) architecture.  On this, we characterize in Theorem \ref{pro_multilayer_info_loss} the individual information loss induced by each layer (IP error) as well as its respective and implicit IS model assumption from the IS characterization presented  in Theorem \ref{th_representation}.

To conclude our IS driven study of ML, we cover the realistic learning setting where both the encoder and the decoder are data-driven elements of design. In this scenario, we show that achieving the optimal performance (i.e., the minimum cross-entropy risk for a task) implies specific conditions for the encoder and the decoder of an ML scheme.   
Our third main result in Theorem \ref{th_consistency} presents necessary and sufficient conditions to meet this learning requirement, i.e., strong consistency for cross-entropy learning. Fundamentally, this implies learning the true predictive model in the strong KL sense and, as a needed requirement, learning an asymptotically IS data-driven representation.
Finally, we show the feasibility of learning IS representation via digitalization in Lemma \ref{lm:expresive_VQ} (confirming the expressive capacity of digital encoders) and the adequacy of the information bottleneck (IB) optimization principle to achieve our IS expressive condition in Lemma \ref{lm:IB_expressive}.

Finally, we design a controlled experimental setting and use our results to explain the learning capability of an encoder-decoder ML scheme that uses a multilayer perceptron architecture (MLP). From two of our main results (Theorems \ref{th_main_IS_mismatch} and \ref{th_consistency}), we present evidence that the well-known functional approximation capability of multilayer NN \cite{hornik_1989,zhou_2020} has the potential to achieve nearly optimal learning performance in the cross entropy sense. We also present evidence supporting that prior IS structural knowledge in the form of a pre-encoder (projector) provides a systematic performance gain, particularly relevant in low-data regimes. 

\subsection{Organization}
The rest of the paper is organized as follows: Section \ref{sec_preliminaries} presents definitions and basic notation. Section \ref{sec_probabilistic_structure} introduces IS to model predictive structure in ML (Theorem \ref{th_representation}) and presents two emblematic cases (in Theorem \ref{lm_invariant_IS} and Theorem \ref{lemma_IS_robust}). In this direction, Section \ref{sub_sec_is_in_learning} illustrates using IS knowledge as a prior for designing encoder-decoder structures in ML. Section \ref{sec_mismatch} focuses on studying the cross-entropy degradation induced by a bias encoder-decoder design (Theorem \ref{th_main_IS_mismatch}), and Section \ref{information_proyection} presents an information projection (IP) interpretation of this analysis (Theorem \ref{th_if_projection_aanlogy}). As a relevant case study, Section \ref{sub_sec_multilayer_IL_analysis} explores the information loss induced by a multi-layer (deep) ML architecture (Theorem \ref{pro_multilayer_info_loss}). Consistency for cross-entropy learning is characterized in Section \ref{sec_ce_learning} (Theorem \ref{th_consistency}) and, in this context, the study of the IB method is presented in Section \ref{sec_IB}. To conclude, Section \ref{sec_numerical} presents an empirical study to illustrate some of the main results presented in this work. The work concludes with a summary and final remarks in Section \ref{sec_structure_analysis}. The proofs of this work's primary results and some support material are relegated to the Appendices.  

\section{Preliminaries}
\label{sec_preliminaries}
Our main object of interest is a joint probability $\mu_{X,Y} \in \mathcal{P}(\mathcal{X}\times \mathcal{Y})$,
where $\mathcal{X}=\mathbb{R}^d$ (for some
$d\in\mathbb{N}_+$) is equipped with the Borel sigma field $\mathcal{B}(\mathcal{X}) = \mathcal{B}(\mathbb{R}^d)$ and $\mathcal{Y}=\left\{1,\ldots,M\right\}$ (for some $M\in\mathbb{N}_+$) is equipped with the power set of $\mathcal{Y}$, $\mathcal{B}(\mathcal{Y})=2^{\mathcal{Y}}$.
$\mu_{X,Y}$ is a probability in  the product space $(\mathcal{X} \times \mathcal{Y},\mathcal{B}(\mathcal{X}\times \mathcal{Y}))$.\footnote{$\mathcal{B}(\mathcal{X}\times \mathcal{Y})$ denote the product sigma field induced by the collection of product events $\mathcal{B}(\mathbb{R}^d) \times 2^{\mathcal{Y}}$ \cite{gray_2009,halmos_1950}.}
%
Moving to the prediction task, we consider a joint vector $(X,Y)\in \mathcal{X} \times \mathcal{Y}$ 
following $\mu_{X,Y} \in \mathcal{P}(\mathcal{X}\times \mathcal{Y})$, where $X$ is the observation and $Y$ is the class label.  
The inference problem is to find a decision rule  $r:\mathcal{X} \longrightarrow \mathcal{Y}$
with the objective of predicting $Y$ from $X$. Let us denote by $\mathcal{F}(\mathcal{X}, \mathcal{Y})$ the family of measurable rules from $\mathcal{X}$ to $\mathcal{Y}$.  Then, the minimum probability of error (MPE) problem is given by: 
\begin{equation}\label{eq_pre_1}
	\ell(\mu_{X,Y}) \equiv \min_{r\in \mathcal{F}(\mathcal{X}, \mathcal{Y})} \ell(r),
\end{equation}
where $\ell(r) \equiv \mathbb{P}(r(X) \neq Y)=\mu_{X,Y}(\left\{(x,y): r(x)\neq y \right\})$ is the probability of error associated to decision rule $r(\cdot)$.\footnote{We denote by $\mathbb{P}$ the probability in the sample space where $X$ and $Y$ (their domain) are defined. It is well known that the MAP rule  $r^*_{\mu_{X,Y}}(x) \equiv \arg \max_{y\in \mathcal{Y}}\mu_{Y|X}(y|x)$ achieves the optimum in (\ref{eq_pre_1}).} 

\subsection{Information-Theoretic Measures}
\label{sub_sec_it_measures}
Mutual information (MI) \cite{cover_2006,gray_1990_b,csiszar_2004} will be used to represent a predictive structure in $\mu_{X,Y}$.  Let us  introduce the entropy and conditional entropy.  Considering our mixed discrete-continuous object $(X,Y)$, the {\em Shannon entropy} of $Y$  is given by \cite{cover_2006}: 
\begin{equation} \label{eq_ps_0}
	H(Y)= \mathcal{H}(\mu_Y)  \equiv  - \sum_{y\in \mathcal{Y}} \mu_{Y} (y) \log  \mu_{Y} (y)  \leq  \log M.
\end{equation}
The conditional entropy of $Y$ given $X$ --- also known as the {\em equivocation entropy} (EE) \cite{feder_1994,ho_2010} ---  is 
\begin{equation} \label{eq_ps_1}
	H(Y|X)   \equiv  
	 \int_{\mathcal{X}} \mathcal{H}(\mu_{Y|X}(\cdot |x))\:d \mu_X(x), 
\end{equation}
where 
$\mathcal{H}(\mu_{Y|X}(\cdot |x)) =  -  \sum_{y\in \mathcal{Y}} \mu_{Y|X} (y|x) \log  \mu_{Y|X} (y|x)$
is 
{\em the Shannon entropy} of  the model $\mu_{Y|X}(\cdot |x)  \in \mathcal{P}(\mathcal{Y})$ \cite{gray_1990_b,cover_2006}. 
Finally, the mutual information (MI) of $\mu_{X,Y}$ is \cite{gray_1990_b,cover_2006}\footnote{The standard notation for the entropy of $Y$ and the MI between $X$ and $Y$ is $H(Y)$ and $I(X;Y)$ respectively \cite{cover_2006}.   However,  we also use $\mathcal{H}(\mu_{Y})$ and  $\mathcal{I}(\mu_{X,Y})$ to emphasize,  in our analysis,  
that these quantities are functionals of the marginal $\mu_Y$ and the joint model $\mu_{X,Y}$, respectively.} 
\begin{equation} \label{eq_ps_3}
	I(X;Y) = \mathcal{I}(\mu_{X,Y})  \equiv H(Y)   - H(Y|X) \geq 0. 
\end{equation}

\subsection{Lossy Compression and Information Sufficiency (IS)}
\label{sub_sec_sufficiency}
A natural strategy adopted in learning to restrict (or regularize) the decision space when solving  (\ref{eq_pre_1}) is the introduction of an encoder of $X$ \cite{bengio_2013,silva_2012b,devroye1996}. The encoder is a measurable 
lossy function $\eta: (\mathcal{X},\mathcal{B}(\mathcal{X})) \longrightarrow (\mathcal{U},\mathcal{B}(\mathcal{U}))$ where $(\mathcal{U},\mathcal{B}(\mathcal{U}))$ denotes the representation space and $U=\eta(X)$ is the induced latent (compressed) variable.  Then, we can say that: 
\begin{definition}\label{def_is}
	An encoder $\eta: \mathcal{X}  \rightarrow \mathcal{U}$ (and  $U=\eta(X)$)  is information sufficient (IS) for $\mu_{X,Y}$ if 
	\begin{equation} \label{eq_ps_7}
		\mathcal{I}(\mu_{X,Y}) =  \mathcal{I}(\mu_{U,Y}),  
	\end{equation}
	where $\mu_{U,Y}$ is the joint  probability of $(U,Y)$ in the representation domain $\mathcal{U}\times \mathcal{Y}$
\end{definition}

It is well-known that $\mathcal{I}(\mu_{X,Y}) -  \mathcal{I}(\mu_{U,Y}) \geq 0$ (by data-processing inequality \cite{cover_2006}), 
and that $\mathcal{I}(\mu_{U,Y})=\mathcal{I}(\mu_{X,Y})$ implies that $\ell(\mu_{U,Y}) = \ell(\mu_{X,Y})$ \cite{silva_2022_aistat}, i.e., IS $\Rightarrow$ sufficiency in the sense of probability of error, where 
$$\ell(\mu_{\eta{(X)},Y}) = {\min_{r\in \mathcal{F}(\mathcal{U},\mathcal{Y})} \mathbb{P}(r(\eta(X)) \neq Y).}$$

\subsection{Cross-Entropy Risk in Learning}
\label{sub_sec_setting_predictive_learning}
For a given model $\mu_{X,Y} \in \mathcal{P}(\mathcal{X}\times \mathcal{Y})$ and a given predictive distribution $v_{\tilde{Y}|\tilde{X}}(\cdot|\cdot) \in \mathcal{P}(\mathcal{Y}|\mathcal{X})$ (produced by a learning agent after the training process), the cross-entropy risk of  $v_{\tilde{Y}|\tilde{X}} (\cdot|\cdot)$ is:
\begin{equation}\label{eq_sec_mis_1}
	r(v_{\tilde{Y}|\tilde{X}}(\cdot|\cdot),\mu_{X,Y}) \equiv \mathbb{E}_{(X,Y)\sim \mu_{X,Y}}  \left\{- \log v_{\tilde{Y}|\tilde{X}}(Y|X) \right\}.
\end{equation}
By the law of large numbers \cite{breiman_1968}, $r(v_{\tilde{Y}|\tilde{X}}(\cdot|\cdot),\mu_{X,Y})$ is the  asymptotic empirical risk of $v_{\tilde{Y}|\tilde{X}}(\cdot|\cdot)$ observed at testing \cite{devroye1996}.
This average risk expresses how good (in average) is the selected model $v_{\tilde{Y}|\tilde{X}}(\cdot|\cdot) \in \mathcal{P}(\mathcal{Y}|\mathcal{X})$ for predicting $Y$ given $X$ in a likelihood sense.\footnote{Interestingly, this risk is tightly related with the cost of compressing (in bits) $Y$ given $X$.} As in the task of lossless compression \cite{cover_2006}, the best risk that can be achieved by any predictive model is lower bounded by an information-theoretic measure.  
For our specific prediction problem  $x\in \mathcal{X} \rightarrow v_{\tilde{Y}|\tilde{X}}(\cdot|x) \in \mathcal{P}(\mathcal{Y})$, this performance bound is given by the {\em conditional Shannon entropy} in (\ref{eq_ps_1}): 
\begin{lemma}\label{th_loss_it_bound}
	For any model $\mu_{X,Y}$ and any predictive distribution $v_{\tilde{Y}|\tilde{X}}(\cdot|\cdot) \in \mathcal{P}(\mathcal{Y}|\mathcal{X})$, we have that 
	\begin{equation}\label{eq_sec_mis_2}
	 	r(v_{\tilde{Y}|\tilde{X}}(\cdot|\cdot),\mu_{X,Y}) \geq  H(Y|X).
	\end{equation}
	Furthermore, the equality in (\ref{eq_sec_mis_2}) is achieved (i.e., optimality) if, and only if, 
	\begin{equation}\label{eq_sec_mis_3}
		v_{\tilde{Y}|\tilde{X}}( \cdot |X) = \mu_{Y|X}(\cdot|X), \ \mu_X-\text{almost surely},
	\end{equation}
	where $\mu_X$ is the marginal distribution of $X$ and the equality in (\ref{eq_sec_mis_3}) is in total variation \cite{devroye_1985,devroye_2001}. 
\end{lemma}

\begin{figure*}
  \centering
   	\includegraphics[width=0.80\textwidth]{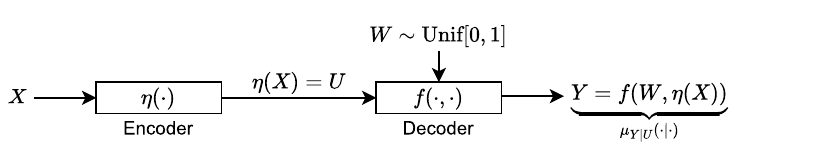}
  \caption{The encoder-decoder structure of $\mu_{Y|X}(\cdot|\cdot)$ when $\mu_{X,Y} \in \mathcal{P}_{\eta}(\mathcal{X}\times \mathcal{Y})$ (see Def.\ref{def_IS_redundant_models}). 
  }
  \label{fig6}
\end{figure*}

\section{IS to Model Predictive Structure in Learning}
\label{sec_probabilistic_structure}
Representation learning focuses on the task of finding a lossy encoder of $X$ that captures all,  or
most of, the information that $X$ has for predicting $Y$.
Aligned with this objective, we are interested in describing the complete class of models $\mu_{X,Y}$ for which a fixed  lossy encoder of $X$ is information sufficient (IS) for $\mu_{X,Y}$ (Def.~\ref{def_is}). 
Importantly, IS,  as a joint condition on the encoder $\eta(\cdot)$ and the model $\mu_{X,Y}$, expresses a strong notion of statistical sufficiency:
$X \rightarrow U \rightarrow Y$, i.e., $X$ and $Y$ are conditional independent given $U=\eta(X)$  \cite{gray_2009,breiman_1968}. Indeed, it is simple to verify that: 
\begin{lemma}\label{lm_is_encode_d-sparation}
$\eta(\cdot)$ is IS for $\mu_{X,Y}$ (Def.~\ref{def_is})  if, and only if, 
$X$ and $Y$ are conditional independent given $U$.  
\end{lemma}
For the rest of this section, we will focus on studying this  IS structure and characterizing the models that meet this strong D-separation requirement $X \rightarrow \eta(X) \rightarrow Y$ for a given encoder $\eta(\cdot)$.

\subsection{A Functional Characterization for Models with a Latent IS Structure}
\label{sec_functional_IS_structure}
The main result of this section provides a functional characterization for models $\mu_{X,Y}$ that have an embedded lossy IS representation. Let us first introduce this class of models: 
\begin{definition}\label{def_IS_redundant_models}	
	Let $\eta:(\mathcal{X},\mathcal{B}(\mathcal{X})) \rightarrow (\mathcal{U},\mathcal{B}(\mathcal{U}))$ be a mesurable function. We denote by $\mathcal{P}_{\eta}(\mathcal{X} \times \mathcal{Y}) \subset  \mathcal{P}(\mathcal{X} \times \mathcal{Y})$  the class of models where if $(X,Y) \sim \mu_{X,Y}\in  \mathcal{P}_{\eta}(\mathcal{X} \times \mathcal{Y})$ then $I(X;Y)=I(\eta(X);Y)$.
\end{definition}
Therefore, $\mathcal{P}_{\eta}(\mathcal{X} \times \mathcal{Y})$ 
is the class of models where $\eta(\cdot)$ is IS for $\mu_{X,Y}$ or, alternatively, the class  of models where $\eta(X)$ $D$-separates $X$ and $Y$, i.e., $X \rightarrow \eta(X) \rightarrow Y$ (from Lemma~\ref{lm_is_encode_d-sparation}).

 \begin{theorem} \label{th_representation}
 	Let us consider $\eta:(\mathcal{X},\mathcal{B}(\mathcal{X})) \rightarrow (\mathcal{U},\mathcal{B}(\mathcal{U}))$. 
$\mu_{X,Y} \in \mathcal{P}_{\eta}(\mathcal{X} \times \mathcal{Y})$ 
if, and only if,  $\exists f:[0,1] \times \mathcal{U} \rightarrow \mathcal{Y}$ (a measurable function) such that for $\mu_X$-almost every $x\in \mathcal{X}$ the conditional distribution of $Y$ given $X=x$ is given by
\begin{equation} \label{eq_ps_9}
	Y=f(W,\eta(x)),
\end{equation}
where $W \sim \emph{Unif}[0,1]$ is a random variable that is insensitive to the choice of $x\in \mathcal{X}$.
\end{theorem}
(The proof of this result is presented in Appendix \ref{proof_th_representation})

Some remarks about this result:
 \begin{itemize}
\item[{i)}] Theorem~\ref{th_representation} offers a precise characterization for the models that have a latent structure determined by a lossy encoder $\eta(\cdot)$. Importantly, the result presents a necessary and sufficient condition to meet $\mu_{X,Y} \in \mathcal{P}_{\eta}(\mathcal{X} \times \mathcal{Y})$.\\
%
\item[ { ii)}]  
Theorem~\ref{th_representation} 
 offers a functional  construction (in (\ref{eq_ps_9})) for all models that meet the IS condition presented in Def.~\ref{def_IS_redundant_models}.
Unlike \cite{Bloem_2019}, this functional description for the class $\mathcal{P}_{\eta}(\mathcal{X} \times \mathcal{Y})$ does not impose any condition on the marginal distribution of $X$ (i.e., on $\mu_X$). Consequently, this IS structure only imposes restrictions on the predictive part $(\mu_{Y|X}(\cdot |x))_{x\in \mathcal{X}}\subset \mathcal{P}(\mathcal{Y})$ of $\mu_{X,Y} \in \mathcal{P}_{\eta}(\mathcal{X} \times \mathcal{Y})$.\\  
%
\item[{ iii)}] Finally in the expression in (\ref{eq_ps_9}),  both $\eta(\cdot)$ and $W$ are fixed, i.e., they are used uniformly to produce each element $\mu_{X,Y}\in \mathcal{P}_{\eta}(\mathcal{X} \times \mathcal{Y})$. Therefore, given $\eta(\cdot)$ and $W \sim \textrm{Unif}[0,1]$,  the expressive power of all measurable functions $f(\cdot)$ from $[0,1] \times \mathcal{U} \rightarrow \mathcal{Y}$ induces the collection of models in $\mathcal{P}_{\eta}(\mathcal{X} \times \mathcal{Y})$.
\end{itemize}

On the relevance of this IS characterization, we show two important classes of models where the presented IS latent structure emerges in ML. More examples of IS structured models, and the interpretation that IS offers for designing expressive encoder-decoder ML algorithms will be covered in  Section \ref{sub_sec_is_in_learning}.

\subsection{Invariance to Transformations}
\label{sec_probabilistic_invariances}
An important class of models with an IS latent structure  
are the models invariant to the action of a compact group \cite{eaton_1889}. 
We prove this connection in Theorem \ref{lm_invariant_IS}.  Models invariant to operations such as rotation, permutations, and scale, among others, have been studied and used as an inductive bias for the design of many ML schemes in image classification and indexing \cite{Bloem_2019, Dubois_2021, zaheer_2017,vasconcelos2004,do_2002,chang_1993}. 

Let us begin with the concept of predictive invariance. 
\begin{definition}\label{def_strong_invariances}
	Given a compact group of measurable transformations $\mathcal{G}= \left\{ g_j, j \in \mathcal{J} \right\}$\footnote{A compact group $\mathcal{G}$ satisfies three properties: for each $g,h\in \mathcal{G}$, $g \circ h\in  \mathcal{G}$, the identity function $id\in \mathcal{G}$ and every $g\in \mathcal{G}$ has an inverse  (i.e., $g^{-1}$ such that $g^{-1}\circ g= id$) and $g^{-1}\in \mathcal{G}$ \cite{eaton_1889,rootma_1995}.},  
	a  model $\mu_{X,Y}$ is said to be predictive invariant 
	w.r.t.  $\mathcal{G}$  (in short $\mu_{Y|X}$ is $\mathcal{G}$-invariant), if $Y|X \overset{d}{=}Y|g(X)$ for any $g\in \mathcal{G}$. More precisely, for any  $g\in \mathcal{G}$ and $A\subset \mathcal{Y}$ the invariant condition 
	\begin{align}\label{eq_pi_1}
		\mu_{Y|X}(A| \left\{x\right\}) =  \mathbb{P}(Y\in A|X=x)
					=  \mathbb{P}(Y\in A| g(X)=x) = \mu_{Y|g(X)}(A|\left\{x\right\})
	\end{align}
	is satisfied for $\mu_X$-almost every point in $\mathcal{X}$. 
We denote by $\mathcal{P}_{\mathcal{G}}(\mathcal{X} \times \mathcal{Y}) \subset \mathcal{P}(\mathcal{X} \times \mathcal{Y})$ the class of $\mathcal{G}$-invariant models. 
\end{definition}

It is known that a  compact group of 
transformations 
$\mathcal{G}$ induces an equivalence relationship in $\mathcal{X}=\mathbb{R}^d$ \cite{eaton_1889,Bloem_2019}: $x \sim z$ if $\exists g\in \mathcal{G}$ such that $z=g(x)$.  This equivalence relationship generates a measurable partition in $\mathcal{X}$ \cite{eaton_1889,Bloem_2019} denoted by $\pi_\mathcal{G}\subset \mathcal{B}(\mathcal{X})$.   
A measurable function $\eta: \mathcal{X} \rightarrow \mathcal{U}$ is said to be maximal invariant if $\eta(x)=\eta(z)$ if, and only if, $x \sim z$ \cite{eaton_1889}.\footnote{On the construction and the existence of maximal invariant transformation for $\mathcal{G}$, please see \cite{eaton_1889} and \cite{Bloem_2019} and references therein. 
} 
\begin{theorem}\label{lm_invariant_IS}
	Let us consider a compact group of  
	transformations $\mathcal{G}= \left\{ g_j, j \in \mathcal{J} \right\}$ equipped with a maximal invariant transformation $\eta^*_\mathcal{G}: \mathcal{X} \rightarrow \mathcal{X}$.  Then,  
	$\mu_{X,Y}$ is predictive invariant w.r.t. $\mathcal{G}$ (Def.~\ref{def_strong_invariances}) if, and only if, $\eta^*_\mathcal{G}(\cdot)$ is IS for $\mu_{X,Y}$ (Def.~\ref{def_is}).
	\\(The proof of this result is presented in Appendix \ref{proof_lm_invariant_IS})
\end{theorem}

Importantly, Theorem~\ref{lm_invariant_IS}  establishes that a necessary and sufficient condition for a model to be predictive invariant w.r.t. $\mathcal{G}$  
is that  $\eta^*_\mathcal{G}(\cdot)$ is IS: i.e.,  $I(\eta^*_\mathcal{G}(X);Y)=I(X;Y)$. 
A direct consequence of Theorem~\ref{th_representation} and Theorem~\ref{lm_invariant_IS} is the following result: 

\begin{corollary}\label{th_representation_invariant}
	$\mu_{X,Y} \in \mathcal{P}_{\mathcal{G}}(\mathcal{X} \times \mathcal{Y})$ if, and only if,  $\exists f:[0,1] \times \mathcal{U} \rightarrow \mathcal{Y}$ such that the conditional distribution of $Y$ given $X=x$ ($\mu_X$-almost surely) is derived by
	\begin{equation} \label{eq_pi_2}
	Y=f(W,\eta^*_\mathcal{G}(x)), 
	\end{equation}
where $W \sim \emph{Unif}[0,1]$ is independent of $X$.
 \end{corollary}
Consistent with the IS definition, Corollary~\ref{th_representation_invariant} establishes a functional characterization for the predictive part of $\mu_{X,Y}\in \mathcal{P}_{\mathcal{G}}(\mathcal{X} \times \mathcal{Y})$.

\begin{remark}
The seminal work by Bloem-Reddy and Teh \cite{Bloem_2019} studied a stronger probabilistic notion of invariance under the action of a compact group $\mathcal{G}$. They addressed the important case where a joint model $\mu_{X,Y}$ is $\mathcal{G}$-invariant 
if for any $g\in \mathcal{G}$ it follows that $(X,Y)=(g(X),Y)$ in distribution when $(X,Y)\sim \mu_{X,Y}$.\footnote{This means that the complete joint distribution $\mu_{X,Y}$ is invariant to the actions of elements of $\mathcal{G}$ and in particular the marginal distribution of $X$ ($\mu_X$).}  
This joint invariant property  is stronger than the predictive $\mathcal{G}$-invariant condition stated in Definition \ref{def_strong_invariances}.\footnote{In fact, if $\mu_{X,Y}$ is $\mathcal{G}$-invariant in the sense that $(X,Y)=(g(X),Y)$ in distribution for any $g\in \mathcal{G}$, then $\mu_{X,Y}$ is predictive $\mathcal{G}$-invariant (Def.~\ref{def_strong_invariances}). See more details about this relationship in \cite[Prop. 1 and Sec. 1.1]{Bloem_2019}.}  Under this joint invariant condition, \cite[Th. 7]{Bloem_2019} showed a functional characterization for $(X,Y) \sim \mu_{X,Y}$ that is equivalent to the statement presented in Corollary~\ref{th_representation_invariant}.
It is worth pointing out that to obtain our result in (\ref{eq_pi_2}),  
we do not require the marginal distribution of $X$ to be  $\mathcal{G}$-invariant. Indeed, this result offers a necessary and sufficient characterization for a model $\mu_{X,Y}$ to be predictive $\mathcal{G}$-invariant and,  consequently,
it  is an improved version of  \cite[Th. 7]{Bloem_2019}.
\end{remark}

\subsection{Robustness to Perturbations}
\label{sec_robust}
Another important IS latent structure worth covering is inspired by the notion of robustness in learning  introduced in \cite{xu_2012}. The idea is that the predictive part of a model (seen in \cite{xu_2012} as the output of a learning process), i.e., $\mu_{Y|X}(\cdot|\cdot)$, is robust (insensitive) 
to some level of observation perturbation 
if this degradation (a form of external noise or adversarial attack) happens within the cells  
of a partition of $\mathcal{X}$. For formalizing this concept, let us consider an indexed measurable partition $\pi_\mathcal{I}= \left\{A_i,i\in \mathcal{I} \right\}\subset \mathcal{B}(\mathcal{X})$ indexed by a set $\mathcal{I}$.

\begin{definition}\label{def_robust}
	A model $\mu_{X,Y}$ is said to be robust to perturbations within the cells of $\pi_\mathcal{I}= \left\{A_i,i\in \mathcal{I} \right\}$, if for any $i\in \mathcal{I}$ and any two points $x,\bar{x}\in A_i$ we have that $\mu_{Y|X}(\cdot|x)=\mu_{Y|X}(\cdot|\bar{x})$.
\end{definition}
The robust structure described in Def.~\ref{def_robust}  expresses a level of critical resolution in the input space $\mathcal{X}$ (dictated by the cells of $\pi_\mathcal{I}$) at which $X$ has discrimination about $Y$. This can be expressed in the language of IS (Def.~\ref{def_is}).  For this, we note that the indexed partition $\pi_\mathcal{I}$ can be equipped with an associated lossy encoder $\eta_{\pi}(\cdot)$.  $\eta_{\pi}(\cdot)$ maps any point $x\in \mathcal{X}$ to a fixed element $x_i\in A_i$ for which we just need to select a representative point $x_i \in \mathcal{X}$ belonging to every cell $A_i$.  Given this encoder, which  is vector quantizer (VQ) \cite{gray_1998_b,gray_1990_b}, we have the following result: 
%
\begin{theorem} \label{lemma_IS_robust}
	A model $\mu_{X,Y}$ is robust w.r.t. $\pi_\mathcal{I} = \left\{A_i,i\in \mathcal{I} \right\}$ (Def.~\ref{def_robust}) if, and only if, $\eta_{\pi}(\cdot)$ is IS for $\mu_{X,Y}$ (Def.~\ref{def_is}). 
\end{theorem}
(The proof of this result is presented in Appendix \ref{proof_lemma_IS_robust})

From Theorem~\ref{lemma_IS_robust}  and Theorem~\ref{th_representation}, we could derive a functional description that fully characterizes the class of models that are robust to perturbations within the cells of $\pi_\mathcal{I}$ (as stated in Eq.~(\ref{eq_ps_9}) but using $\eta_{\pi}(\cdot)$ as the encoder).  As in Corollary~\ref{th_representation_invariant}, the encoder $\eta_{\pi}(\cdot)$ fully characterizes this family. An example of this class (with its respective functional characterization) is presented in Section~\ref{sub_sec_digital} for the discrete case  when $ \left| \mathcal{I}  \right| <\infty$, i.e., when $\eta_{\pi}(\cdot)$ is a digital compressor. 

\subsection{Interpreting IS for Encoder-Decoder Design in ML}
\label{sec_IS_ML_algorithms}
Theorem~\ref{th_representation} tells us that the predictive part of $\mu_{X,Y} \in \mathcal{P}_{\eta}(\mathcal{X} \times \mathcal{Y}) $ is given by the following functional structure: 
\begin{equation}\label{eq_IS_ML_1}
	Y=f(W,\eta(X)).
\end{equation}
In (\ref{eq_IS_ML_1}), we recognize two key design elements commonly used in modern ML architecture \cite{bengio_2013,alemi_2017,kingma_2014,achille_2018}.
On the one hand, the role of an {\bf encoder}, represented by $\eta(\cdot)$ in (\ref{eq_IS_ML_1}).  The encoder represents the redundancy of $\mu_{Y|X}(\cdot|\cdot)$ in the sense of the existence of a compressed  
representation of $X$ (or latent variable) that is sufficient (Def.~\ref{def_is}). On the other hand, an {\bf stochastic decoder} that maps elements from $U=\eta(X)\in \mathcal{U}$, i.e., the latent domain, to a predictive model $\mu_{Y|U}(\cdot | u) \in \mathcal{P}(\mathcal{Y})$ for all $u\in \mathcal{U}$. In this second stage, the posterior distribution is derived by a functional equation indexed by  $f(\cdot, \cdot)$ and an universal noise $W\sim \textrm{Unif}[0,1]$. 
Interestingly, $W$ can be seen as the nuisance variable, 
which models the random interference between $U$ and $Y$. Figure~\ref{fig6} summarizes the encoder-decoder predictive structure 
presented in (\ref{eq_IS_ML_1}) when $\mu_{X,Y}\in \mathcal{P}_{\eta}(\mathcal{X} \times \mathcal{Y})$.  Further analyses about the interpretation of this functional characterization when adopting a learning scheme with 
an encoder-decoder structure will be presented in Sections \ref{sub_sec_is_in_learning} and \ref{sec_mismatch}.

\section{Assuming an IS (Encoder-Decoder) Structure in Learning}
\label{sub_sec_is_in_learning}
In ML,  we do not know $\mu_{X,Y}$. Still, we could access prior knowledge indicating that $\mu_{X,Y}  \in \mathcal{P}_{\eta}(\mathcal{X}\times \mathcal{Y})$, for example, that the model is permutation invariant, robust or sparse as illustrated below in  Sections~\ref{sub_sec_digital} and \ref{sub_sec_sparse_canonical}.  In light of Theorem~\ref{th_representation},  IS 
can be used as a prior (inductive bias) for the design of probabilistic ML algorithms. 
Assuming   that $\mu_{X,Y} \in \mathcal{P}_{\eta}(\mathcal{X}\times \mathcal{Y})$,  for some encoder $\eta(\cdot)$, 
the result in (\ref{eq_IS_ML_1})  means that learning the predictive model $\mu_{Y|X}(\cdot|\cdot)$  within $\mathcal{P}_{\eta}(\mathcal{X}\times \mathcal{Y})$ reduces to ``estimating the function $f(\cdot, \cdot)$ from data". 
For this estimation task, we could use the machinery of deep neural networks (DNN) \cite{hornik_1989,zhou_2020} and the well-known reparametrization trick \cite{kingma_2014, rezende_2014, alemi_2017}  to index a collection of expressive measurable functions $\left\{f_\theta(\cdot, \cdot): [0,1] \times \mathcal{U} \rightarrow \mathcal{Y}, \theta \in \Theta \right\}$ (indexed by the parameters of a neural network denoted by $\theta$)  and use standard learning algorithms to select $\hat{\theta}(S_n) \in \Theta$ from $S_n=  \left\{ (\eta(X_1),Y_1)),\ldots, (\eta(X_n),Y_n)) \right\}$. Here, $S_n$  is the supervised data projected on the representation domain $\mathcal{U} \times \mathcal{Y}$. In this framework, $\eta(\cdot)$ is acting as a pre-encoder that projects the data from $\mathcal{X} \times \mathcal{Y}$ to a representation domain $\mathcal{U} \times \mathcal{Y}$. 
This pre-encoder 
offers a practical way to learn a data-driven predictive model $Y=f_{\hat{\theta}(S_n)}(W,\eta(X))$ that belongs to 
$\mathcal{P}_{\eta}(\mathcal{X}\times \mathcal{Y})$ (from Theorem~\ref{th_representation}) with standard DNN architectures and data-fitting algorithms. 
Following this path, we illustrate some relevant classes of IS structured models (organized by their encoder) and their consistent neural network encoder-decoder architecture.

\subsection{Digital Models: Vector Quantizer} 
\label{sub_sec_digital}
Inspired by the important role played by digitalization in 
machine learning (see Section \ref{sec_robust}), we introduce the collection of models that are $D$-separable by a vector quantizer (VQ)\cite{gray_1998_b,gersho_1992,gray_1990}. Digitalization has been used extensively as a preprocessor in pattern recognition and ML applications \cite{Dubois_2021,alemi_2017,strouse_2017,tishby_1999,amjad_2019,tishby_2015}. 
Consequently, the collection of digital models, presented in (\ref{eq_ps_12}), is a relevant object to study.  
Let us consider a finite measurable partition of $\mathcal{X}$  of size $K\in\mathbb{N}_+$,  
 $\pi=\left\{A_i, i=1,\ldots,K \right\}$, 
and its respective VQ (analog to digital converter):\footnote{${\bf 1}_{A_i}(\cdot)$ denotes the indicator function of the set $A\subset \mathcal{X}$.}
\begin{equation} \label{eq_ps_11}
\eta_\pi(x) = \sum_{i=1}^K {\bf 1}_{A_i}(x) \cdot \hat{x}_i  \in \mathcal{X},
\end{equation}  
where $\hat{x}_i$ is an element in $A_i$.  Then, we can consider the class that are $D$-separable by  $\eta_\pi(\cdot)$:
\begin{equation} \label{eq_ps_12}
\mathcal{P}_\pi(\mathcal{X} \times \mathcal{Y}) \equiv \left\{\mu \in  \mathcal{P}(\mathcal{X} \times \mathcal{Y}),\:\text{s.t.,}\:{I}({X;Y}) = {I}({\eta_\pi(X);Y}) \right\}.
\end{equation}  
Importantly, from Theorem~\ref{th_representation}, we have the following functional characterization for $\mathcal{P}_\pi(\mathcal{X} \times \mathcal{Y})$: 
\begin{corollary}\label{cor_digital_class}
	$\mu_{X,Y}\in \mathcal{P}_\pi(\mathcal{X} \times \mathcal{Y})$ if, and only if, the distribution of $Y$ given $X=x$ can be obtained by  $Y=f(W,\eta_\pi(x))$, 
	where $W\sim \emph{Unif}[0,1]$ and $f:[0,1] \times  \left\{\hat{x}_1,\ldots,\hat{x}_K \right\} \rightarrow \mathcal{Y}$ is a measurable function. 
\end{corollary}

{\bf Digital Neural Networks (Di-NN)}:
A Di-NN is a neural network where the initial layer is a VQ in the form presented in Eq.~(\ref{eq_ps_11}). 
After that,  we can have deep expressive layers to represent the function $f(\cdot, \cdot)$ in Eq.~(\ref{eq_IS_ML_1}). 
This makes the Di-NNs functional expressive and consistent with the  assumption that we are learning within $\mathcal{P}_\pi(\mathcal{X} \times \mathcal{Y})$. 

\subsection{Sparse Models: Feature Selector} 
\label{sub_sec_sparse_canonical}
Inspired by sparse assumption used in signal processing  \cite{donoho_2006,candes_2006,candes_2008,candes_2006b,candes_2005,cohen_2009,silva_2015, silva_2022} and feature selection in machine learning 
\cite{kim2021dropbottleneck,tibshirani1996sparse-lasso,molchanov2016pruningcnn},
let us consider a feature selector of size $q<d$ of 
the form $\eta_{j_1,\ldots,j_q}: \mathbb{R}^d \rightarrow  \mathbb{R}^q$ where for any $\bar{x}=(x_1,\ldots,x_d)\in \mathcal{X}$,  $\eta_{j_1,\ldots,j_q}(\bar{x})=(x_{j_1},\ldots,x_{j_q}) \in \mathbb{R}^q$. 
\footnote{$\eta_{j_1,\ldots,j_q}(\cdot)$ is a linear operator $\eta_{j_1,\ldots,j_q}(\bar{x})= P_{j_1,\ldots,j_q} \cdot \bar{x}$, being  $P_{j_1,\ldots,j_q}$ a projection matrix of dimension $q\times d$ where the $k$-row of $P_{j_1,\ldots,j_q}$ is $\bar{e}_{j_k}^\mathsf{T}$ and  $\left\{\bar{e}_j, j=1,\ldots,d \right\}$ denotes the canonical orthonormal basis for $\mathbb{R}^d$. } 
Then, we  introduce the family of $q$-sparse models in the components $j_1,\ldots,j_q\in \left\{1,\ldots,d \right\}$ as follows: $\mathcal{P}_{\eta_{j_1,\ldots,j_q}}(\mathcal{X} \times \mathcal{Y}) \equiv$ 
\begin{equation} \label{eq_ps_13b}
 \left\{\mu_{X,Y} \in  \mathcal{P}(\mathcal{X} \times \mathcal{Y}),\:\text{s.t.,}\:{I}({X;Y}) = {I}({\eta_{j_1,\ldots,j_q}(X);Y}) \right\}.
\end{equation}  
$\mu_{X,Y}\in \mathcal{P}_{\eta_{j_1,\ldots,j_q}}$ means that $X_{j_1},X_{j_2},\ldots,X_{j_q}$ is IS, 
 i.e.,  $I(X;Y)= I(X_{j_1},X_{j_2},\ldots,X_{j_q};Y)$. 
From Theorem~\ref{th_representation}, we have the following functional description for $\mathcal{P}_{\eta_{j_1,\ldots,j_q}}(\mathcal{X} \times \mathcal{Y})$: 
\begin{corollary}\label{cor_sparse_class}
	$\mu_{X,Y}\in \mathcal{P}_{\eta_{j_1,\ldots,j_q}}(\mathcal{X} \times \mathcal{Y})$ if, and only if, the distribution of $Y$ given $X=x$ (for $\mu_X$-almost every point) can be obtained by 
		$Y=f(W,\eta_{j_1,\ldots,j_q}(x))$
	where $W\sim \emph{Unif}[0,1]$ and $f:[0,1] \times \mathbb{R}^q \rightarrow \mathcal{Y}$ is a measurable function. 
\end{corollary}

{\bf Sparse Neural Networks (S-NN)}:
To learn $\mu_{X,Y}$ within $\mathcal{P}_{\eta_{j_1,\ldots,j_q}}(\mathcal{X} \times \mathcal{Y})$,  
the first layer of a S-NN encodes the operation $P_{j_1,\ldots,j_{q}} \cdot \bar{x}$, where  
$P_{j_1,\ldots,j_{q}}$ is a linear projection  (a point-to-point layer) 
that can be interpreted as a pooling layer (or feature selector). After that,  we can have fully connected expressive layers to represent the function $f(\cdot, \cdot)$ in Eq.~(\ref{eq_IS_ML_1}). 
This encoder-decoder structure makes a network expressive and fully consistent with the functional structure of $\mathcal{P}_{\eta_{j_1,\ldots,j_q}}(\mathcal{X} \times \mathcal{Y})$ in Corollary~\ref{cor_sparse_class}.

\subsection{More IS Classes} 
\label{sub_sec_morel}
Other relevant classes of models with an IS structure and their consistent encoder-decoder architectures are presented in Appendix \ref{sec_IS_ML_algorithms}. These include transform-based sparse models, which utilize a linear full-rank projection as its encoder and permutation-invariant models, which uses the empirical distribution or set as its encoder. 

\section{IS Mismatch Analysis: Encoder-Decoder Expressive Analysis in Learning}
\label{sec_mismatch}
In Section~\ref{sub_sec_is_in_learning}, we used Theorem~\ref{th_representation} to inform the design a ML encoder-decoder architecture that  is consistent with the assumption that $\mu_{X,Y}  \in \mathcal{P}_{\eta}(\mathcal{X}\times \mathcal{Y})$ (see Def.~\ref{def_IS_redundant_models}).  
Now, we consider the more practical scenario where  this structural assumption does not hold. 
In other words, we look at the performance cost to pay (if any) when using an encoder-decoder structure that follows the functional assumption of  $\mathcal{P}_{\eta}(\mathcal{X}\times \mathcal{Y})$ (see Eq.(\ref{eq_IS_ML_1}) in Section \ref{sec_probabilistic_structure}), however this IS condition is not met by the true  data generating distribution $\mu_{X,Y}$, i.e., we have that $\mu_{X,Y}  \notin \mathcal{P}_{\eta}(\mathcal{X}\times \mathcal{Y})$. 
From Def.~\ref{def_is}, this means that 
\begin{equation}\label{eq_sec_mismatch_0}
I(X;Y|U=\eta(X)) = \underbrace{I(X;Y) - I(U;Y)}_{\text{mutual information loss (MIL)}} > 0. 
\end{equation}
We use the cross-entropy risk in Eq.(\ref{eq_sec_mis_1}) as the performance indicator for this mismatch expressive analysis.  The main result of this section (Theorem \ref{th_main_IS_mismatch})  shows that $I(X;Y|U)>0$, i.e., the mutual information loss (MIL)  induced by $\eta(\cdot)$ in $\mu_{X,Y}$ in \eqref{eq_sec_mismatch_0}, precisely expresses this cross-entropy performance degradation. 

\subsection{The IS Mismatch Result}
\label{main_result_mismatch}
The next result  uses Theorem~\ref{th_representation} to quantify the effect of 
using an encoder-decoder architecture 
that is consistent with the hypothesis that $\eta(\cdot)$ is IS  (see Section \ref{sub_sec_is_in_learning} and Figure~\ref{fig6}). For the statement, we need to introduce some learning concepts.

Following the design strategy presented in Section \ref{sub_sec_is_in_learning}, for a given $\eta: \mathcal{X} \rightarrow \mathcal{U}$, let us denote by\footnote{For any $f_\theta(\cdot,\cdot)$, $v^\theta_{\tilde{Y}|U}(\cdot|u)$ in (\ref{eq_sec_mis_3b}) denotes the conditional distribution induced by the expression $\tilde{Y}=f_\theta(W,u)$ for any $u\in \mathcal{U}$ and $W\sim \textrm{Unif}[0,1]$.}  
\begin{equation}\label{eq_sec_mis_3b}
\Lambda_{\Theta,\eta} \equiv \left\{v^\theta_{\tilde{Y}|U}(\cdot|\eta(\cdot)), \theta \in \Theta \right\}\subset \mathcal{P}(\mathcal{Y}|\mathcal{X})
\end{equation}
the collection of predictive distributions  induced by a family of measurable functions $\left\{f_\theta : [0,1]\times\mathcal{U}  \rightarrow \mathcal{Y}, \theta \in \Theta  \right\}$, indexed by $\Theta$, and the encoder($\eta(\cdot)$)-decoder($f_\theta(\cdot,\cdot)$) structure that is consistent with assuming that $\eta(\cdot)$ is IS (see Eq.(\ref{eq_IS_ML_1}) and Theorem~\ref{th_representation}).
For performance, we consider i.i.d. samples $S_n=(X_1,Y_1),....,(X_n,Y_n)$ generated from the true (data-generated) 
distribution $\mu_{X,Y}$, where the empirical cross entropy risk of $v^\theta_{\tilde{Y}|U}(\cdot|\cdot) \in \Lambda_{\Theta,\eta}$ is 
\begin{equation}\label{eq_sec_mis_4}
	 \hat{r}(v^\theta_{\tilde{Y}|U}(\cdot|\eta(\cdot)),S_n) \equiv - \frac{1}{n} \sum^n_{i=1} \log v^\theta_{\tilde{Y}|U}(Y_i|\eta(X_i)).
\end{equation}

\begin{theorem}\label{th_main_IS_mismatch}
Let us consider an unknown model $\mu_{X,Y} \in \mathcal{P}(\mathcal{X}\times \mathcal{Y})$, a lossy encoder $\eta: \mathcal{X} \rightarrow \mathcal{U}$ (the inductive bias) and family of functions $\left\{f_\theta : [0,1]\times\mathcal{U} \rightarrow \mathcal{Y}, \theta \in \Theta  \right\}$ (the hypothesis space) to be selected by a learning agent (using some training resources).  If we have i.i.d. samples $S_n=(X_1,Y_1),....,(X_n,Y_n)$ generated from $\mu_{X,Y}$ (at testing time), 
\begin{itemize}
\item[i)] for any possible candidate $v^\theta_{\tilde{Y}|U}(\cdot|\cdot) \in \Lambda_{\Theta,\eta}$,  it follows that
\begin{align}\label{eq_sec_mis_5a}
	 	\lim_{n \rightarrow \infty}\hat{r}(v^\theta_{\tilde{Y}|U}(\cdot|\eta(\cdot)),S_n) &= r(v^\theta_{\tilde{Y}|U}(\cdot|\eta(\cdot)),\mu_{X,Y}) \geq\nonumber\\  
		&\underbrace{I(X;Y|U)}_{\substack{\emph{cross-entropy}\\\emph{penalization}}} + H(Y|X),
\end{align}
where the convergence in (\ref{eq_sec_mis_5a}) is almost surely with respect to the process distribution of $(X_n,Y_n)_{n\geq 1}$. 
\item[ii)] Furthermore, the 
lower bound $I(X;Y|U) + H(Y|X)$ in (\ref{eq_sec_mis_5a}) is achieved if, and only if, the selected model satisfies the following matching condition 
\begin{equation}\label{eq_sec_mis_5b}
	 	v^\theta_{\tilde{Y}|U}(\cdot|\eta(X)) = \mu_{Y|U}(\cdot|\eta(X)), \mu_X-\text{almost surely},  
\end{equation}
where $\mu_{Y|U}(\cdot|\cdot) \in \mathcal{P}(\mathcal{Y}|\mathcal{U})$ represents the true conditional distribution of $Y$ given $U=\eta(X)$.  
\end{itemize}
\end{theorem}
(The proof is presented in Appendix \ref{proof_th_main_IS_mismatch})

Interpretations of Theorem~\ref{th_main_IS_mismatch}:
 \begin{itemize}
 \item[i)] Theorem~\ref{th_main_IS_mismatch} offers an achievable performance bound for the task of cross-entropy learning: $I(X;Y|U) + H(Y|X)$. This information quantity is a function of $\mu_{X,Y}$ and $\eta(\cdot)$, and it is independent of the learning agent and the functional expressiveness of 
 $\left\{f_\theta : [0,1]\times\mathcal{U} \rightarrow \mathcal{Y}, \theta \in \Theta  \right\}$. 
 \item[ii)]  Contrasting this result with Lemma~\ref{th_loss_it_bound}, we notice that the fact that $\eta(\cdot)$ is not IS (for the unknown model $\mu_{X,Y}$) translates into a performance degradation that is measured by $I(X;Y|U)> 0$ in (\ref{eq_sec_mismatch_0}).  This expression is the MIL induced by the encoder $\eta(\cdot)$  \cite{silva_2022_aistat}. Then, this information loss maps directly to an increase in the minimum expected risk that we could achieve: MIL = cross entropy degradation.  
 \item[iii)] In other words, $I(X;Y|U)\geq 0$ measures the ``{\em lack of cross-entropy expressiveness}'' of the hypothesis space $\Lambda_{\Theta,\eta}= \left\{v^\theta_{\tilde{Y}|U}(\cdot|\eta(\cdot)), \theta \in \Theta \right\}\subset \mathcal{P}(\mathcal{Y}|\mathcal{X})$ that is consistent with an $\eta(\cdot)$-decoder structure (see Figure~\ref{fig6}). 
 \item[iv)]  From the encoder perspective $\eta(\cdot)$,  Theorem~\ref{th_main_IS_mismatch} tells us that zero (cross-entropy) degradation is obtained if, and only if, $\mu_{X,Y} \in \mathcal{P}_{\eta}(\mathcal{X} \times \mathcal{Y})$. Therefore, Theorem~\ref{th_main_IS_mismatch} confirms that each of the encoder-decoder architectures illustrated in Section~\ref{sub_sec_is_in_learning} 
are expressive for their respective classes of models in the cross-entropy sense. 
 \item[v)] From the decoder perspective $f_\theta(\cdot,\cdot)$, Theorem~\ref{th_main_IS_mismatch} 
 determines the condition of an expressive decoder stage in (\ref{eq_sec_mis_5b}):  
decoder-optimality,  in the sense of achieving $I(X;Y|U) + H(Y|X)$ in (\ref{eq_sec_mis_5a}), is met if, and only if, the selected predictive distribution matches the true posterior $\mu_{Y|U}(\cdot|\cdot) $ in the strong total variational sense expressed in (\ref{eq_sec_mis_5b}). In the 
next section, we elaborate further on this decoder expressiveness 
by using  
Theorem~\ref{th_representation} in conjunction with Theorem~\ref{th_main_IS_mismatch}.
 \end{itemize}
 
\subsection{Decoder Probabilistic Expressiveness}  
\label{sub_sec_decoder_expresiviness}
For any 
fixed encoder $\eta(\cdot)$, Theorem \ref{th_main_IS_mismatch} provides a necessary and sufficient condition to achieve the minimum expected risk $I(X;Y|U=\eta(X))+ H(Y|X)$  in (\ref{eq_sec_mis_5b}). This condition 
stipulates a strong expressiveness requirement 
for the class $\Lambda_{\Theta,\eta}= \left\{v^\theta_{\tilde{Y}|U}(\cdot|\eta(\cdot)), \theta \in \Theta \right\}\subset \mathbb{P}(\mathcal{Y}|\mathcal{X})$ induced by $\left\{f_\theta : \mathcal{U}\times [0,1] \rightarrow \mathcal{Y}, \theta \in \Theta  \right\}$,  that is worth expressing it formally:

\begin{definition}\label{def_prob_expressiviness}
	A class of predictive models $\Lambda_{\Theta,\eta}= \left\{v^\theta_{\tilde{Y}|U}(\cdot|\eta(\cdot)), \theta \in \Theta \right\}\subset \mathcal{P}(\mathcal{Y}|\mathcal{X})$, induced by a family of measurable functions $\left\{f_\theta : \mathcal{U}\times [0,1] \rightarrow \mathcal{Y}, \theta \in \Theta  \right\}$ (decoders) and $\eta(\cdot)$ (the encoder), is said to be expressive if $\forall \mu_{X,Y}\in \mathcal{P}(\mathcal{X}\times \mathcal{Y})$ $\exists \theta \in \Theta$ such that 
	$$v^\theta_{\tilde{Y}|U}(\cdot|\eta(X)) = \mu_{Y|U}(\cdot|\eta(X)), \mu_X-\text{almost surely}.$$
\end{definition}
The condition in Definition \ref{def_prob_expressiviness} is very strong: it says that $\left\{v^\theta_{\tilde{Y}|U}(\cdot|\cdot), \theta \in \Theta \right\}$ covers all the elements in $\mathcal{P}(\mathcal{Y}|\mathcal{U})$, which is the complete collection of conditional distribution from $\mathcal{U}$ to $\mathcal{Y}$ \cite{gray_2009}. Importantly from Theorem \ref{th_representation}, 
we can derive a universal (distribution-free) functional expressiveness condition for the class of transformations $\left\{f_\theta : \mathcal{U}\times [0,1] \rightarrow \mathcal{Y}, \theta \in \Theta  \right\}$. The result is the following: 

\begin{lemma}\label{lemma_functional_expressiviness}
	A  sufficient condition for $\Lambda_{\Theta,\eta}$ to be expressive 
	(Definition \ref{def_prob_expressiviness}) is that for any measurable function $f: [0,1] \times \mathcal{U} \rightarrow \mathcal{Y}$ there exists $\theta \in \Theta$ such that $f(W,u)=f_\theta(W, u)$ $\mu_W$-almost surely for all $u\in \mathcal{U}$ where $W\sim \mu_W = \emph{Unif}[0,1]$.\\ 
	(The proof of Lemma \ref{lemma_functional_expressiviness} is presented in Appendix \ref{proof_lemma_functional_expressiviness})
\end{lemma}
Lemma \ref{lemma_functional_expressiviness}  is relevant as it connects functional expressiveness (over $\{ f_\theta:[0,1]\times\mathcal{U}\rightarrow\mathcal{Y},\theta\in\Theta \}$) with probabilistic expressiveness (over $\{v^\theta_{\tilde{Y}|U}(\cdot|\cdot),\theta\in\Theta\}$) for which the functional characterization of Theorem \ref{th_representation} is instrumental.
Finally, we have the following expressiveness 
result: 
\begin{corollary}\label{cor_decoder_expressiveness}
	Let $\eta: \mathcal{X} \rightarrow \mathcal{U}$ be a lossy encoder. If the class $\Lambda_{\Theta,\eta}= \left\{v^\theta_{\tilde{Y}|U}(\cdot|\eta(\cdot)), \theta \in \Theta \right\}\subset \mathcal{P}(\mathcal{Y}|\mathcal{X})$ is expressive for $\mathcal{P}_\eta(\mathcal{X}\times \mathcal{Y})$ (Definition \ref{def_prob_expressiviness}), 
		for any $\mu_{X,Y}\in \mathcal{P}(\mathcal{X}\times \mathcal{Y})$ there exists 
		$v^\theta_{\tilde{Y}|U}(\cdot|\eta(\cdot)) \in \Lambda_{\Theta,\eta}$ that is cross-entropy 
		optimal (from Theorem \ref{th_main_IS_mismatch}),  
		i.e., $r(v^\theta_{\tilde{Y}|U}(\cdot|\eta(\cdot)), \mu_{X,Y})=I(X;Y|U) + H(Y|X)$.
\end{corollary}

\begin{remark}
Achieving the expressiveness condition in Definition \ref{def_prob_expressiviness} with a parametric family of functions, i.e., $\Lambda_{\Theta,\eta}$, 
is not evident. It means we can reproduce any conditional distribution from $\mathcal{U}$ to $\mathcal{Y}$ with the functions $\left\{f_\theta :  [0,1] \times \mathcal{U} \rightarrow \mathcal{Y}, \theta \in \Theta  \right\}$.  This objective might look simple to verify, but it is not, particularly when the latent space is dense and continuous, for example, $(\mathbb{R}^q,\mathcal{B}(\mathbb{R}^q))$. A rich literature in ML shows the capacity of multilayer networks (more recently, deep neural networks (DNN) and convolutional neural networks (CNN)) to be universal approximators of complex continuous functions 
\cite{cybenko_1989,hornik_1989,montufar_2014,poole_2016,delalleau_2011,mhaskar_2016,telgarsky_2016,zhou_2020,shen2022approximation,mao2022approximation}. Establishing a connection between well-known functional expressiveness results and predictive expressiveness (as Def. \ref{def_prob_expressiviness}) is a relevant area of theoretical exploration that is challenging when $\mathcal{U}$ is a finite-dimensional continuous space.  In this context, Theorem \ref{th_representation} and Lemma \ref{lemma_functional_expressiviness}   offer a bridge to connect  functional approximation properties with the probabilistic expressiveness stated in Definition \ref{def_prob_expressiviness}. 
\end{remark}

\section{The Information Projection (IP) Analogy for Encoder Expressiveness}
\label{information_proyection}
Focusing on the expressive role of the encoder $\eta(\cdot)$,
here we show that the task of learning over a class of predictive models consistent with $\eta(\cdot)$ being IS, presented in Theorem \ref{th_main_IS_mismatch}, is equivalently as the task of projecting the true model $\mu_{X,Y}$ into its closest representative in $\mathcal{P}_\eta(\mathcal{X} \times \mathcal{Y})$ (see Def.\ref{def_IS_redundant_models}).  

For making this connection, we assume the ideal scenario that for any encoder $\eta(\cdot)$, the collection $\left\{f_\theta : \mathcal{U}\times [0,1] \rightarrow \mathcal{Y}, \theta \in \Theta  \right\}$ meets the sufficient condition stated in Lemma \ref{lemma_functional_expressiviness}  and, consequently, we have a fully expressive decoder stage in the strong probabilistic sense declared in Definition \ref{def_prob_expressiviness}.
Under this assumption, Corollary \ref{cor_decoder_expressiveness} shows that 
it is feasible to achieve the  information lower bound $I(X;Y|U) + H(Y|X)$ in  Eq.~(\ref{eq_sec_mis_5a}).  
Achieving this lower bound 
can be seen as projecting the true model $\mu_{X,Y}$ into its closest representative in $\mathcal{P}_\eta(\mathcal{X} \times \mathcal{Y})$. 
More precisely, we have that: 
\begin{theorem}\label{th_if_projection_aanlogy}
Let us consider a joint distribution $\mu_{X,Y}$ and a  given lossy encoder $\eta: \mathcal{X} \rightarrow \mathcal{U}$.  
Under the assumption that $\Lambda_{\Theta,\eta}= \left\{v^\theta_{\tilde{Y}|U}(\cdot|\eta(\cdot)), \theta \in \Theta \right\}\subset \mathbb{P}(\mathcal{Y}|\mathcal{U})$ is expressive (see Def. \ref{def_prob_expressiviness}), selecting the optimal decoder in the  sense stated in Eq.(\ref{eq_sec_mis_5b}) (Theorem \ref{th_main_IS_mismatch})   reduces to solving the following information projection (IP) task
\begin{equation}\label{eq_app_IS_projection_1}
	\min_{\tilde{\mu}_{X,Y} \in \mathcal{P}_\eta(\mathcal{X} \times \mathcal{Y})} D(\mu_{X,Y} || \tilde{\mu}_{X,Y}),   
\end{equation}
where the optimal solution of (\ref{eq_app_IS_projection_1})  has the following IS factorization $\mu_X \cdot \mu_{Y|U} \in \mathcal{P}_\eta(\mathcal{X} \times \mathcal{Y})$ (with $U=\eta(X)$). 
\end{theorem}
(The proof of Theorem \ref{th_if_projection_aanlogy}  is presented in Appendix \ref{app_IS-projection}) 

Some remarks about Theorem \ref{th_if_projection_aanlogy}: 
\begin{itemize}
\item[i)] The problem in (\ref{eq_app_IS_projection_1}) selects the closest model to $\mu_{X,Y}$ over the class 
that is consistent with the assumption that $\eta(\cdot)$ is IS (Theorem \ref{th_representation}) in KL divergence sense \cite{cover_2006}. 
Furthermore, it follows that 
\begin{equation}\label{eq_infp_2}
	\min_{\tilde{\mu}_{X,Y} \in \mathcal{P}_\eta(\mathcal{X} \times \mathcal{Y})} D(\mu_{X,Y} || \tilde{\mu}_{X,Y})=I(X;Y|U),
\end{equation}
with an optimal projected model\footnote{This projected model is consistent with the latent IS structure: $X\rightarrow U \rightarrow Y$.} given 
by $\tilde{\mu}^*_{X,Y}=\mu_X \cdot \mu_{Y|U}\in \mathcal{P}_\eta(\mathcal{X} \times \mathcal{Y})$ where $\mu_{Y|U}\in \mathcal{P}(\mathcal{Y}|\mathcal{U})$ is 
the predictive distribution of the joint vector $(U=\eta(X),Y)\sim \mu_{U,Y}$ induced by $\mu_{X,Y}$ and $\eta(\cdot)$.
\item[ii)] 
$I(X;Y|U)\geq 0$ measures the projection error (or lack of expressiveness) of $\mathcal{P}_\eta(\mathcal{X} \times \mathcal{Y})$ to represent $\mu_{X,Y}$ in the KL sense, in analogy with the result in Theorem \ref{th_main_IS_mismatch} that shows that $I(X;Y|U)$ measures the cross-entropy degradation (structural bias) induced by the class of predictive models $\Lambda_{\Theta,\eta}$ in (\ref{eq_sec_mis_3b}). 
\item[iii)] We arrive to this equivalent IP result 
from the two main results of this work: Theorems \ref{th_representation} and \ref{th_main_IS_mismatch}. 
\end{itemize}

In conclusion, learning with an encoder($\eta(\cdot)$)-decoder structure is equivalent to projecting $\mu_{X,Y}$ on $\mathcal{P}_\eta(\mathcal{X} \times \mathcal{Y})$  and inducing a 
performance degradation (the equivalent approximation error in (\ref{eq_app_IS_projection_1})) given by $I(X;Y|U) \geq 0$. 

\section{Information Loss in a Multi-Layer (Deep) Learning Setting}
\label{sub_sec_multilayer_IL_analysis}
In Deep Learning, it is standard to have a multilayer (deep) architecture for the encoder and a final soft-max layer for the decoder.   Importantly, this deep 
architecture is a particular case of the encoder-decoder stages  
that Theorem \ref{th_representation} characterizes in the form of a 
class of models. 
Complementing that interpretation, Theorems \ref{th_main_IS_mismatch} and \ref{th_if_projection_aanlogy} can be used to quantity the approximation error (or lack of cross-entropy expressiveness) that is produced 
by this sequential encoder (deep) architecture and see each layer's isolated effect in the forward inference path. 

Following the structure presented in Fig. \ref{fig6},  let us consider an encoder stage composes by $K$-parametric mappings 
$$\left\{\eta_1^{\theta_1}(\cdot), \eta_2^{\theta_2}(\cdot),\ldots, \eta_K^{\theta_K}(\cdot)\right\}$$ 
associated to the latent spaces $\left\{\mathcal{U}_1, \ldots, \mathcal{U}_K\right\}$  with $\eta^{\theta_i}_i: \mathcal{U}_{i-1}  \rightarrow \mathcal{U}_i$ for any $i\in  \left\{1,\ldots, K \right\}$ and $\mathcal{U}_0=\mathcal{X}$. The application of the first $j$-layers ($j\geq 2$) is denoted by $\eta_{1,2,\ldots,j} (\cdot) \equiv \eta^{\theta_j}_j(\cdot) \circ \ldots \circ \eta^{\theta_1}_1(\cdot)$. Consequently, the multilayer encoder is $\eta^{\bar{\theta}} (\cdot) \equiv \eta_{1,2,\ldots,K} (\cdot) = \eta^{\theta_K}_K(\cdot) \circ \ldots \circ \eta^{\theta_1}_1(\cdot)$ with $\bar{\theta} \equiv (\theta_1,\ldots,\theta_K)$. The interpretation of Theorem \ref{th_main_IS_mismatch} as an IP task (in Section \ref{information_proyection}: Theorem \ref{th_if_projection_aanlogy}) 
is insightful here:  every layer represented by $\eta_j^{\theta_j}(\cdot)$ projects the learning task to a smaller hypothesis space (the optimization in (\ref{eq_app_IS_projection_1})) inducing a non-negative approximation error (the mutual information loss in (\ref{eq_infp_2})). 
Formally, we have the following result:  
\begin{theorem}\label{pro_multilayer_info_loss}
	Let us consider a multilayer encoder with $K$-layers of processing $\left\{\eta_1^{\theta_1}(\cdot), \eta_2^{\theta_2}(\cdot),\ldots, \eta_K^{\theta_K}(\cdot)\right\}$. 
	For the IP task in (\ref{eq_app_IS_projection_1}), it follows that 
	\begin{equation*}
	\mathcal{P}(\mathcal{X}\times \mathcal{Y})  \xmapsto[layer1]{\eta_1^{\theta_1}(\cdot)} \mathcal{P}_{\eta^{\theta_1}_1}(\mathcal{X}\times \mathcal{Y})  \xmapsto[layer2]{\eta_2^{\theta_2}(\cdot)}  \mathcal{P}_{\eta^{\theta_2}_2 \circ \eta^{\theta_1}_1}(\mathcal{X}\times \mathcal{Y})  \ldots \xmapsto[layerK]{\eta_K^{\theta_K}(\cdot)} \mathcal{P}_{\eta^{\theta_K}_K \circ \ldots \circ \eta^{\theta_1}_1}(\mathcal{X}\times \mathcal{Y}), 
	\end{equation*}
	where this collection of hypothesis spaces is embedded, i.e, 
	\begin{equation}\label{eq_multilayer_IL_analysis_1b}
		\mathcal{P}_{\eta^{\theta_K}_K \circ \ldots \circ \eta^{\theta_1}_1}(\mathcal{X}\times \mathcal{Y})  \subset \ldots  \subset \mathcal{P}_{\eta^{\theta_2}_2 \circ \eta^{\theta_1}_1 }(\mathcal{X}\times \mathcal{Y})   \subset   \mathcal{P}_{\eta^{\theta_1}_1}(\mathcal{X}\times \mathcal{Y}) \subset \mathcal{P}(\mathcal{X}\times \mathcal{Y}).
	\end{equation}	
	In addition, for any $j\in  \left\{2,\ldots,K\right\}$,\footnote{For $j=1$, we can adopt Theorem \ref{th_if_projection_aanlogy}.}
	\begin{align}\label{eq_multilayer_IL_analysis_2_lyj}
		\min_{\tilde{\mu} \in \mathcal{P}_{\eta^{\theta_j}_j \circ \ldots \circ \eta^{\theta_1}_1}(\mathcal{X} \times \mathcal{Y})} D(\mu_{X,Y} || \tilde{\mu})
				&= \min_{\tilde{\mu} \in \mathcal{P}_{\eta^{\theta_{j-1}}_{j-1} \circ \ldots \circ \eta^{\theta_1}_1}(\mathcal{X} \times \mathcal{Y})} D(\mu_{X,Y} || \tilde{\mu}) + I(U_{j-1};Y|U_{j})
	\end{align}
	with $U_1=\eta^{\theta_1}_1(X)$, $U_2=\eta^{\theta_2}_2(U_1),\ldots$, and $U_j=\eta^{\theta_j}_j(U_{j-1})$.
\end{theorem}
(The proof of this result is presented in Appendix \ref{proof_pro_multilayer_info_loss})

Analysis and interpretation of Theorem \ref{pro_multilayer_info_loss}:
\begin{itemize}
	\item[i)] The multilayer structure of deep-leaning architectures induces a collection of embedded IS hypotheses spaces, see  Eq.(\ref{eq_multilayer_IL_analysis_1b}). 
	\item[ii)] In light of Theorems \ref{th_main_IS_mismatch} and \ref{th_if_projection_aanlogy}, 
	this embedded model structure 
	implies that each layer of the architecture produces a degradation on the best cross-entropy loss that 
	can be achieved with this encoder-decoder design.
	\item[iii)] 
Each individual  layer induces a performance degradation that is measured by the non-negative MIL: $I(U_{j-1};Y|U_{j})\geq 0$ in (\ref{eq_multilayer_IL_analysis_2_lyj}). Consequently, from Lemma \ref{lm_is_encode_d-sparation} and Definition \ref{def_IS_redundant_models}, the isolated effect on the information loss of the $j$-layer is zero if, and only if, $\eta^{\theta_j}_j(\cdot)$ is IS for the model $\mu_{U_{j-1},Y}$, i.e., $\mu_{U_{j-1},Y}\in \mathcal{P}_{\eta^{\theta_j}_j}(\mathcal{U}_k\times \mathcal{Y})$, which means that $U_{j-1}\rightarrow U_{j} \rightarrow Y$.  Importantly, this zero approximation error class $\mathcal{P}_{\eta^{\theta_j}_j}(\mathcal{U}_k\times \mathcal{Y})$ is fully characterized in Theorem \ref{th_representation}.
	\item[iv)] For the first layer, we have from Theorem \ref{th_if_projection_aanlogy} that  
		$\min_{\tilde{\mu} \in \mathcal{P}_{\eta^{\theta_1}_1}(\mathcal{X} \times \mathcal{Y})} D(\mu_{X,Y} || \tilde{\mu})=I(X;Y|U_1)$. Then using (\ref{eq_multilayer_IL_analysis_2_lyj}) and Theorem \ref{th_main_IS_mismatch},
         the accumulated degradation (approximation error) of the first $j$-layers of the inference path (i.e., the encoder $\eta_{1,2,\ldots,j} (\cdot)$) is measured by 
	\begin{align} \label{eq_multilayer_IL_analysis_2_lyj_b}
				I(X;Y|U_j)= I(X;Y|U_1) + I(U_1;Y|U_2) +\ldots + I(U_{j-1};Y|U_{j}).   
	\end{align}				
	\item[v)] Adopting Theorem \ref{pro_multilayer_info_loss} and the additive decomposition in (\ref{eq_multilayer_IL_analysis_2_lyj_b}), we have the following multilayer result from Theorem \ref{th_main_IS_mismatch}: 
	\begin{corollary}
	Every predictive model $v^{\bar{\theta},\phi}_{\tilde{Y}|U_K}(\cdot| \eta_{\bar{\theta}}(\cdot))$ obtained from the deep encoder-decoder introduced in Theorem \ref{pro_multilayer_info_loss} satisfies that 
	\begin{align}\label{eq_multilayer_IL_analysis_1}
		r(v^{\bar{\theta},\phi}_{\tilde{Y}|U_K}(\cdot|\cdot),\mu_{X,Y}) 	\geq &  
		\underbrace{I(X;Y|U_1) + I(U_1;Y|U_2) +\ldots + I(U_{K-1};Y|U_{K})}_{\substack{\emph{layer by layer cross-entropy penalization}}} 
											+ H(Y|X). 
	\end{align}
	\end{corollary}	
	
	\item[vi)] We observe in (\ref{eq_multilayer_IL_analysis_2_lyj_b})  and (\ref{eq_multilayer_IL_analysis_1}), the additive nature of the degradation effect induced by each layer in the architecture. 
	
	\item[vii)] Finally, the optimal projected models (the solutions of Eq.(\ref{eq_multilayer_IL_analysis_2_lyj})) are:  $\tilde{\mu}^1_{X,Y}\equiv \mu_X \cdot \mu_{Y|U_1}\in \mathcal{P}_{\eta^{\theta_1}_1}(\mathcal{X}\times \mathcal{Y})$,  $\tilde{\mu}^2_{X,Y}\equiv \mu_X \cdot \mu_{Y|U_2}\in \mathcal{P}_{\eta^{\theta_2}_2 \circ \eta^{\theta_1}_1}(\mathcal{X}\times \mathcal{Y}), \ldots, $ and  $\tilde{\mu}^2_{X,Y}\equiv \mu_X \cdot \mu_{Y|U_K}\in \mathcal{P}_{\eta^{\theta_K}_K \circ, \ldots, \circ \eta^{\theta_1}_1}(\mathcal{X}\times \mathcal{Y})$. 
\end{itemize}

\begin{remark}
The result presented in this section indicates that every layer of a deep structure introduces a degradation in performance.   At first glance, this suggests that there is no reason to use a deep structure for learning, which contradicts numerous evidence about the great benefit of using these structures in ML. It is worth noting that the presented analysis is limited to the approximation error, which is the lack of cross-entropy expressiveness attributed to using an encoder\footnote{This performance analysis is biased by assuming an oracle (perfect) decoder.}. This analysis does not consider a crucial aspect of a learning task: the estimation error incurred by selecting the decoder from data. To address this lack of perspective, the following section addresses a learning task in which an estimation error appears as an additive degradation effect in our cross-entropy analysis. 
\end{remark}

\section{Cross-Entropy Learning with an Encoder-Decoder Architecture} 
\label{sec_ce_learning}
In representation learning, the encoder and the decoder are learned from data.  We analyze this problem here. 
Let $\mu_{X,Y}$ be the true model (or data-generating distribution) producing $n$ i.i.d. supervised samples $S_n=(X_1,Y_1),\ldots,(X_n,Y_n)$. Using the encoder-decoder structure of Fig.\ref{fig6}, we could represent a learning rule $\xi_n(\cdot)$ 
as a mapping from $S_n\in (\mathcal{X}\times \mathcal{Y})^n$  
to $(\bar{\theta}(S_n), \phi(S_n))\in \Theta \times \Psi$ (the learned parameters), which represents 
for example the parameter of a feed-forward NN.   Following the notation of the previous two Sections, the obtained data-driven predictive distribution 
is $v^{\phi(S_n)}(\cdot|\eta_{\bar{\theta}(S_n)}(\cdot)) \in \mathcal{P}(\mathcal{Y}|\mathcal{X})$ where both the encoder and the decoder are selected from $S_n \in (\mathcal{X}\times \mathcal{Y})^n$. In this context, the hypothesis space (or collection of predictive models) 
 is  $\mathcal{H}_{\Theta,\Psi} \equiv  \left\{ v^{\bar{\theta},\phi}(\cdot|\cdot), \bar{\theta}\in \Theta, \phi \in \Psi \right\} \subset \mathcal{P}(\mathcal{Y}|\mathcal{X})$, where $v^{\bar{\theta},\phi}(\cdot|\cdot)$ is a short-hand for the 
 predictive distribution $v^{\phi}(\cdot|\eta_{\bar{\theta}}(\cdot))$.  Finally, an ML scheme $\Xi=\left\{ \xi_n(\cdot), n\geq 1 \right\}$ is a collection of learning rules for the different data-lengths.

\subsection{Consistency}
In light of Lemma \ref{th_loss_it_bound}, 
we say that a ML scheme is consistent if it achieves the cross-entropy lower bound $H(Y|X)$ of the true unknown model $\mu_{X,Y}$ as $n$ tends to infinity. 
\begin{definition}\label{cross_entropy_consistency}
	A ML scheme $\Xi=\left\{ \xi_n(\cdot), n\geq 1 \right\}$, represented by a collection of learning rule $\xi_n(S_n) = (\bar{\theta}(S_n), \phi(S_n))$ (for each $n\geq 1$), is said to be strongly consistent for the cross-entropy loss, if for any model $\mu_{X,Y}\in \mathcal{P}(\mathcal{X} \times \mathcal{Y})$ 
	\begin{equation} \label{eq_sec_ce_learning_1}
		\lim_{n \rightarrow \infty} r(v^{\phi(S_n)}(\cdot|\eta_{\bar{\theta}(S_n)}(\cdot)),\mu_{X,Y}) = H(Y|X),
	\end{equation}
	where $(X_i,Y_i)_{i\geq 1}$ are i.i.d. samples from $\mu_{X,Y}$ and the convergence in (\ref{eq_sec_ce_learning_1}) is a.s. w.r.t. the  process distribution of $(X_i,Y_i)_{i\geq 1}$. 
\end{definition}
Achieving consistency means achieving the best performance for the inference task, which means learning in some way the predictive component of $\mu_{X,Y}$ as presented in the following result (Theorem \ref{th_consistency}).

\subsection{The Result}
\label{sub_sec_ce_learning_main_result}
It is relevant to understand what it means 
to achieve the optimal performance $H(Y|X)$ and the concrete requirements, if any, we need to ask the encoder and the decoder 
of an ML scheme of the form  $v^{\phi(S_n)}(\cdot|\eta_{\bar{\theta}(S_n)}(\cdot))$ to meet the condition stated in Definition \ref{cross_entropy_consistency}. We answer these two questions in the following result: 
\begin{theorem} 
\label{th_consistency}
Let us consider a ML encoder-decoder scheme $\Xi=\left\{ \xi_n(\cdot), n\geq 1 \right\}$ determined by the rules $\xi(S_n)=(\bar{\theta}(S_n), \phi(S_n))$ for any $n\geq 1$. The scheme is consistent (Def. \ref{cross_entropy_consistency}) if, and only if,  for any model $\mu_{X,Y}\in \mathcal{P}(\mathcal{X}\times \mathcal{Y})$
\begin{equation} \label{eq_sec_ce_learning_2_pre} 
	\lim_{n \rightarrow \infty} D(\mu_{Y|X}(\cdot |\cdot)|| v^{\phi(S_n)}(\cdot|\eta_{\bar{\theta}(S_n)}(\cdot)) |\mu_X)=0,
\end{equation}
where 
\begin{equation}\label{eq_sec_ce_learning_2} 
	D(\mu_{Y|X}(\cdot |\cdot)|| v^{\phi(S_n)}(\cdot|\eta_{\bar{\theta}(S_n)}(\cdot)) |\mu_X) \equiv \mathbb{E}_{X\sim \mu_X}   \left\{ D(\mu_{Y|X}(\cdot | X) || v^{\phi(S_n)}(\cdot|\eta_{\bar{\theta}(S_n)}(X))) \right\}
\end{equation}
is the average (w.r.t. to $X\sim \mu_X$) KL divergence \cite{kullback1958,cover_2006}\footnote{$D(p||q) \equiv \sum_{y\in \mathcal{Y}} \log p(y) \frac{p(y)}{q(y)}$ when $p\ll q$ and,  otherwise, $D(p||q) \equiv \infty$ \cite{kullback1958}.} between the true predictive model $\mu_{Y|X}(\cdot | X) \in \mathcal{P}(\mathcal{Y})$ and the learned model $v^{\phi(S_n)}(\cdot|\eta_{\bar{\theta}(S_n)}(X)) \in \mathcal{P}(\mathcal{Y})$, and the convergence in (\ref{eq_sec_ce_learning_2_pre}) is a.s. w.r.t. the process distribution of $(X_i,Y_i)_{i\geq 1}$.\\

Looking at the isolated role of the encoder and the decoder in $\Xi$, the condition in (\ref{eq_sec_ce_learning_2_pre}) is met if, and only if,
 \begin{itemize}
 	\item[{\bf i)}] $\lim_{n \rightarrow \infty} I(X;Y| \eta_{\bar{\theta}(S_n)}(X))=0$  and 
	\item[{\bf ii)}] $\lim_{n \rightarrow \infty} D(\mu_{Y|U_n}(\cdot |\cdot)|| v^{\phi(S_n)}(\cdot|\cdot) |\mu_{U_n})=0$
\end{itemize}
where $U_n \equiv \eta_{\bar{\theta}(S_n)}(X)$ is the learned data-driven representation,  $\mu_{U_n}$ denotes the induced true distribution of $U_n$ and 
\begin{equation}\label{eq_sec_ce_learning_2b} 
	D(\mu_{Y|U_n}(\cdot |\cdot)|| v^{\phi(S_n)}(\cdot|\cdot) |\mu_{U_n})  \equiv \mathbb{E}_{U_n\sim \mu_{U_n}}  \left\{ D(\mu_{Y|U_n}(\cdot | U_n) || v^{\phi(S_n)}(\cdot|U_n)) \right\}. 
\end{equation}
The convergences in {\bf i)} and {\bf ii)} are almost surely w.r.t. the process distribution of $(X_i,Y_i)_{i\geq 1}$.
\end{theorem}
(The proof of this result is presented in Appendix \ref{proof_th_consistency})

Interpretations of Theorem \ref{th_consistency}:
\begin{enumerate}
	\item First, the result establishes a necessary and sufficient condition on a ML scheme 
	to meet optimality in the cross-entropy sense. 
	The optimality condition in (\ref{eq_sec_ce_learning_2_pre}) means 
	that the learned model $v^{\phi(S_n)}(\cdot|\eta_{\bar{\theta}(S_n)}(\cdot))$ matches asymptotically (in the KLD sense) the true predictive distribution. 
	Then, achieving $H(Y|X)$ means no less than learning the true predictive model $\mu_{Y|X}(\cdot|\cdot)$. 
	
	\item Importantly, the proof of Theorem \ref{th_consistency} shows that 
	$D(\mu_{Y|X}(\cdot |\cdot)|| v^{\phi(S_n)}(\cdot|\eta_{\bar{\theta}(S_n)}(\cdot)) |\mu_X)$ quantifies the performance gap (or lack of optimality) w.r.t. $H(Y|X)$ of the learned (data-driven) model  $v^{\phi(S_n)}(\cdot|\eta_{\bar{\theta}(S_n)}(\cdot))$. Then, the discrepancy in the KLD sense in (\ref{eq_sec_ce_learning_2}) 
	is the performance overhead.
	
	\item Looking at the encoder-decoder structure of the scheme,  Theorem \ref{th_consistency} isolates in two additive terms, see Eq.(\ref{eq_sec_ce_learning_3}) below, the expressive effect of (or structured biased induced by) the encoder, by the MI loss  $I(X;Y| \eta_{\bar{\theta}(S_n)}(X))\geq 0$, and the expressive effect of (or bias induced by) the stochastic decoder $(f_{\phi(S_n)}(W, u))_{u\in \mathcal{U}} \sim (v^{\phi(S_n))}(\cdot|u))_{u\in \mathcal{U}} \in \mathcal{P}(\mathcal{Y}|\mathcal{U})$ by the KLD in (\ref{eq_sec_ce_learning_2b}). 
	Importantly, 
	we have that for 
	any finite $n>0$: 
	\begin{align}\label{eq_sec_ce_learning_3} 
		 &r(v^{\phi(S_n)}(\cdot|\eta_{\bar{\theta}(S_n)}(\cdot)),\mu_{X,Y}) =\nonumber\\
		 &\underbrace{I(X;Y| \eta_{\bar{\theta}(S_n)}(X))}_{\text{Encoder Effect}\geq 0 } + \underbrace{D(\mu_{Y|U_n}(\cdot |\cdot)|| v^{\phi(S_n)}(\cdot|\cdot) |\mu_{U_n})}_{\text{Decoder Effect} \geq 0} + H(Y|X).
	\end{align}
	
	\item In (\ref{eq_sec_ce_learning_3}),  
	it is worth noting that $I(X;Y| \eta_{\bar{\theta}(S_n)}(X)) \geq 0$ 
	is 
	the approximation error of assuming that $\mu_{X,Y}$ belongs to the data-driven class $\mathcal{P}_{\eta_{\bar{\theta}}(S_n)}(\mathcal{X}\times \mathcal{Y})$ (from Theorem \ref{th_main_IS_mismatch}).  On the other hand,  
	$D(\mu_{Y|U_n}(\cdot |\cdot)|| v^{\phi(S_n)}(\cdot|\cdot) |\mu_{U_n}) \geq 0$ can be seen as a variance (or estimation error): it measures the average discrepancy between the data-driven prediction $S_n \rightarrow f_{\phi(S_n)}(W,\cdot) \sim (v^{ \phi(S_n)}(\cdot| u))_{u\in \mathcal{U}} \in \mathcal{P}(\mathcal{Y}|\mathcal{U})$ and the true model $\mu_{Y|U_{n}}(\cdot|\cdot)$ 
 in the transform (projected by the encoder) 
	space $\mathcal{P}(\mathcal{Y}|\mathcal{U})$. Indeed in the proof of Theorem \ref{th_consistency}, we derive the following additive decomposition: 
	\begin{align}\label{eq_sec_ce_learning_3b} 
	D(\mu_{Y|X}(\cdot |\cdot)|| v^{\phi(S_n)}(\cdot|\eta_{\bar{\theta}(S_n)}(\cdot)) |\mu_X)=I(X;Y| \eta_{\bar{\theta}(S_n)}(X)) + D(\mu_{Y|U_n}(\cdot |\cdot)|| v^{\phi(S_n)}(\cdot|\cdot) |\mu_{U_n}).
	\end{align}	
	
	\item Condition {\bf i)} represents an expressiveness requirement for the data-driven encoders $\left\{\eta_{\bar{\theta}(S_n)}(\cdot), n\geq 1 \right\}$. The encoders need to capture all the information that $X$ has about $Y$ in the MI sense as $n$ tends to infinity. In other words, the collection of representations $\left\{ \eta_{\bar{\theta}(S_n)}(\cdot), n\geq 1 \right\}$ needs to be asymptotically IS for $\mu_{X,Y}$  (Definition \ref{def_is}). This IS asymptotic criterion over a collection of representations was introduced and studied in \cite{silva_2022_aistat} from the angle of studying encoder expressiveness in the classical probability of error sense. In contrast, and  
	from the angle of achieving optimality in the cross-entropy sense, Theorem \ref{th_consistency} states that this asymptotic IS condition is necessary but not sufficient for consistency.
	
	\item Condition {\bf ii)} represents an expressiveness requirement for the stochastic (soft) decoder, $\left\{ v^{ \phi(S_n)}(\cdot| u))_{u\in \mathcal{U}}, n\geq 1 \right\}$. It expresses the necessity to approximate the true projected  
	model\footnote{This observation is from the fact that the KLD  is zero iff the compared distributions are the same (in total variations).} $\mu_{Y|U_n}(\cdot|\cdot)$, which is a moving target with $n$.
\end{enumerate}

In summary,  an ML scheme is strongly consistent if the learning rules 
have the ability to learn an IS representation for $\mu_{X,Y}$ with the encoders (in the sense of {\bf i)} in Th. \ref{th_consistency}) and 
the true predictive distribution in the induced latent domain with the decoder (in the KL sense of {\bf ii)} in Th. \ref{th_consistency}).

\subsection{Encoder Expressiveness in Cross-Entropy Learning}  
\label{IP_view_in_learning}
Theorem \ref{th_main_IS_mismatch} shows that if $\mu_{X,Y}\in \mathcal{P}_\eta(\mathcal{X} \times \mathcal{Y})$ an $\eta(\cdot)$-decoder architecture is expressive (assuming Def.\ref{def_prob_expressiviness}).  Then, we have a model 
explanation (IS structure) to justify the adoption an encoder-decoder ML design. We can derive a similar interpretation for the encoder-decoder learning scheme $\Xi=\left\{ \xi_n(\cdot), n\geq 1 \right\}$ in Theorem \ref{th_consistency}. 
We note that $\Xi$ has a collection of encoders $\left\{\eta_{\bar{\theta}}(\cdot): \theta \in \Theta \right\}$ equipped with their respective class of IS models $ \left\{ \mathcal{P}_{\eta_\theta}(\mathcal{X} \times \mathcal{Y}):  \bar{\theta} \in \Theta \right\}$ (from Theorem \ref{th_representation}). If  
$\Xi$
is consistent, using Theorem \ref{th_consistency} and the IP analogy (in Theorem \ref{th_if_projection_aanlogy}), we have that 
\begin{equation} \label{eq_IP_view_in_learning_1}
	\min_{\tilde{\mu} \in \mathcal{P}_{\eta_{\bar{\theta}}(S_n)}(\mathcal{X} \times \mathcal{Y})} D(\mu_{X,Y} || \tilde{\mu})=  I(X;Y| \eta_{\bar{\theta}(S_n)}(X)) \rightarrow 0. 
\end{equation}

This asymptotic IS condition implies that
\begin{equation} \label{eq_IP_view_in_learning_1b}
	\max_{\mu_{X,Y} \in \mathcal{P}(\mathcal{X} \times \mathcal{Y})}\min_{\tilde{\mu} \in \bigcup_{\bar{\theta} \in \Theta} \mathcal{P}_{\eta_{\bar{\theta}}}(\mathcal{X} \times \mathcal{Y})} D(\mu_{X,Y} || \tilde{\mu})=0.  
\end{equation}
In other words, we have that the space $\bigcup_{\bar{\theta} \in \Theta} \mathcal{P}_{\eta_{\bar{\theta}}}(\mathcal{X} \times \mathcal{Y})$ is expressive in the sense that it approximate arbitrary closely any distribution in $\mathcal{P}(\mathcal{X} \times \mathcal{Y})$ in the KL sense. 
It is important to remember that the expressiveness requirement in (\ref{eq_IP_view_in_learning_1}) is necessary but not sufficient for a scheme to be consistent.  
Conversely, if 
\begin{equation} \label{eq_IP_view_in_learning_2}
\delta(\Theta) \equiv \max_{\mu_{X,Y} \in \mathcal{P}(\mathcal{X} \times \mathcal{Y})}\min_{\tilde{\mu} \in \bigcup_{\bar{\theta} \in \Theta} \mathcal{P}_{\eta_{\bar{\theta}}}(\mathcal{X} \times \mathcal{Y})} D(\mu_{X,Y} || \tilde{\mu})>0,
\end{equation}
$\Xi=\left\{ \xi_n(\cdot), n\geq 1 \right\}$ is not consistent independent of the way the learning rules select the parameters in $\Theta \times \Psi$ from data. 
More precisely, 
for an arbitrary small $\epsilon>0$
there is a model $\mu_{X,Y} \in \mathcal{P}(\mathcal{X} \times \mathcal{Y})$ and $N>0$, such that  $\forall n\geq N$ 
\begin{equation} \label{eq_IP_view_in_learning_3}
	r(v^{\phi(S_n)}(\cdot|\eta_{\bar{\theta}(S_n)}(\cdot)),\mu_{X,Y}) >  {\delta(\Theta) }+ H(Y|X) - \underbrace{\epsilon}_{\approx 0}.
\end{equation}
From (\ref{eq_IP_view_in_learning_3}), the expression in (\ref{eq_IP_view_in_learning_2}) is a structural performance degradation attributed to the lack of expressiveness of $\bigcup_{\theta \in \Theta} \mathcal{P}_{\eta_\theta}(\mathcal{X} \times \mathcal{Y})$.  
This 
max-min model approximation error is independent of how good the learning rules in $\Xi$ operates to select the (learned) parameters in $\Theta \times \Psi$ from the data (empirical) process $(S_n)_{n\geq 1}$.

\subsection{Digital Encoders are Expressive}
\label{sub_sec_expresive_VQ}
Here we show that vector quantizers (VQs) are universally expressive as an encoder strategy for ML.  This is shown from the perspective of achieving an arbitrary small IL for any model in $\mathcal{P}(\mathcal{X} \times \mathcal{Y})$. Let $\mathcal{Q}(\mathcal{X})$ denote the family of finite-size measurable partitions of $\mathcal{X}$. Then, any $\pi\in \mathcal{Q}(\mathcal{X})$ is an indexed measurable partition of the form $\pi=\left\{A_i, i\in \mathcal{I} \right\}\subset \mathcal{B}(\mathcal{X})$  with $\left| \mathcal{I} \right| < \infty$, which is equipped with its respective digital encoder (or VQ) $\eta_\pi(\cdot)$ in (\ref{eq_ps_11}). The family of finite-size VQs is $\left\{ \eta_\pi(\cdot), \pi\in \mathcal{Q}(\mathcal{X})  \right\}$. 
In the setting of Theorem \ref{th_consistency}, if the encoders are selected within the mentioned class of VQs, the following result shows this class is expressive offering the possibility of achieving a vanishing IL: i.e., meeting the max-min KL vanishing 
condition in (\ref{eq_IP_view_in_learning_1b}). 
\begin{lemma}
\label{lm:expresive_VQ}
	\begin{equation} \label{eq_expresive_VQ_1}
	\max_{\mu_{X,Y} \in \mathcal{P}(\mathcal{X} \times \mathcal{Y})} \min_{\tilde{\mu} \in \bigcup_{\pi \in \mathcal{Q}(\mathcal{X})} \mathcal{P}_{\pi}(\mathcal{X} \times \mathcal{Y})} D(\mu_{X,Y} || \tilde{\mu})=0,  
\end{equation}
where $\mathcal{P}_{\pi}(\mathcal{X} \times \mathcal{Y})$ is the digital IS class introduced in (\ref{eq_ps_12}) and characterized in Section \ref{sub_sec_digital}.\\ 
(The proof of Lemma \ref{lm:expresive_VQ} is presented in Appendix \ref{proof_lm:expresive_VQ})
\end{lemma}
Lemma \ref{lm:expresive_VQ} justifies the adoption of VQ as an expressive alternative to optimally learn a prediction task in the cross-entropy sense. There are many ML algorithms that do use VQ as an encoder strategy. On this, we can highlight the deterministic IB method \cite{strouse_2017}, and the recently introduced lossy compression for lossless prediction method \cite{Dubois_2021}, both information-driven strategies that learn digital encoders (VQ)  from data. In Section \ref{sec_IB}, we discuss further the appropriateness of the IB learning principle.

\section{IS Learning and Information Bottleneck} 
\label{sec_IB}
The information bottleneck (IB) problem was introduced in \cite{tishby_1999} as a particular case of the celebrated rate-distortion problem \cite{cover_2006,berger_1971}. Recently, it has been adopted in representation learning as a principle to learn expressive encoders 
from data \cite{tishby_2015,alemi_2017,strouse_2017,amjad_2019} within the encoder-decoder framework studied in this work. In light of Theorem \ref{th_consistency}, we revisit the IB principle to evaluate its ability to learn good encoders for cross-entropy learning. 

Given a model $\mu_{X,Y}$, the IB method solves the following problem:\footnote{Different versions of the IB problem can be considered depending on the selection of $\mathcal{U}$. For simplicity, we just focus on $\mathcal{U}=\mathbb{R}$.} 
\begin{equation} \label{eq_IB_1}
	\max_{P_{U|X} \in \mathcal{P}(\mathcal{U}|\mathcal{X}), st. I(X;U) \leq B}  I(U;Y),  
\end{equation}
where $\mathcal{P}(\mathcal{U}|\mathcal{X})$ is the collection of conditional probabilities from $\mathcal{X}$
to the latent space $\mathcal{U}=\mathbb{R}$ and the MIs in (\ref{eq_IB_1}) are obtained from the 
random object $(Y,X,U)\sim \mu_{Y,X}\cdot P_{U|X}$.  The IB problem in (\ref{eq_IB_1}) 
finds the conditional probabilities (or soft-encoders) with the best tradeoff between information $I(U;Y)$  and compression $I(X;U)$. The condition $I(X;U) \leq B$ is the IB restriction. $I(X;U)$ has been interpreted in ML (from the related lossy source coding task \cite{berger_1971}) as the number of bits that $U$ retains about $X$. In the RL setting studied in this work, we concentrate on deterministic encoders $\eta: \mathcal{X} \rightarrow \mathbb{R}$, and, consequently, on the deterministic version of the IB problem: 
\begin{equation} \label{eq_IB_2}
	\max_{\eta(\cdot) \mathcal{X} \rightarrow \mathbb{R}, st. H(\eta(X)) \leq B}  I(\eta(X);Y).  
\end{equation}
Here, the IB restriction $H(\eta(X)) \leq B$ is very strict because $U=\eta(X)$ needs to have a finite Shannon entropy (number of bits). 
Importantly, a way to induce finite-entropy latent variables is by the family of VQs presented in Section \ref{sub_sec_expresive_VQ}.\footnote{For any finite-size  VQ  $\eta_\pi(\cdot)$, we have that $H(\eta_\pi(X)) \leq \log \left| \pi  \right| < \infty$.} 
It is important to notice that by design, the IB method minimizes the IL $I(X;Y|\eta(X))$ over a family of finite entropy encoders and, consequently, in theory the IB method might offer the capacity to learn expressive finite-entropy encoders that make $I(X;Y|\eta(X)) \approx 0$ as the bandwidth (or compression) bound $B$ grows. Using the expressive quality of VQs in Lemma \ref{lm:expresive_VQ}, the next result shows that the IB method in (\ref{eq_IB_2}) selects encoders with the capacity to meet the IS asymptotic condition stated in Theorem \ref{th_consistency} ({\bf i)}.
\begin{lemma}
\label{lm:IB_expressive}
	Let $\eta^B(\cdot)$ denote the solution of the IB problem in (\ref{eq_IB_2}) for the IB restriction $B>0$.
	For any model $\mu_{X,Y}$ and $\epsilon>0$,  there exists $B(\epsilon, \mu_{X,Y})>0$ such that 
	 $\forall B\geq B(\epsilon, \mu_{X,Y})$, $I(X;Y|\eta^B(X)) < \epsilon$.\\    
	 (The proof of this result is presented in Appendix \ref{proof_lm:IB_expressive})
\end{lemma}

This result expresses that no matter how much self-information $X$ has (potentially an infinite number of bits \cite{renyi_1959}), the IB method offers a principle 
to learn a finite bit (lossy) description of $X$ that retains an arbitrarily large proportion of the MI that $X$ has about $Y$.   Indeed,  Lemma \ref{lm:IB_expressive} implies that $\lim_{B \longrightarrow \infty} I(X;Y|\eta^B(X))=0$ for any model $\mu_{X,Y}$, which is 
precisely the 
condition for the encoders stated in Theorem \ref{th_consistency} ({\bf i)}.
Therefore,  the IB method (and Lemma \ref{lm:IB_expressive}) confirms that digital compression of $X$ 
with an info-max criterion offers model-dependent expressive representations 
for ML in the cross-entropy sense. 

\section{A Controlled Empirical Study}
\label{sec_numerical}
We conclude this paper with a controlled numerical analysis to evaluate the expressiveness of ML schemes with an encoder-decoder structure. We use this paper's theory and mathematical formalization to guide the interpretation. We look at two important practical aspects: first evaluate the power of a ML architecture in its capacity to achieve optimal cross-entropy performance  (from Def. \ref{cross_entropy_consistency} and Th. \ref{th_consistency}), and second, analyze numerically the performance effect of the encoder and decoder stages, studied in Section \ref{sec_mismatch}, using  the information loss and KLD gaps predicted in Theorems \ref{th_main_IS_mismatch} and \ref{th_consistency}, respectively.

\subsection{Models}
\label{sec_numerical_model}
Because of the  mixed  discrete-continuous setting $\mathcal{X}=\mathbb{R}^d$ and $\mathcal{Y}=[M]$, we have designed a rich family of models $\mu_{X,Y}$  for which
  $I(Y;X)$ can be computed in closed-form. We consider encoders $\left\{\eta_1(\cdot),..,\eta_K(\cdot) \right\}$ that are feature selectors, as   presented in Section \ref{sub_sec_sparse_canonical}. The proposed model $\mu_{X,Y}$ construction along with these encoders  provide analytical expressions for $I(Y; \eta_i(X))$ and $I(Y;X|\eta_i(X))$  
in (\ref{eq_sec_mismatch_0}).\footnote{The model construction and the analytical expressions for  $I(Y;X)$ and $\left\{ I(Y; \eta_i(X)), i=1,\ldots,K \right\}$ are presented in Appendix \ref{supp:emp-study}.} 
In particular, we use a reference model $\mu_{X,Y}$ with $d=15$ and $M=3$. 
We produce two other reference models in the same space $\mathbb{R}^d \times  \left\{1,..,M \right\}$ by masking specific coordinates of $X$.  We obtain $\mu_{\tilde{X},Y}$ by masking the coordinate 1 of $X$ to produce $\tilde{X}$ and $\mu_{\bar{X},Y}$  by masking the coordinates 1,3 and 5 of $X$ to produce $\bar{X}$. Therefore, under these three models,  it follows that $I(X;Y)> I(\tilde{X},Y) > I(\bar{X},Y)=0$, which offers three distinctive discrimination scenarios (reflected in the bound in Th.\ref{th_main_IS_mismatch}) and learning difficulties.  In addition, $\mu_{X,Y}$ was designed with a low dimensional IS sparse structure (see Corollary \ref{cor_sparse_class}) in the sense that the 5D coordinate projector $\eta_{1,2,3,4,5}:\mathbb{R}^{15} \rightarrow \mathbb{R}^5$ (introduced in Section \ref{sub_sec_sparse_canonical})  is IS for $\mu_{X,Y}$ (Def. \ref{def_IS_redundant_models}). Consequently for the  three models, we have that $I(X;Y|\eta_{1,2,3,4,5}(X)) = I(\tilde{X};Y|\eta_{1,2,3,4,5}(\tilde{X})) = I(\bar{X};Y|\eta_{1,2,3,4,5}(\bar{X})) =0$, i.e., $\eta_{1,2,3,4,5}(\cdot)$ is IS.  We will use  $\eta_{1,2,3,4,5}(\cdot)$ to represent a learning scenario where prior structural IS  knowledge is available. 

\subsection{Performance Metric}
For each of the mentioned models, for example $\mu_{X,Y}$, we produce two set of i.i.d. samples: the training set $S_n=(X_1,Y_1),\ldots,(X_n,Y_n)$ and validation set $\tilde{S}_m=(\tilde{X}_1,\tilde{Y}_1),\ldots,(\tilde{X}_m,\tilde{Y}_m)$.  We also have a collection of ML schemes $\Xi^1,\ldots,\Xi^K$ (see Section \ref{sec_ce_learning}) where $\Xi^i=\left\{ \xi^i_n(\cdot), n\geq 1 \right\}$ and  $\xi^i_n(\cdot)$ is equipped with an encoder-decoder structure. In this context, given the training set $S_n$ (of length  $n$), the rule of  $\Xi^i$ maps  $S_n$ to $(\bar{\theta}_n(S_n),\phi_n(S_n))$ and from these parameters we have the induced data-driven  predictive model $v^{\bar{\theta}_n(S_n),\phi_n(S_n)}(\cdot|\cdot)= v^{\phi_n(S_n)}(\cdot|\eta_{\bar{\theta}_n(S_n)}(\cdot)) \in \mathcal{P}(\mathcal{Y}|\mathcal{X})$, which is the output of the learning process. 

For performance evaluation, we use the validation set $\tilde{S}_m$. In particular, given an induced data-driven  model $v^{\bar{\theta}_n(S_n),\phi_n(S_n)}(\cdot|\cdot)$ its (empirical) cross-entropy risk is
\begin{equation}\label{eq_sec_numerical_1}
	 \hat{r}(v^{\bar{\theta}_n(S_n),\phi_n(S_n)}(\cdot|\cdot),\tilde{S}_m) = - \frac{1}{m} \sum^m_{i=1} \log 
	 v^{\phi_n(S_n)}(\tilde{Y}_i |\eta_{\bar{\theta}_n(S_n)}(\tilde{X}_i)).
\end{equation}
By the law of large numbers, as $m$ becomes large,  $ \hat{r}(v^{\bar{\theta}_n(S_n),\phi_n(S_n)}(\cdot|\cdot),\tilde{S}_m) $ tends (almost surely w.r.t. the process distribution of $(\tilde{S}_m)_{m\geq 1}\sim \mu_{X,Y}$) to the true average risk: 
\begin{equation}\label{eq_sec_numerical_2}
	 r(v^{\bar{\theta}_n(S_n),\phi_n(S_n)}(\cdot|\cdot),\mu_{X,Y}) = \mathbb{E}_{(X,Y)\sim \mu_{X,Y}}  \left\{- \log 
	 v^{\phi_n(S_n)}({Y} |\eta_{\bar{\theta}_n(S_n)}({X}))) \right\},
\end{equation}
which is the performance indicator in our analysis. 
Then, we are interested in analyzing the dynamic of the performance gap $r(v^{\bar{\theta}_n(S_n),\phi_n(S_n)}(\cdot|\cdot),\mu_{X,Y}) -  H(Y|X)$ as  $H(Y|X)$ is available  in our controlled
setting. 
From Theorem \ref{th_consistency},  
this gap (or performance overhead) has two distinctive non-zero information components:  
\begin{equation}\label{eq_sec_numerical_3}
	\underbrace{I(X;Y| \eta_{\bar{\theta}_n(S_n)}({X}))}_{\text{encoder bias: } \bar{\theta}_n(S_n)} +  \underbrace{D(\mu_{Y|U_n}(\cdot |\cdot)|| v^{\phi(S_n)}(\cdot|\cdot) |\mu_{U_n})}_{\text{decoder error: }\phi_n(S_n)}.
\end{equation}

\subsection{MLP Architectures}
Concerning the ML scheme $\Xi=\left\{ \xi_n(\cdot), n\geq 1 \right\}$,  we choose three multi-layer perceptrons (MLP) with a different number of parameters and a ReLU activation function: $\text{MLP}32$ is an MLP with one hidden layer of width 32, $\text{MLP}256$ is an MLP with two hidden layers of with 256 and finally  $\text{MLP}1024$ is an MLP with two hidden layers of width 1024.  For training, the cross-entropy loss is used with stochastic gradient descent (SGD). 
The practical details of the training process used in each case are presented in Appendix \ref{supp:emp-study}. 

\begin{figure*}
  \centering
   \includegraphics[width=0.99\textwidth]{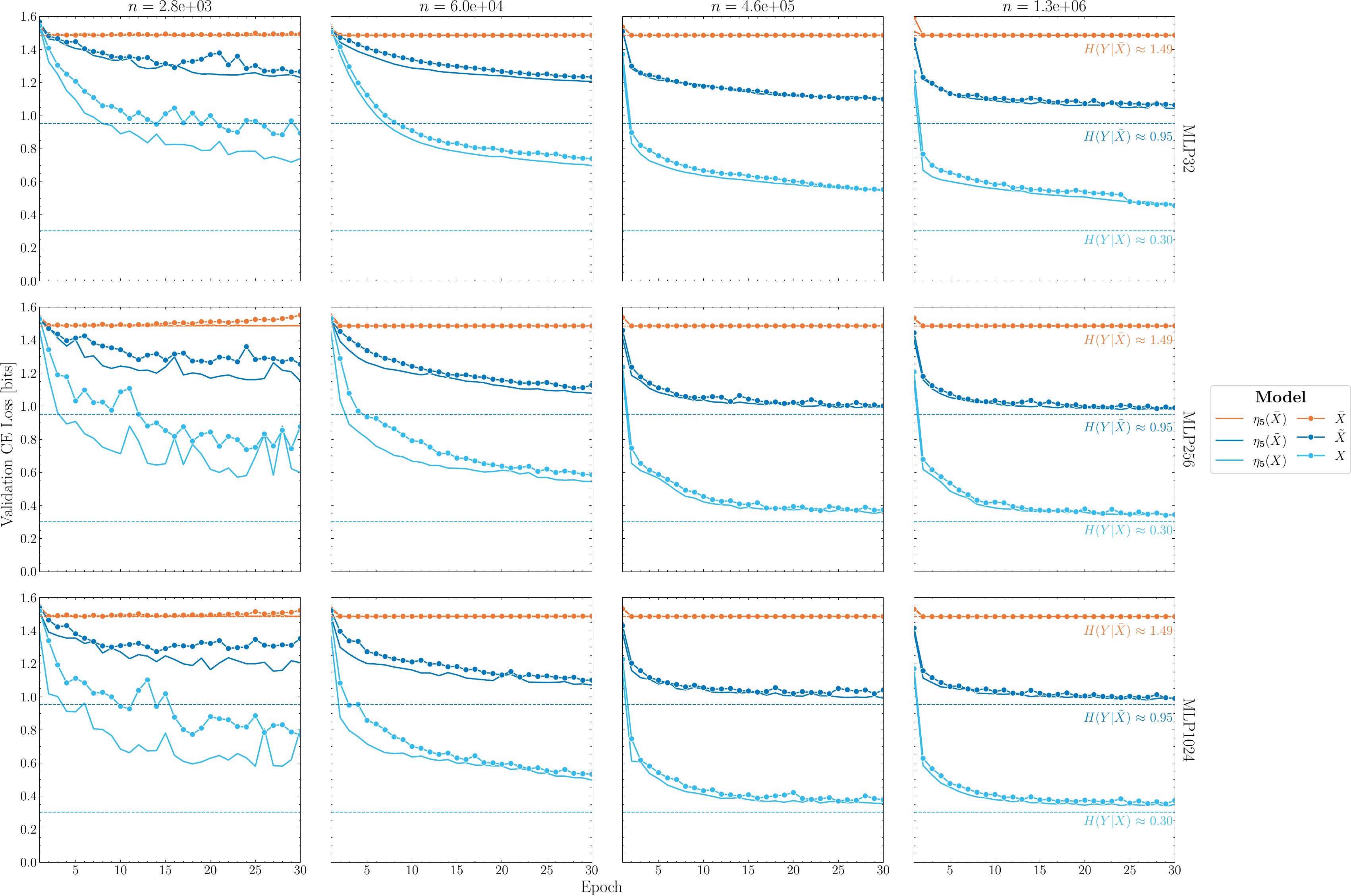}
  \caption{Cross-entropy losses curves per-epoch for different MLP schemes, models and training data-lengths ($n$). The curves for $\text{MLP}32$, $\text{MLP}256$ and $\text{MLP}1024$ are presented in the first, second and third rows, respectively. The horizontal lines present the cross-entropy lower bounds of Theorem \ref{th_main_IS_mismatch} where $H(Y|X)=0.303532$, $H(Y|\tilde{X})=0.952762$ and $H(Y|\bar{X})=H(Y)=1.485475$. In the right caption, $\eta_5(\cdot)$ is a short-hand for $\eta_{1,2,3,4,5}(\cdot)$.}
  \label{fig2}
\end{figure*}

\subsection{Performance Analyses}
Given the universal functional expressiveness of multilayer networks \cite{hornik_1989}, it is interesting to evaluate its capacity to approximate optimal performance in the sense of Def. \ref{cross_entropy_consistency} (learnability or consistency) and, in the process, how the dynamics of the gap in (\ref{eq_sec_numerical_3}) depends on the number of parameters of the scheme, the width of the hidden layer,  the underlying distribution, the number of training examples, the number of training epoch, etc.
In particular, for each scheme ($\Xi^1=\text{MLP}32$,  $\Xi^2=\text{MLP}256$, $\Xi^3=\text{MLP}1024$), model ($\mu_{X,Y}$, $\mu_{\tilde{X},Y}$ and $\mu_{\bar{X},Y}$), training epoch ($k\in \left\{1,\ldots, 20 \right\}$),  and data-lengths ($n \in \left\{2.78e^{+3}, 2.15e^{+4}, 5.99e^{+4}, 4.64e^{+5}, 1.29e^{+6} \right\}$), the loss $r(v^{\bar{\theta}^i_n(S_n),\phi^i_n(S_n)}(\cdot|\cdot),\mu_{X,Y})$ is estimated from (\ref{eq_sec_numerical_1}) and compared with $H(Y|X)$ to characterize the encoder-decoder gap in (\ref{eq_sec_numerical_3}).  
In addition, we compute the same performance metrics when using the IS pre-encoder $\eta_{1,2,3,4,5}(\cdot)$ (that projects the learning task to a smaller dimension). The idea here is to observe, if any, the benefit of the IS structural knowledge when learning the true predictive model (i.e., meeting (\ref{eq_sec_ce_learning_1})).  In all these results, we consider $m=800,000$ to compute (\ref{eq_sec_numerical_1}), a sufficient number of validation samples to have a precise estimation of the true loss in  (\ref{eq_sec_numerical_2}).

Figure \ref{fig2} presents these performance curves (function of $k$) organized by columns (associated with a fixed $n$) and rows (associated with a fixed MLP architecture).  In each sub-figure, we observe six loss curves (function of $k$): one for each of the three models ($\mu_{X,Y}$, $\mu_{\tilde{X},Y}$ and $\mu_{\bar{X},Y}$) with and without the use of the IS pre-encoder $\eta_{1,2,3,4,5}(\cdot)$. In addition, we include the cross-entropy lower bound for each of the three models (the dashed horizontal lines).  On the analysis, we can say the following:
\begin{itemize}
	\item {\bf Discrimination $\&$ Performance Dynamics:} Each of the three models (learning scenario) offers learning curves with distinctive performance dynamics. The most discriminative model $\mu_{X,Y}$ (with the smallest conditional entropy) is the most challenging to learn from data. As this case has the smallest performance bound, the task requires more complex schemes, more data ($n$), and a higher number of epochs ($k$) to be able to meet a performance that is closer to $H(Y|X)$. This is in clear contrast with the non-discriminative model $\mu_{\bar{X},Y}$,  $H(Y)=H(Y|\bar{X})$, where all performance curves meet optimality after few training epoch independent of the sample size and the complexity of the network. There is  also a clear difference on the curves of $\mu_{X,Y}$ and the second most discriminative model $\mu_{\tilde{X},Y}$. Here, the performance gaps  in (\ref{eq_sec_numerical_3}) of $\mu_{\tilde{X},Y}$ are smaller than the respective gap of $\mu_{X,Y}$  when compering the same scheme, $k$ and $n$. This evidence suggests that $H(Y|X)$ is an informative indicator of the difficulty of learning the true predictive distribution with an NN.	
	
	\item {\bf More Parameters  is Better:} Complementing the previous point, we  observe that more complex schemes (in the number of parameters) have better learning dynamics overall. This is  particularly clear in the sample size regime when $n\in \left\{2.78e^{+3}, 1.29e^{+6} \right\}$ for the two discriminative models: $\mu_{X,Y}$ and  $\mu_{\tilde{X},Y}$.  This finding is consistent with some evidence in the literature pointing to the surprising performance of over-parametrized NN architecture. The difference in the performance curves is more prominent when learning the most discriminative model ($\mu_{X,Y}$). In contrast, on the trivial non-discriminative model (where the observations are irrelevant), there are no differences when increasing the parameter size of the scheme. This evidence indicates that the potential gains in cross-entropy attributed to adopting more complex NN architectures are proportional to the predictability (in number of bits) of $Y$ given $X$.
	
	\item {\bf IS Learnability and IL:} Importantly, our loss curves show that MLP can meet results near the optimal performance bounds. Hence,  the well-known functional approximation quality of MLP does translate in a capacity to learn the true predictive distribution in the KL sense, which Theorem \ref{th_consistency} shows is a necessary and sufficient condition to achieve $H(Y|X)$. 
	On the other hand, we also observe a non-vanishing performance gap for the single-layer architecture ($\text{MLP}32$).  This discrepancy is not mitigated by increasing the number of epochs ($k$) or training size ($n$). Then, we observe a structural bottleneck in this scenario.
	In this,  we recognized from Theorem \ref{th_consistency} two sources of discrepancy: a non-expressive encoder $I(X;Y| \eta_{\bar{\theta}_n(S_n)}({X}))>0$ and a bias decoder $D(\mu_{Y|U_n}(\cdot |\cdot)|| v^{\phi(S_n)}(\cdot|\cdot) |\mu_{U_n})>0$. As we can't disentangle these two sources, we argue that the dominant term is the lack of expressiveness of the encoder that introduces an information loss that cannot be remediated even by selecting the optimal decoder, as predicted by Theorem \ref{th_main_IS_mismatch}. 
	
	\item {\bf IS Prior in Performance} Finally, when comparing side-by-side the curves with the IS  pre-encoder (the application of $\eta_{1,2,3.4,5}(\cdot)$) to the ones that don't use this IS projection, 
	we see consistently better performance across all the models, MLP schemes,  epochs ($k$) and sample sizes ($n$). The gain attributed to using this prior knowledge is more prominent for the two discriminative models and in the small sampling-size regime. In addition, when $n$ increases, the gain attributed to the IS encoder vanishes. This trend indicates that the supervised information of the training set dominates over the prior IS knowledge in the large sampling-size regime, as expected. 
\end{itemize}

To summarize our empirical findings, 
we observe a relevant dependency between how difficult it is to learn (with NN encoder-decoder architectures) the true predictive distribution of a model and the magnitude of its conditional entropy. This last information indicator is also proportional to the performance gain of using more complex NN architectures. Importantly, evidence supports the expressive power of MLPs to learn the actual predictive distribution (in the KLD sense from Theorem \ref{th_consistency}) and, consequently, that  MLPs can achieve the optimal information bound (Theorem \ref{th_main_IS_mismatch}) for cross-entropy learning. To our knowledge, this is the first empirical analysis that uses formal performance results and presents evidence in this direction. Finally, we show that IS knowledge introduced in this work consistently provides a performance advantage, and this gain is particularly relevant in low-data regimes.  

\section{Summary and Discussion} 
\label{sec_structure_analysis}
In this work, we expand the theory of representation learning to model and understand the role of encoder-decoder design in ML from an information-theoretic angle. Our results show that information sufficiency (IS) and information loss (IL) are central elements to understanding and measuring encoder expressiveness in modern ML, respectively. 
Here, we highlight some of the consequences and interpretations of the presented results:

\begin{itemize}
\item 
We study probabilistic structure driven by the ML task of predicting $Y$ from $X$.  
In this scenario, the role played by an encoder $\eta(\cdot)$ as a sufficient representation of discrimination is central to the analysis. We show that $\eta(\cdot)$ offers a strategy to organize classes of models (indexed by $\eta(\cdot)$) where we focus exclusively on the predictive dimension of $\mu_{X,Y}$ (i.e., on $\mu_{Y|X}(\cdot|\cdot)$). 

\item   
Theorem~\ref{th_representation}
tells us that if $\mu_{X,Y}\in \mathcal{P}_{\eta}(\mathcal{X}\times \mathcal{Y})$ (see Def.~\ref{def_IS_redundant_models}),  its predictive distribution $\mu_{Y|X}(\cdot|\cdot)$ is fully characterized by the following functional expression: 
	$Y=f(W,\eta(X))$.
Interpreting this result in the language of representation learning, we recognize the encoder $\eta(\cdot)$ (constant over $\mathcal{P}_{\eta}(\mathcal{X}\times \mathcal{Y})$) and a stochastic decoder given by the function $f(\cdot,\cdot)$. This encoder-decoder structure (see Fig. \ref{fig6}) 
implies that 
	$\mu_{Y|X}(\cdot|x) = \mu_{Y|U}(\cdot|\eta(x))$,
i.e.,  the computation of the posterior probability of $\mu_{X,Y}\in \mathcal{P}_{\eta}(\mathcal{X}\times \mathcal{Y})$ 
can be made from the  latent domain $\mathcal{U}$ with no prediction loss. 
The description presented  in Theorem~\ref{th_representation} offers an interpretation that can be used for ML design, and it is naturally aligned with the encoder-decoder stages used by many ML algorithms.

\item In Theorem~\ref{lm_invariant_IS}, we show that the important collection of the models that are invariant to the action of a compact group (Def.~\ref{def_strong_invariances}) is an instance of models with an  IS latent structure.  
Other relevant examples of models with an IS latent structure are presented in Sections \ref{sec_robust} and  \ref{sub_sec_is_in_learning}. Here, we offer connections with the type of model structure widely used for data-compression \cite{candes_2006b,gersho_1992} (digital models) and compressed sensing \cite{donoho_2006} (sparse models).

\item Studying the realistic possibility of an IS mismatch scenario, Theorem~\ref{th_main_IS_mismatch} shows that the mutual information loss 
$I(X;Y|U)\geq 0$  (induced by $\eta(\cdot)$ in $\mu_{X,Y}$) precisely measures the performance degradation, in cross-entropy risk, induced by a learning agent that assumes that $\mu_{X,Y}\in \mathcal{P}_{\eta}(\mathcal{X}\times \mathcal{Y})$. 
Importantly, Theorem~\ref{th_main_IS_mismatch} also confirms that the encoder-decoder architectures presented in Section~\ref{sub_sec_is_in_learning} are expressive (optimal in the cross-entropy sense) when $\mu_{X,Y}\in \mathcal{P}_{\eta}(\mathcal{X}\times \mathcal{Y})$.  

\item An elegant information projection (IP) interpretation of the IS mismatch scenario (studied in Theorem~\ref{th_main_IS_mismatch}) is presented in Theorem~\ref{th_if_projection_aanlogy}. This model projection analogy 
is insightful and offers a tool to analyze the individual effect (information expressiveness) of each layer in a modern multilayer sequential architecture (Theorem \ref{pro_multilayer_info_loss}).

\item On the problem of universal cross-entropy learning, we establish in Theorem \ref{th_consistency} a necessary and sufficient condition to achieve the best performance: $H(Y|X)$. Complementing this finding, we look at the individual role of the encoder and the decoder stages presenting specific conditions to meet strong consistency (or asymptotic learnability). To meet the optimal learning performance bound $H(Y|X)$, we show that the (data-driven) encoder stage needs to find an IS representation and the (data-driven) soft decoder needs to approximate (in the KL divergence sense) the true predictive distribution in the latent transform domain. 

\item Equipped with these results, we confirm the expressive power of digitalization for cross-entropy learning (in Lemma \ref{lm:expresive_VQ}) and the adequacy of the celebrated information bottleneck (IB) principle as a criterion to guide the selection of a compressed (digital) IS latent presentation (in Lemma \ref{lm:IB_expressive}).
\end{itemize}

\subsection{Learning Design and Entropy Estimation}
Our empirical results in Section \ref{sec_numerical} provide evidence that conditional entropy is not only fundamental in theory (see Theorems \ref{th_main_IS_mismatch} and \ref{th_consistency}) but is very informative as a practical indicator of the dynamic of learning a specific task in the cross entropy sense. Given that, an estimation of this quantity could be used to understand the complexity of the problem and predict the type of ML architecture that better fits the scenario. The literature on information measure estimation is rich \cite{paninski_2003, silva_2018, silva_2010, silva_isit_2007,silva_2010b,silva_2012}, and many methods can be adapted to provide a consistent proxy for the conditional entropy in the mixed continuous-discrete setting. On the other hand, extending the presented numerical analysis over a large class of models and ML schemes and exploring the practical use of data-driven entropy estimators as a proxy to condition the ML design are exciting directions for future work.  

\section{Acknowledgment}
This material is based on work supported by grants of CONICYT-Chile, Fondecyt 1210315 and the Advanced Center for Electrical and Electronic Engineering, Basal Project FB0008. C. Ramírez is supported by ANID-Subdirección de Capital Humano/Magíster-Nacional/2023 - 22230232 master's scholarship.

\newpage
\appendices
\section{Proof of Theorem \ref{th_representation}}
\label{proof_th_representation}

\subsection{Preliminaries}
For the proof of Theorem \ref{th_representation}, we use the following two results: 
 
{\bf Lemma \ref{lm_is_encode_d-sparation}:} 
	{\em $\eta(\cdot)$ is information sufficient (IS) for $\mu_{X,Y}$ (Def.~1)  if, and only if, $U=\eta(X)$ D-separates $X$ and $Y$ in the sense that $X$ and $Y$ are independent given $U$, i.e., $X \rightarrow U \rightarrow Y$.}
	
\begin{proof}
Using the assumption that $\eta(X)$ offers a D-separation between $X$ and $Y$, this means that we have the following {\em Markov chain}: $X \rightarrow U \rightarrow Y$. Then, by the data-processing inequality of the MI \cite{cover_2006}, we have that $I(X;Y)\leq I(U;Y)$. On the other hand,  $U$ is a deterministic (measurable) function of $X$, this means that $I(X;Y)\geq I(U;Y)$ by the same data-processing inequality and, consequently, $I(X;Y)=I(U;Y)$. Then,  $U$ is IS for $\mu_{X,Y}$ by definition.  
	
	For the other implication, let us assume that $U=\eta(X)$ is IS.  This means that $I(X;Y)=I(U;Y) \Leftrightarrow I(X;Y|U)=0$ \cite{cover_2006}.  This last equivalence is obtained by the following well-known identity \cite{cover_2006}:
	\begin{equation} \label{eq_ps_5}
		I(X;Y) - I(U;Y)= I((X,U);Y) - I(U;Y) = I(X;Y | U) \geq 0,
	\end{equation}
	where 
	\begin{equation} \label{eq_ps_5b}
		 I(X;Y | U) \equiv \int_{\mathcal{U}} \mathcal{I}(\mu_{X,Y|U}(\cdot | u))\:\mathrm{d}\mu_U(u) \geq 0.
	\end{equation}
	In (\ref{eq_ps_5b}), $\mathcal{I}(\mu_{X,Y|U}(\cdot | u))$ is the MI of the joint model $\mu_{X,Y|U}(\cdot | u)\in \mathcal{P}(\mathcal{X} \times \mathcal{Y}),\forall u\in\mathcal{U}$, and $\mu_U$  denotes the marginal distribution of $U$. Using the expression in (\ref{eq_ps_5b}), we have from our IS assumption that 
	\begin{equation} \label{eq_ps_5c}
	\int_{\mathcal{U}} \mathcal{I}(\mu_{X,Y|U}(\cdot | u))\:\mathrm{d}\mu_U(u) = 0.
	\end{equation}
As the mutual information is non-negative \cite{cover_2006}, the previous equality implies that the term $\mathcal{I}(\mu_{X,Y|U}(\cdot | u))=0$ (as a function of $u$) for $\mu_U$-almost every point in $\mathcal{U}$.\footnote{Formally, this means that the measurable set $ A= \left\{u\in \mathcal{U},\text{ s.t. }\mathcal{I}(\mu_{X,Y|U}(\cdot | u))=0  \right\}$ satisfies that $\mu_U(A)=1$ \cite{halmos_1950}.} At this point, we use the fact  that \cite{gray_1990_b}
	\begin{equation} \label{eq_ps_5d}
	\mathcal{I}(\mu_{X,Y|U}(\cdot | u))= D(\mu_{X,Y|U}(\cdot | u)\,||\, \mu_{X|U}(\cdot | u) \times \mu_{Y|U}(\cdot | u) )   ,
	\end{equation}
where $D(p||q)$ is the Kullback-Leibler (KL) divergence between two probability distributions \cite{kullback1958}.   Then, the condition $\mathcal{I}(\mu_{X,Y|U}(\cdot | u))=0$ from (\ref{eq_ps_5d}) implies that $\mu_{X,Y|U}(\cdot | u) = \mu_{X|U}(\cdot | u) \times  \mu_{Y|U}(\cdot | u)$, i.e., the joint distribution of $(X,Y)$ given $U=u$ is equal to the multiplication of the marginals of $X$ and $Y$ given $U=u$.   Finally, using that $\mathcal{I}(\mu_{X,Y|U}(\cdot | u))=0$ for $\mu_U$-almost every $u$, this means that $X$ and $Y$ are independent given $U=u$ for $\mu_U$-almost every point.  This  is equivalent to state that $U=\eta(X)$ D-separates $X$ and $Y$. 
\end{proof}

In addition, we will use the following result by Bloem-Reddy and Teh \cite{Bloem_2019}: 
\begin{lemma}\label{lm_functional_d_separation} \cite[Lemma 5, pp. 15]{Bloem_2019}
	Let $(X,Y)$ be our joint observation-class random variable following $\mu_{X,Y}\in \mathcal{P}(\mathcal{X} \times  \mathcal{Y})$. Let $\eta:(\mathcal{X},\mathcal{B}(\mathcal{X})) \rightarrow (\mathcal{U},\mathcal{B}(\mathcal{U}))$ be a lossy encoder. If $\eta(\cdot)$ D-separates $(X,Y)$ in the sense that $X \rightarrow \eta(X) \rightarrow Y$, then there exists a measurable function $f:[0,1] \times \mathcal{U} \rightarrow \mathcal{Y}$ such that 
	\begin{equation} \label{eq_ps_8}
		Y=f(W,\eta(X))
	\end{equation}
	almost surely\footnote{Almost surely with respect to the joint (product) distribution of the pair $(X,W)$.}, where $W$ is a random variable in $[0,1]$ with uniform distribution (i.e., $W \sim \emph{Unif}[0,1]$) that is independent of $X$. 
\end{lemma}

\subsection{Main Argument}	
\begin{proof}
	For the direct (forward) implication, the condition $\mu_{X,Y}\in\mathcal{P}_{\eta}(\mathcal{X}\times\mathcal{Y})$  implies that $\eta(\cdot)$ is IS for $\mu_{X,Y}$, then $\eta(X)$ D-separates $X$ and $Y$ (by~Lemma \ref{lm_is_encode_d-sparation}) and, consequently, from Lemma~\ref{lm_functional_d_separation}  the functional structure stated in Eq.(\ref{eq_ps_9}) follows from (\ref{eq_ps_8}).
	For the converse implication, it is simple to note that if a pair $(X,Y)$ is constructed as in (\ref{eq_ps_9}) (using a function $f(\cdot,\cdot)$ and a noise $W$ that is independent of $X$) then 
	\begin{equation}\label{eq_proof_th1}
	 I(X;Y|\eta(X))=H(Y|\eta(X))- H(Y|X, \eta(X)) = H(Y|\eta(X))- H(Y|\eta(X))= 0. 
	 \end{equation}
	The first equality in (\ref{eq_proof_th1}) is by definition of the conditional MI \cite{cover_2006},  and the last equality is by the functional construction of $Y$  given $\eta(X)=u\in \mathcal{U}$ in (\ref{eq_ps_9}) and the fact that $W$ has a distribution that is invariant (independent of) of the value of $X$ by construction.
	 Consequently, we have from (\ref{eq_ps_5}) that $X \rightarrow \eta(X) \rightarrow Y$, then $\eta(\cdot)$ is IS for $\mu_{X,Y}$ from Lemma \ref{lm_is_encode_d-sparation}, which means that $\mu_{X,Y}\in\mathcal{P}_{\eta}(\mathcal{X}\times\mathcal{Y})$.
\end{proof}

\section{Proof of Theorem \ref{lm_invariant_IS}}
\label{proof_lm_invariant_IS}
We first prove that if $\mu_{X,Y}$ is predictive invariant w.r.t. $\mathcal{G}$ (Def. \ref{def_strong_invariances})  then $\eta^*_\mathcal{G}(\cdot)$ is IS for $\mu_{X,Y}$ (Def. \ref{def_is}):\\ 
\begin{proof}	
Let us assume that $\mu_{Y|X}$ is $\mathcal{G}$-invariant. Using Eq.~(\ref{eq_pi_1}), we have that for any $g \in \mathcal{G}$ and any event $\cdot \subset \mathcal{Y}$\footnote{From definition in (\ref{eq_pi_1}) and the fact that $g^{-1}\in \mathcal{G}$ for any $g\in \mathcal{G}$.}
\begin{equation}\label{eq_proof_lm_invariant_IS_1}
	\mu_{Y|X}(\cdot |\left\{ x \right\})= \mu_{Y|X}(\cdot | \left\{ g(x) \right\})
\end{equation}
$\mu_X$-almost surely in $\mathcal{X}$.  Let us denote by $U\equiv \eta^*_\mathcal{G}(X)$ and by 
\begin{equation}\label{eq_proof_lm_invariant_IS_1b}
\text{orbit}(x)\equiv \left\{g(x),g \in \mathcal{G}\right\} = {\eta^*_\mathcal{G}}^{-1}( \left\{\eta^*_\mathcal{G}(x) \right\}),
\end{equation}
where we know that $\text{orbit}(x) \in \mathcal{B}(\mathcal{X})$\footnote{It is known that $\eta^*_{\mathcal{G}}(\cdot)$ is a measurable function when $\mathcal{G}$ is a compact group \cite{eaton_1889}.}. 
From
(\ref{eq_proof_lm_invariant_IS_1}), we have the following result: 
\begin{proposition}\label{pro_orbit} If $\mu_{Y|X}$ is $\mathcal{G}$-invariant then 
	$\mu_{Y|X}(\cdot | \left\{ x \right\}) = \mu_{Y|X}(\cdot\,|\,\emph{orbit}(x))$ 
	$\mu_X$-almost surely. 
	\\(The proof is presented in Appendix \ref{proof_pro_orbit}).
\end{proposition}
From this result, we have that for any $A\subset \mathcal{Y}$ and $\mu_X$-almost every $x\in \mathcal{X}$:
\begin{align}\label{eq_proof_lm_invariant_IS_2}
	\mathbb{P}(Y\in A|X=x) = \mathbb{P}(Y\in A|X=x, U=\eta^*_\mathcal{G}(x)) =  \mathbb{P}(Y\in A|U=\eta^*_\mathcal{G}(x)).
\end{align}
The first equality comes from the fact that $x\in \text{orbit}(x)$ and the definition of $U$.  The second equality in (\ref{eq_proof_lm_invariant_IS_2}) comes from Proposition \ref{pro_orbit}
considering that the event $U=\eta^*_\mathcal{G}(x)$ is equivalent to $X\in {\eta^*_\mathcal{G}}^{-1}( \left\{\eta^*_\mathcal{G}(x) \right\})=\text{orbit}(x)$. This last equality means that $X$ and $Y$ are independent given $U=u\in\text{range}(\eta^*_\mathcal{G}) \equiv  \left\{\eta^*_\mathcal{G}(x), x\in \mathcal{X} \right\}$, 
and, consequently, from Lemma \ref{lm_is_encode_d-sparation} we have that $\eta^*_\mathcal{G}(\cdot)$ is IS for $\mu_{X,Y}$.
\end{proof}

Conversely, we need to prove that if  $\eta^*_\mathcal{G}(\cdot)$ is IS for $\mu_{X,Y}$  then $\mu_{X,Y}$ is predictive invariant w.r.t. $\mathcal{G}$:\\  
\begin{proof}
Here we assume that $\eta^*_\mathcal{G}(\cdot)$ is IS for $\mu_{X,Y}$.  This means that $X$ and $Y$ are independent given $U=\eta^*_\mathcal{G}(X)$ (from Lemma \ref{lm_is_encode_d-sparation}), and in particular, that for $\mu_X$-almost every $x \in \mathcal{X}$:
\begin{equation} \label{eq_proof_lm_invariant_IS_3}
	\mu_{Y|X}(\cdot | \left\{ x \right\}) = \mu_{Y|X,U} (\cdot | \left\{ x \right\} \times \left\{ \eta^*_\mathcal{G}(x) \right\})=  \mu_{Y|U} (\cdot | \left\{ \eta^*_\mathcal{G}(x) \right\})= \mu_{Y|X}(\cdot \:|\: \text{orbit}(x)), 
\end{equation}
where the first equality is by definition of conditional probability considering that\footnote{$\Omega$ denotes the sample space in which $(X,Y)$ is defined.}  
$$\left\{\omega\in\Omega: X(\omega)=x\right\} \subset \left\{\omega\in\Omega: U(\omega)= \eta^*_\mathcal{G}(x) \right\} \subset \Omega.$$  The second and third equalities in (\ref{eq_proof_lm_invariant_IS_3}) are obtained  by the Markov chain structure $X \rightarrow U \rightarrow Y$ (using the IS hypothesis) and the fact that $\text{orbit}(x)= {\eta^*_\mathcal{G}}^{-1}( \left\{\eta^*_\mathcal{G}(x) \right\})$, respectively.  

On the other hand, using the fact that $ \eta^*_\mathcal{G}(\cdot)$ is maximal $\mathcal{G}$-invariant, it follows that  for any $g\in \mathcal{G}$ and $x\in \mathcal{X}$  
(see Proposition \ref{prop_invariant_property} and its proof in Appendix \ref{sec_proof_prop_invariant_property})
\begin{equation} \label{eq_proof_lm_invariant_IS_5}
	 {\eta^*_\mathcal{G}}^{-1}(  \left\{ \eta^*_\mathcal{G}(x) \right\})  =   {\eta^*_\mathcal{G}}^{-1}(  \left\{ \eta^*_\mathcal{G}(g^{-1}(x)) \right\}). 
\end{equation}
Consequently, from (\ref{eq_proof_lm_invariant_IS_5}) and (\ref{eq_proof_lm_invariant_IS_3}), we have that for $\mu_X$-almost every $x$
\begin{align} 
    \label{eq_proof_lm_invariant_IS_6}
	\mathbb{P}(Y\in \cdot | X =  x ) = \mu_{Y|X}(\cdot |\left\{x \right\}) &= \mu_{Y|X} (\cdot |  {\eta^*_\mathcal{G}}^{-1} (\left\{\eta^*_\mathcal{G}(x) \right\}))\\
    \label{eq_proof_lm_invariant_IS_6b}
    &= \mu_{Y|X} (\cdot | {\eta^*_\mathcal{G}}^{-1}(  \left\{ \eta^*_\mathcal{G}(g^{-1}(x)) \right\}) )\\
    \label{eq_proof_lm_invariant_IS_6c}
    &= \mu_{Y|X} (\cdot |  g^{-1}(\left\{x \right\}))\\
    &= \mathbb{P}(Y\in \cdot | g(X) = x), 
\end{align}
meaning that $\mu_{Y|X}$ is $\mathcal{G}$-invariant (see Def. \ref{def_strong_invariances}).  The equality in (\ref{eq_proof_lm_invariant_IS_6}) comes from (\ref{eq_proof_lm_invariant_IS_3}), the equality in (\ref{eq_proof_lm_invariant_IS_6b}) from (\ref{eq_proof_lm_invariant_IS_5}), and the condition in (\ref{eq_proof_lm_invariant_IS_6c}) from (\ref{eq_proof_lm_invariant_IS_3}) again. 
\end{proof}

\section{Proof of Theorem \ref{lemma_IS_robust}}
\label{proof_lemma_IS_robust}
\begin{proof}
For the direct implication, we know by Def.~\ref{def_robust} that  $\mu_{Y|X}(\cdot| \left\{ x \right\} )=\mu_{Y|X}(\cdot|  \left\{ \bar{x} \right\})$ for any $x, \bar{x}\in A_i$ and for any  $A_i \in \pi_{\mathcal{I}}$. This relationship implies that for all $B\subset \mathcal{Y}$ (following the same argument used to  derive Proposition \ref{pro_orbit}): 
\begin{equation}\label{eq_appendix_c_1}
	\mu_{Y|X}(B | \left\{ x \right\}) = \mu_{Y|X}(B | \pi_{\mathcal{I}}(x)), 
\end{equation}
where $\pi_{\mathcal{I}}(x)\in \pi_{\mathcal{I}}\subset \mathcal{B}(\mathcal{X})$ is a short-hand for the 
cell in $\pi_{\mathcal{I}}$ that contains $x\in \mathcal{X}$. On the other hand, by definition of $U=\eta_{\pi}(X)$ it follows that
\begin{equation}\label{eq_appendix_c_2}
	\mu_{Y|U}(B| \left\{ \eta_{\pi}(x) \right\})= \mu_{Y|X}(B | \pi_{\mathcal{I}}(x)).
\end{equation}
From (\ref{eq_appendix_c_1}) and (\ref{eq_appendix_c_2}) we have that 
\begin{equation}\label{eq_appendix_c_2b}
\mu_{Y|X}(B | \left\{ x \right\})=\mu_{Y|U}(B| \left\{ \eta_{\pi}(x) \right\})
\end{equation}
and using the fact 
that $\mu_{U|X}( \left\{ \eta_{\pi}(x) \right\} |  \left\{ x \right\} )=1$,\footnote{This equality is obtained from the fact that $\eta_\pi(\cdot)$ is a deterministic mapping.} it follows from (\ref{eq_appendix_c_2b}) that
\begin{equation}\label{eq_appendix_c_3}
	\mu_{Y|X}(B | \left\{ x \right\}) = \mu_{U|X}( \left\{ \eta_{\pi}(x) \right\} |  \left\{ x \right\} ) \times  \mu_{Y|U}(B| \left\{ \eta_{\pi}(x) \right\}).
\end{equation}
This result is valid for any $x\in \mathcal{X}$, which implies the following Markov Chain $X \rightarrow U \rightarrow Y$. From Lemma 1, this is equivalent to the condition that $\eta_{\pi}(\cdot)$ is IS for $\mu_{X,Y}$. 

For the converse implication,  we assume that $\eta_{\pi}(\cdot)$ is IS for $\mu_{X,Y}$.  Then we have the following: 
\begin{align} 
    \label{eq_appendix_c_4}
	\mu_{Y|X}(B | \left\{ x \right\}) &= \mu_{Y|X,U}(B | \left\{ x \right\} \times \left\{ \eta_{\pi}(x) \right\} )\\
    \label{eq_appendix_c_4b}
    &=  \mu_{Y|U}(B |  \left\{ \eta_{\pi}(x) \right\}).
\end{align}
The first equality in (\ref{eq_appendix_c_4}) is by definition of conditional probability and from the observation that $U$ is a deterministic r.v. given $X=x$. The second equality in (\ref{eq_appendix_c_4b}) derives from  the fact that $X \rightarrow U \rightarrow Y$ under the assumption that $\eta_{\pi}(\cdot)$ is IS for $\mu_{X,Y}$ and Lemma \ref{lm_is_encode_d-sparation}. From the identity in (\ref{eq_appendix_c_4b}), it follows directly that for any $i\in \mathcal{I}$ and any pair $x,\bar{x} \in A_i$, we have that 
\begin{equation}\label{eq_appendix_c_3}
\mu_{Y|X}(B | \left\{ x \right\})= \mu_{Y|X}(B | \left\{ \bar{x} \right\}) = \mu_{Y|X}(B | A_i),
\end{equation}
which concludes that $\mu_{X,Y}$ is robust to perturbations within the cells of $\pi_\mathcal{I}= \left\{A_i,i\in \mathcal{I} \right\}$ (see Def.~\ref{def_robust}). 
\end{proof}

\section{Proof of Theorem \ref{th_main_IS_mismatch}}
\label{proof_th_main_IS_mismatch}
\begin{proof} 
	Concerning the almost sure convergence of $\hat{r}(v^\theta_{\tilde{Y}|U}(\cdot|\cdot),S_n)$ to $r(v^\theta_{\tilde{Y}|U}(\cdot|\cdot),\mu_{X,Y})$ as $n$ tends to infinity  in (\ref{eq_sec_mis_5a}) this is by the law of large numbers \cite{breiman_1968}, where 
\begin{align}\label{eq_p_th_main_IS_mismatch_1}
	r(v^\theta_{\tilde{Y}|U}(\cdot|\cdot),\mu_{X,Y}) 
		&\equiv  \mathbb{E}_{(X,Y)\sim \mu_{X,Y}} \left\{ - \log v^\theta_{\tilde{Y}|U}(Y|\eta(X)) \right\}\nonumber\\  
		&= \mathbb{E}_{(U,Y)\sim \mu_{U,Y}} \left\{ - \log v^\theta_{\tilde{Y}|U}(Y|U) \right\}.  
\end{align}
In the last expression, $U=\eta(X)$ and $\mu_{U,Y} \in \mathcal{P}(\mathcal{U} \times \mathcal{Y})$ is the (true) joint distribution of $(U,Y)$ induced by $\eta(\cdot)$ and $\mu_{X,Y}$ \cite{gray_2004}.  Working with the (transform domain) cross-entropy term in (\ref{eq_p_th_main_IS_mismatch_1}), it follows that
\begin{align}\label{eq_p_th_main_IS_mismatch_2}
\mathbb{E}_{(U,Y)\sim \mu_{U,Y}} \left\{  \log \frac{1}{v^\theta_{\tilde{Y}|U}(Y|U)} \right\} &=   \mathbb{E}_{(U,Y)} \left\{  \log \frac{\mu_{Y|U}(Y|U)}{v^\theta_{\tilde{Y}|U}(Y|U)} \right\} + \mathbb{E}_{(U,Y)} \left\{  \log \frac{1}{\mu_{Y|U}(Y|U)} \right\} \nonumber\\
	&=D(\mu_{Y|U}(\cdot|\cdot)||v^\theta_{\tilde{Y}|U}(\cdot|\cdot) |\mu_U) + H(Y|U),
\end{align}
where $\mu_{Y|U}(\cdot|\cdot) \in \mathcal{P}(\mathcal{Y} | \mathcal{U})$ is the true posterior obtained from $\mu_{U,Y}$, $H(Y|U)$ is the conditional entropy of $Y$ given $U$ \cite{cover_2006} and
\begin{align}\label{eq_p_th_main_IS_mismatch_3}
D(\mu_{Y|U}(\cdot|\cdot)||v^\theta_{\tilde{Y}|U}(\cdot|\cdot) |\mu_U) \equiv \mathbb{E}_{U\sim \mu_U}  \left\{D(\mu_{Y|U}(\cdot|U)||v^\theta_{\tilde{Y}|U}(\cdot|U)) \right\}.
\end{align}	 
In (\ref{eq_p_th_main_IS_mismatch_3}), $D(p||q)$ denotes the discrete KL divergence \cite{cover_2006} between any $q,p \in \mathcal{P}(\mathcal{Y})$ and $\mu_U$ is the marginal distribution of $U$ obtained from the joint $\mu_{U,Y}$.  Integrating, we have that: 
\begin{align}\label{eq_p_th_main_IS_mismatch_4}
	r(v^\theta_{\tilde{Y}|U}(\cdot|\cdot),\mu_{X,Y}) &=  \mathbb{E}_{U\sim \mu_U}  \left\{D(\mu_{Y|U}(\cdot|U)||v^\theta_{\tilde{Y}|U}(\cdot|U)) \right\} + H(Y|U)\nonumber\\ 
		&= \mathbb{E}_{U}  \left\{D(\mu_{Y|U}(\cdot|U) ||v^\theta_{\tilde{Y}|U}(\cdot|U)) \right\} +I(X;Y|U) +  H(Y|X)
\end{align}
where the last identity in (\ref{eq_p_th_main_IS_mismatch_4}) comes from the application of the chain rule of the MI \cite{cover_2006}, i.e.,\footnote{The first equality in (\ref{eq_p_th_main_IS_mismatch_5}) from the fact that $U$ is a deterministic function of $X$.} 
\begin{align}\label{eq_p_th_main_IS_mismatch_5}
	I(X;Y) = I((X,U); Y) = I(U;Y) + I(X;Y|U),
\end{align}
which implies that 
\begin{align}\label{eq_p_th_main_IS_mismatch_5b}
	H(Y|U) = H(Y|X) + I(X;Y|U).
\end{align}

To conclude the argument, we use the following \cite{cover_2006,gray_1990_b}:
\begin{itemize}
	\item[a)] $D(p||q)\geq 0$ for any $q,p \in \mathcal{P}(\mathcal{Y})$. 
	\item[b)] $D(p||q)= 0$ if, and only if, $q=p$ in total variation. 
\end{itemize}

From a),  it follows that  $\mathbb{E}_{U\sim \mu_U}  \left\{D(\mu_{Y|U}(\cdot|U)||v^\theta_{\tilde{Y}|U}(\cdot|U)) \right\}\geq 0$.  This last inequality implies the main bound in Eq.(\ref{eq_sec_mis_5a}) from the identity stated in (\ref{eq_p_th_main_IS_mismatch_4}). 

Concerning the task of achieving the optimal cross-entropy lower bound $I(X;Y|U) +  H(Y|X)$ in (\ref{eq_p_th_main_IS_mismatch_4}),  the evident optimality condition $\mathbb{E}_{U\sim \mu_U}  \left\{D(\mu_{Y|U}(\cdot|U)||v^\theta_{\tilde{Y}|U}(\cdot|U)) \right\}=0$ is equivalent to the condition that $D(\mu_{Y|U}(\cdot|u)||v^\theta_{\tilde{Y}|U}(\cdot|u))=0$ for $\mu_U$ almost every point $u\in \mathcal{U}$ \cite{halmos_1950}, which is equivalent to say from b) that  $v^\theta_{\tilde{Y}|U}(\cdot|u)\in \mathcal{P}(\mathcal{Y})$ is the same (in total variation) to the true posterior $\mu_{Y|U}(\cdot|u) \in \mathcal{P}(\mathcal{Y})$ for $\mu_U$-almost every conditional value $u\in  \mathcal{U}$, i.e., 
\begin{align}\label{eq_p_th_main_IS_mismatch_5c}
	\mu_U  \left(   \left\{u\in \mathcal{U}: V(\mu_{Y|U}(\cdot|u),v^\theta_{\tilde{Y}|U}(\cdot|u)))=0  \right\} \right)=1
\end{align}
where $V(p,q)\equiv \sup_{B\subset \mathcal{Y}}  \left| p(B)-  q(B) \right|$ is the total variational distance in $\mathcal{P}(\mathcal{Y})$ \cite{devroye_2001}. The result in (\ref{eq_p_th_main_IS_mismatch_5c}) is precisely the condition stated in (\ref{eq_sec_mis_5b}).
\end{proof}

\section{Proof of The Information Projection Analogy: Theorem \ref{th_if_projection_aanlogy}}
\label{app_IS-projection}

\begin{proof}
Let us analize the problem 
\begin{equation}\label{eq_app_IS_projection_3}
	\min_{\tilde{\mu}_{X,Y} \in \mathcal{P}_\eta(\mathcal{X} \times \mathcal{Y})} D(\mu_{X,Y} || \tilde{\mu}_{X,Y}).  
\end{equation}
We know from Theorem \ref{th_representation} that  for any $\tilde{\mu}_{X,Y}\in \mathcal{P}_\eta(\mathcal{X} \times \mathcal{Y})$, $\exists f:[0,1] \times \mathcal{U} \rightarrow \mathcal{Y}$ such that $\tilde{Y}=f(W,\eta(\tilde{X}))$ where $W \sim Unif[0,1]$.  Considering $(\tilde{X},\tilde{Y})\sim \tilde{\mu}_{X,Y}$, the mentioned functional expression meets that $I(\tilde{X};\tilde{Y}|\eta(\tilde{X}))=0$ implying that $\tilde{\mu}_{\tilde{Y}|\eta(\tilde{X}),\tilde{X}} (\cdot |\cdot,\cdot) = \tilde{\mu}_{\tilde{Y}|\eta({X})}(\cdot|\cdot)$ and, consequently, we have the following factorization $\tilde{\mu}_{\tilde{X},\tilde{Y}}= \tilde{\mu}_{\tilde{X}} \cdot \tilde{\mu}_{\tilde{Y}|\eta(\tilde{X})}(\cdot|\cdot)$. 

Using this factorization, for any $\tilde{\mu}_{X,Y} \in \mathcal{P}_\eta(\mathcal{X} \times \mathcal{Y})$: 
\begin{align}\label{eq_app_IS_projection_4}
	D(\mu_{X,Y} || \tilde{\mu}_{X,Y})
		&= \mathbb{E}_{(X,Y)\sim \mu_{X,Y}}  \left\{ \log \frac{\mu_{X,Y}(X,Y)}{\tilde{\mu}_{X,Y}(X,Y)} \right\}\nonumber\\
		&= \mathbb{E}_{X\sim \mu_{X}}  \left\{ \log \frac{\mu_{X}(X)}{\tilde{\mu}_{X}(X)} \right\} + \mathbb{E}_{(X,Y)\sim \mu_{X,Y}}  \left\{ \log \frac{\mu_{Y|X}(Y|X)}{\tilde{\mu}^f_{Y|\eta(X)}(Y|\eta{(X)})} \right\}, 
\end{align}
where $f(\cdot)$ is the function associated to $\tilde{\mu}_{X,Y}$ (from Theorem \ref{th_representation}) and 
$\mu^f_{Y|\eta(X)}\in \mathcal{P}(\mathcal{Y}|\mathcal{U})$ is the predictive distribution induced by $f(\cdot)$ using 
the functional expression in (\ref{eq_ps_9}).  

Returning to (\ref{eq_app_IS_projection_3}), we use the decomposition in (\ref{eq_app_IS_projection_4}) and 
the functional characterization of $\mathcal{P}_\eta(\mathcal{X} \times \mathcal{Y})$  (in Theorem \ref{th_representation}) to obtain that
\begin{align} \label{eq_app_IS_projection_5}
	\min_{\tilde{\mu}_{X,Y} \in \mathcal{P}_\eta(\mathcal{X} \times \mathcal{Y})} D(\mu_{X,Y} || \tilde{\mu}_{X,Y}) 
			&= \min_{\tilde{\mu}_{X} \in \mathcal{P}(\mathcal{X})} D(\mu_X||\tilde{\mu}_{X}) +  \min_{f:[0,1] \times \mathcal{U} \rightarrow \mathcal{Y}} \mathbb{E}_{(X,Y)\sim \mu_{X,Y}}  \left\{ \log \frac{\mu_{Y|X}(Y|X)}{\tilde{\mu}^f_{Y|\eta(X)}(Y|\eta{(X)})} \right\}\\
			\label{eq_app_IS_projection_5b}
			&= \min_{f:[0,1] \times \mathcal{U} \rightarrow \mathcal{Y}} \mathbb{E}_{(X,Y)\sim \mu_{X,Y}}  \left\{ \log \frac{\mu_{Y|X}(Y|X)}{\tilde{\mu}^f_{Y|\eta(X)}(Y|\eta{(X)})} \right\}.
\end{align}
The decomposition in (\ref{eq_app_IS_projection_5}) derives from noting that when $\tilde{\mu}_{X,Y}\in \mathcal{P}_\eta(\mathcal{X} \times \mathcal{Y})$ Theorem \ref{th_representation} tells us that we only restrict the predictive part of $\tilde{\mu}_{X,Y}$. Then (\ref{eq_app_IS_projection_5b}) derives from the fact  that $\mu_X\in \mathcal{P}(\mathcal{X})$ which implies that $\min_{\tilde{\mu}_{X} \in \mathcal{P}(\mathcal{X})} D(\mu_X||\tilde{\mu}_{X})=0$.  

We claim at this point that solving (\ref{eq_app_IS_projection_5b}) (i.e., solving (\ref{eq_app_IS_projection_1})) is equivalent to selecting the optimal decoder (in the cross-entropy sense) given the model $\mu_{X,Y}$.  Indeed, for any $f:[0,1] \times \mathcal{U} \rightarrow \mathcal{Y}$
\begin{align} \label{eq_app_IS_projection_6}
\mathbb{E}_{(X,Y)\sim \mu_{X,Y}}  \left\{ \log \frac{\mu_{Y|X}(Y|X)}{\tilde{\mu}^f_{Y|\eta(X)}(Y|\eta{(X)})} \right\} &=
\mathbb{E}_{(X,Y)}  \left\{\log \frac{\mu_{Y|X}(Y|X)}{\mu_{Y|U}(Y|\eta{(X)})} \right\} + \mathbb{E}_{(X,Y)}  \left\{ \log \frac{\mu_{Y|U}(Y|\eta(X))}{\tilde{\mu}^f_{Y|\eta(X)}(Y|\eta{(X)})} \right\} \nonumber\\
	&=H(Y|U) - H(Y|X) + \mathbb{E}_{(U,Y)\sim \mu_{U,Y}} \left\{ \log \frac{\mu_{Y|U}(Y|U)}{\tilde{\mu}^f_{Y|U}(Y|U)} \right\}\\
	 \label{eq_app_IS_projection_6b}
	&= I(X;Y|U) + \mathbb{E}_{(U,Y)\sim \mu_{U,Y}} \left\{ \log \frac{\mu_{Y|U}(Y|U)}{\tilde{\mu}^f_{Y|U}(Y|U)} \right\}, 	 
\end{align}
where $\mu_{U,Y}$ in (\ref{eq_app_IS_projection_6}) is the distribution of $(U=\eta(X),Y)$ induced by $\mu_{X,Y}$ and $\eta(\cdot)$.  Using (\ref{eq_app_IS_projection_6b}) in (\ref{eq_app_IS_projection_5b})
\begin{align} \label{eq_app_IS_projection_7}
	\min_{\tilde{\mu}_{X,Y} \in \mathcal{P}_\eta(\mathcal{X} \times \mathcal{Y})} D(\mu_{X,Y} || \tilde{\mu}_{X,Y}) =
	  I(X;Y|U) + \min_{f:[0,1] \times \mathcal{U} \rightarrow \mathcal{Y}} D(\mu_{Y|U}(\cdot |\cdot) || \tilde{\mu}^f_{Y|U}(\cdot|\cdot) |\mu_U)
\end{align}
where 
\begin{align} \label{eq_app_IS_projection_8}
	D(\mu_{Y|U}(\cdot |\cdot) || \tilde{\mu}^f_{Y|U}(\cdot|\cdot) |\mu_U) = \mathbb{E}_{U\sim \mu_U}  \left\{D(\mu_{Y|U}(\cdot |U) || \tilde{\mu}^f_{Y|U}(\cdot|U) \right\}.
\end{align}

Coming back to the expression in (\ref{eq_p_th_main_IS_mismatch_4}) (in Appendix \ref{proof_th_main_IS_mismatch}), we note the claimed analogy:  selecting the optimal predictor within $\Lambda_{\Theta,\eta}= \left\{v^\theta_{\tilde{Y}|U}(\cdot|\eta(\cdot)), \theta \in \Theta \right\}\subset \mathbb{P}(\mathcal{Y}|\mathcal{U})$ reduces to solving $ \min_{f:[0,1] \times \mathcal{U} \rightarrow \mathcal{Y}} D(\mu_{Y|U}(\cdot |\cdot) || \tilde{\mu}^f_{Y|U}(\cdot|\cdot) |\mu_U)$ (under the expressiveness assumption in Def. \ref{def_prob_expressiviness}), which is equivalent to the information projection task presented in (\ref{eq_app_IS_projection_7}).

To conclude the argument, 
we know that  
$$\min_{f:[0,1] \times \mathcal{U} \rightarrow \mathcal{Y}} D(\mu_{Y|U}(\cdot |\cdot) || \tilde{\mu}^f_{Y|U}(\cdot|\cdot) |\mu_U)=0$$
as any distribution in $P(\mathcal{Y}|\mathcal{X})$ can be produced by a measurable function $f(\cdot,\cdot)$ and the 
functional construction $Y=f(u,W)$ with $W \sim Unif[0,1]$. Therefore, we have that 
\begin{align} \label{eq_app_IS_projection_9}
	\min_{\tilde{\mu}_{X,Y} \in \mathcal{P}_\eta(\mathcal{X} \times \mathcal{Y})} D(\mu_{X,Y} || \tilde{\mu}_{X,Y}) = I(X;Y|U),
\end{align}
where the distribution in $\mathcal{P}_\eta(\mathcal{X} \times \mathcal{Y})$ achieving the minimum is $\mu_X \cdot \mu_{Y|U}(\cdot|\cdot)$. This is simple to verify by looking again at the expressions in (\ref{eq_app_IS_projection_5}) for $\mu_x$ and (\ref{eq_app_IS_projection_7}) for $\mu_{Y|U}(\cdot|\cdot)$. 
\end{proof}

\section{Proof of Theorem \ref{pro_multilayer_info_loss}}
\label{proof_pro_multilayer_info_loss}
\begin{proof}
For proving the embedded structure of the problem in (\ref{eq_multilayer_IL_analysis_1b}), Theorem \ref{th_if_projection_aanlogy} tells us the information loss induced by an encoder $\eta(\cdot)$ can be seen as the projection of $\mu_{X,Y}$ to $ \mathcal{P}_{\eta}(\mathcal{X}\times \mathcal{Y})$ . Consequently, given the multilayer setting determined by $\left\{\eta_1^{\theta_1}(\cdot), \eta_2^{\theta_2}(\cdot),\ldots, \eta_K^{\theta_K}(\cdot)\right\}$, we look at the collections of probabilities 
$$\left\{ \mathcal{P}_{\eta^{\theta_K}_K \circ \ldots \circ \eta^{\theta_1}_1}(\mathcal{X}\times \mathcal{Y}),  \ldots,  \mathcal{P}_{\eta^{\theta_2}_2 \circ \eta^{\theta_1}_1 }(\mathcal{X}\times \mathcal{Y}) ,  \mathcal{P}_{\eta^{\theta_1}_1}(\mathcal{X}\times \mathcal{Y}) \right\},$$ 
and their inter-relationship.  
Let us first focus on $\mathcal{P}_{\eta^{\theta_2}_2 \circ \eta^{\theta_1}_1 }(\mathcal{X}\times \mathcal{Y})$ and $\mathcal{P}_{\eta^{\theta_1}_1}(\mathcal{X}\times \mathcal{Y})$.  From Theorem \ref{th_representation}, if
$\mu_{X,Y}\in \mathcal{P}_{\eta^{\theta_2}_2 \circ \eta^{\theta_1}_1 }(\mathcal{X}\times \mathcal{Y}) \Leftrightarrow \exists f: \mathcal{U}_2\times [0,1] \rightarrow \mathcal{Y}$ such that $Y=f(\eta^{\theta_2}_2 \circ \eta^{\theta_1}_1(X), W)$ where $W\sim \emph{Unif}[0,1]$ is independent of $X$.  Using $f(\cdot,\cdot)$, we can create $\tilde{f}:\mathcal{U}_1\times [0,1] \rightarrow \mathcal{Y}$ by $\tilde{f}(\eta_1(x),w) \equiv f(\eta^{\theta_2}_2 \circ \eta^{\theta_1}_1(x), w)$ for every pair $(x,w) \in \mathcal{X} \times [0,1]$ where $Y=\tilde{f}(\eta^{\theta_1}_1(X),W)$  with probability one.  Then, using again the functional characterization in  Theorem \ref{th_representation}, this means that $\mu_{X,Y}\in \mathcal{P}_{\eta^{\theta_1}_1 }(\mathcal{X}\times \mathcal{Y})$. In other words, $\mathcal{P}_{\eta^{\theta_2}_2 \circ \eta^{\theta_1}_1 }(\mathcal{X}\times \mathcal{Y}) \subset \mathcal{P}_{\eta^{\theta_1}_1}(\mathcal{X}\times \mathcal{Y})$. We can apply the same argument recursively to conclude that: 
$$\mathcal{P}_{\eta^{\theta_K}_K \circ \ldots \circ \eta^{\theta_1}_1}(\mathcal{X}\times \mathcal{Y})  \subset \ldots  \subset \mathcal{P}_{\eta^{\theta_2}_2 \circ \eta^{\theta_1}_1 }(\mathcal{X}\times \mathcal{Y})   \subset   \mathcal{P}_{\eta^{\theta_1}_1}(\mathcal{X}\times \mathcal{Y}) \subset \mathcal{P}(\mathcal{X}\times \mathcal{Y}).$$

For the second part of the result, let us consider $j\in  \left\{2,\ldots,K\right\}$.  The IP task after the application of the first $j$-layer of processing reduces to 
\begin{equation}\label{eq_proof_pro_multilayer_info_loss_1}
	I(X;Y|U_j) = \min_{\tilde{\mu} \in \mathcal{P}_{\eta^{\theta_j}_j \circ \ldots \circ \eta^{\theta_1}_1}(\mathcal{X} \times \mathcal{Y})} D(\mu_{X,Y} || \tilde{\mu}), 
\end{equation}
where $U_j=\eta^{\theta_j}_j(\cdot) \circ \ldots \circ \eta^{\theta_1}_1(X)$.  
At this stage, we know that $\mathcal{P}_{\eta^{\theta_j}_j \circ \ldots \circ \eta^{\theta_1}_1}(\mathcal{X} \times \mathcal{Y}) \subset \mathcal{P}_{\eta^{\theta_{j-1}}_{j-1} \circ \ldots \circ \eta^{\theta_1}_1}(\mathcal{X} \times \mathcal{Y})$, then we have from (\ref{eq_proof_pro_multilayer_info_loss_1}) that $I(X;Y|U_j) \geq I(X;Y|U_{j-1})$. In addition, by the construction of the multilayer processing setting we know that $U_j=\eta^{\theta_j}_{j}(U_{j-1})$. This implies that
\begin{align}
	\label{eq_proof_pro_multilayer_info_loss_2}
	I(X;Y|U_j) 	&= I((X,U_{j-1});Y|U_j)\\ 
	\label{eq_proof_pro_multilayer_info_loss_2b}
			&= I(U_{j-1};Y|U_j) + I(X;Y|U_j,U_{j-1})\\ 
	\label{eq_proof_pro_multilayer_info_loss_2c}
			&= I(U_{j-1};Y|U_j) + I(X;Y|U_{j-1}), 
\end{align}
where (\ref{eq_proof_pro_multilayer_info_loss_2}) derives from the chain-rule of MI and the observation that $I(U_{j-1};Y|U_j,X)=0$  (as $U_{j-1}$ is a deterministic function of $X$), the equality in (\ref{eq_proof_pro_multilayer_info_loss_2b}) comes from the chain-rule of MI and (\ref{eq_proof_pro_multilayer_info_loss_2c}) comes from the observation that $I(X;Y|U_j,U_{j-1})=I(X;Y|U_{j-1})$ (as $U_j$ is a deterministic function of $U_{j-1}$). Using IP identity presented in (\ref{eq_proof_pro_multilayer_info_loss_1}) in the additive decomposition in (\ref{eq_proof_pro_multilayer_info_loss_2c}), it follows that 
\begin{equation}\label{eq_proof_pro_multilayer_info_loss_3}
\min_{\tilde{\mu} \in \mathcal{P}_{\eta^{\theta_j}_j \circ \ldots \circ \eta^{\theta_1}_1}(\mathcal{X} \times \mathcal{Y})} D(\mu_{X,Y} || \tilde{\mu}) =   \min_{\tilde{\mu} \in \mathcal{P}_{\eta^{\theta_{j-1}}_{j-1} \circ \ldots \circ \eta^{\theta_1}_1}(\mathcal{X} \times \mathcal{Y})} D(\mu_{X,Y} || \tilde{\mu}) + \underbrace{I(U_{j-1};Y|U_{j})}_{\geq 0}.
\end{equation}
Finally, from the proof of Theorem \ref{th_if_projection_aanlogy}, we know that $\tilde{\mu}^j_{X,Y}= \mu_X \cdot \mu_{Y|U_j}(\cdot|\cdot) \in \mathcal{P}_{\eta^{\theta_j}_j \circ \ldots \circ \eta^{\theta_1}_1}(\mathcal{X} \times \mathcal{Y})$ and $\tilde{\mu}^{j-1}_{X,Y}=\mu_X \cdot \mu_{Y|U_{j-1}}(\cdot|\cdot) \in \mathcal{P}_{\eta^{\theta_{j-1}}_{j-1} \circ \ldots \circ \eta^{\theta_1}_1}(\mathcal{X} \times \mathcal{Y})$ are the optimal solutions of the two IP problems in (\ref{eq_proof_pro_multilayer_info_loss_3}). 
\end{proof}

\section{Proof of Theorem \ref{th_consistency}}
\label{proof_th_consistency}
\begin{proof}
The learning rule $\xi_n(\cdot)$ of data length $n>0$ is a mapping from $S_n\in (\mathcal{X}\times \mathcal{Y})^n$ to $(\bar{\theta}(S_n), \phi(S_n)) \in \Theta \times \Psi$. In this context, we have the collection of encoders $\left\{\eta_{\bar{\theta}}(\cdot): \mathcal{X} \rightarrow \mathcal{U}, \bar{\theta} \in \Theta  \right\}$  and the collection of 
soft decoders (conditional distributions) $\left\{v^\phi(\cdot|\cdot), \phi \in \Psi \right\}\subset \mathcal{P}(\mathcal{Y}|\mathcal{U})$.\footnote{Using the functional characterization in Theorem \ref{th_representation}, we might consider that $v^\phi(\cdot| u)$ is the distribution induced by the relationship $Y=f_\phi(W,u)$ for all $u\in \mathcal{U}$, where $f_\phi(\cdot,\cdot)$ is a collection of parametric functions (deep learning) from $[0.1] \times \mathcal{U}$ to $\mathcal{Y}$.} With these two elements, the hypothesis space (with an encoder-decoder structure)  is: 
\begin{equation}\label{eq_proof_th_consistency_1}
	\mathcal{H}_{\Theta,\Psi} \equiv \left\{ v^{\bar{\theta}, \phi}(\cdot|\cdot) = v^{\phi}(\cdot|\eta_{\bar{\theta}}(\cdot)), \bar{\theta} \in \Theta, \phi \in \Psi \right\} \subset \mathcal{P}(\mathcal{Y}|\mathcal{X}).
\end{equation}

Let us consider an arbitrary pair of points $(\bar{\theta},\psi) \in \Theta \times \Psi$ in our decision space. Then for the induced predictive model in $v^{\bar{\theta}, \phi}(\cdot|\cdot) \in \mathcal{H}_{\Theta,\Psi}$, we have that: 
\begin{align}\label{eq_proof_th_consistency_2}
	r(v^{\bar{\theta}, \phi}(\cdot|\cdot), \mu_{X,Y}) 
			&= \mathbb{E}_{(X,Y)\sim \mu_{X,Y}} \left\{- \log v^{\bar{\theta}, \phi}(Y|X)  \right\}\\
			&=  \mathbb{E}_{(X,Y)\sim \mu_{X,Y}} \left\{- \log v^{\phi}(Y|\eta_{\bar{\theta}}(X))  \right\} \\
			\label{eq_proof_th_consistency_2_last}
			&= \mathbb{E}_{(U_{\bar{\theta}},Y)\sim \mu_{U_{\bar{\theta}},Y}} \left\{- \log v^{\phi}(Y| U_{\bar{\theta}})  \right\}, 
\end{align}
where looking at (\ref{eq_p_th_main_IS_mismatch_1}) and using the derivations in (\ref{eq_p_th_main_IS_mismatch_2}), it follows that  
\begin{align}\label{eq_proof_th_consistency_2b}
\mathbb{E}_{(U_{\bar{\theta}},Y)} \left\{- \log v^{\phi}(Y| U_{\bar{\theta}})  \right\}= D(\mu_{Y|U_{\bar{\theta}}}(\cdot |\cdot)|| v^{\phi}(\cdot|\cdot) |\mu_{U_{\bar{\theta}}}) + H(Y|U_{\theta}).  
\end{align}
Then, from the derivation used in (\ref{eq_p_th_main_IS_mismatch_4}), (\ref{eq_proof_th_consistency_2_last}) can be expressed as: 
\begin{align}\label{eq_proof_th_consistency_3}
	r(v^{\bar{\theta}, \phi}(\cdot|\cdot), \mu_{X,Y}) = D(\mu_{Y|U_{\bar{\theta}}}(\cdot |\cdot)|| v^{\phi}(\cdot|\cdot) |\mu_{U_{\bar{\theta}}}) + I(X;Y|U_{\bar{\theta}}) + H(Y|X).
\end{align}

On the other hand, for any predictive model $v(\cdot|\cdot)  \in \mathcal{P}(\mathcal{Y}|\mathcal{X})$, we have that (from the same derivations presented in (\ref{eq_p_th_main_IS_mismatch_2}))
\begin{align}\label{eq_proof_th_consistency_4}
	r(v(\cdot|\cdot), \mu_{X,Y}) = D(\mu_{Y|X}(\cdot |\cdot)|| v(\cdot|\cdot) |\mu_{X}) + H(Y|X), 
\end{align}
which is particularly true for  any of our encoder-decoder models $v^{\bar{\theta}, \phi}(\cdot|\cdot) \in \mathcal{H}_{\Theta,\Psi}$. From (\ref{eq_proof_th_consistency_3}) and (\ref{eq_proof_th_consistency_4}), we have the following  additive decomposition: for any $(\bar{\theta},\psi) \in \Theta \times \Psi$ 
\begin{align}\label{eq_proof_th_consistency_5}
	D(\mu_{Y|X}(\cdot |\cdot)|| v^{\bar{\theta}, \phi}(\cdot|\cdot)   |\mu_{X}) = \underbrace{I(X;Y|U_{\bar{\theta}})}_{\text{encoder $\eta_{\bar{\theta}}(\cdot)$ regret}} + \underbrace{D(\mu_{Y|U_{\bar{\theta}}}(\cdot |\cdot)|| v^{\phi}(\cdot|\cdot) |\mu_{U_{\bar{\theta}}})}_{\text{decoder $v^{\phi}(\cdot|\cdot)$ regret}},
\end{align}
where each term on the RHS of (\ref{eq_proof_th_consistency_5}) (associated to the individual role of the encoder and decoder) is non-negative. 

Let us consider a learning scheme $\Xi=\left\{ \xi_n(\cdot), n\geq 1 \right\}$ driven by the empirical i.i.d. process $(X_i,Y_i)_{i\geq 1}$ where $(X_i,Y_i) \sim \mu_{X,Y}$ and $\mu_{X,Y}\in \mathcal{P}(\mathcal{X} \times \mathcal{Y})$ is an arbitrary data-generated model. Asking for consistency, in the sense introduced in Def. \ref{cross_entropy_consistency}, means that 
\begin{align}\label{eq_proof_th_consistency_6}
	\lim_{n \rightarrow \infty} r(v^{\bar{\theta}(S_n), \phi(S_n)}(\cdot|\cdot), \mu_{X,Y})= 
	H(Y|X), 
\end{align}
where $(\bar{\theta}(S_n), \phi(S_n))=\xi_n(S_n)$.  From the equality in (\ref{eq_proof_th_consistency_4}), (\ref{eq_proof_th_consistency_6}) is equivalent to the condition that
\begin{align}\label{eq_proof_th_consistency_7}
	\lim_{n \rightarrow \infty} D(\mu_{Y|X}(\cdot |\cdot)|| v^{\bar{\theta}(S_n),\phi(S_n)}(\cdot|\cdot) |\mu_X)=0,
\end{align}
a.s. w.r.t. process distribution of $(X_i,Y_i)_{i\geq 1}$. This proves the first part of the result in Eq.(\ref{eq_sec_ce_learning_2_pre}).

Furthermore, using the additive encoder-decoder decomposition in (\ref{eq_proof_th_consistency_5}), we have that for any $n\geq 1$ 
\begin{align}\label{eq_proof_th_consistency_8}
	D(\mu_{Y|X}(\cdot |\cdot)|| v^{\bar{\theta}(S_n),\phi(S_n)}(\cdot|\cdot) |\mu_X) = I(X;Y| \eta_{\bar{\theta}(S_n)}(X)) + D(\mu_{Y|U_n}(\cdot |\cdot)|| v^{\phi(S_n)}(\cdot|\cdot) |\mu_{U_n}), 
\end{align}
where $U_n=\eta_{\bar{\theta}(S_n)}(X)$ denotes the (data-driven) representation with distribution denoted by  $\mu_{U_n}$ (which is obtained from $\mu_X$ and  the encoder $\eta_{\bar{\theta}(S_n)}(\cdot)$). It is worth pointing out that both expressions in the RHS of (\ref{eq_proof_th_consistency_8}) are non-negative and functions of $S_n$ (i.e., random variables). In light of this observation, achieving consistency in the sense of the convergence result in (\ref{eq_proof_th_consistency_7}) is equivalent to asking that: 
\begin{itemize}
	\item $\lim_{n \rightarrow \infty} I(X;Y| \eta_{\bar{\theta}(S_n)}(X))=0$ and 
	\item $\lim_{n \rightarrow \infty} D(\mu_{Y|U_n}(\cdot |\cdot)|| v^{\phi(S_n)}(\cdot|\cdot) |\mu_{U_n})=0$, 
\end{itemize}
a.s. w.r.t. process distribution of $(X_i,Y_i)_{i\geq 1}$. This concludes the second part of the proof. 
\end{proof}

\section{Supporting Results}
\label{app_supporting}
\subsection{Proof of Lemma \ref{th_loss_it_bound}}
\label{proof_th_loss_it_bound}
\begin{proof} 
Let us consider $\mu_{X,Y}\in \mathcal{P}(\mathcal{X}\times \mathcal{Y})$ and $v_{\tilde{Y}|\tilde{X}}(\cdot|\cdot) \in \mathcal{P}(\mathcal{Y}|\mathcal{X})$.  From (\ref{eq_sec_mis_1}), it follows that 
\begin{align} \label{eq_th_loss_it_bound_1}
	r(v_{\tilde{Y}|\tilde{X}}(\cdot|\cdot),\mu_{X,Y}) 
	&= - \mathbb{E}_{(X,Y)\sim \mu_{X,Y}} \left\{ \log v_{\tilde{Y}|\tilde{X}}(Y|X) \right\}\nonumber\\
	&= \mathbb{E}_{(X,Y)}  \left\{\log \frac{\mu_{Y|X}(Y|X)}{v_{\tilde{Y}|\tilde{X}}(Y|X)} \right\} + \mathbb{E}_{(X,Y)} \left\{ \log \frac{1}{\mu_{Y|X}(Y|X) } \right\}\nonumber\\
	&= D(\mu_{Y|X}(\cdot|\cdot) || v_{\tilde{Y}|\tilde{X}}(\cdot| \cdot) |\mu_X) + H(Y|X), 
\end{align}
where the expected value is using that $(X,Y)\sim \mu_{X,Y}$, $\mu_X$ is the marginal of $X$ and 
\begin{align} \label{eq_th_loss_it_bound_2}
	D(\mu_{Y|X}(\cdot| \cdot)|| v_{\tilde{Y}|\tilde{X}}(\cdot| \cdot ) |\mu_X) \equiv \mathbb{E}_{X\sim \mu_X}  \left\{ D(\mu_{Y|X}(\cdot|X)|| v_{\tilde{Y}|\tilde{X}}(\cdot|X)) \right\}. 
\end{align}

It is known that $D(p||q) \geq 0$ for any pair  $p,q \in \mathcal{P}(\mathcal{Y})$ and, consequently,  $D(\mu_{Y|X}(\cdot|X)|| v_{\tilde{Y}|\tilde{X}}(\cdot|X) |\mu_X) \geq 0$ from (\ref{eq_th_loss_it_bound_2}). Using (\ref{eq_th_loss_it_bound_1}), this implies that $r(v_{\tilde{Y}|\tilde{X}}(\cdot|\cdot),\mu_{X,Y}) \geq H(Y|X)$.

It is also known that $D(p||q)=0$ if, and only if, $p=q$ in the total variational distance sense \cite{cover_2006,gray_1990_b}. On the other hand,  $\mathbb{E}_{X\sim \mu_X}  \left\{ D(\mu_{Y|X}(\cdot|X)|| v_{\tilde{Y}|\tilde{X}}(\cdot|X)) \right\}=0$ if, and only if, the argument of the integration (function of $X$) is $0$ for $\mu_X$ almost every point in $\mathcal{X}$ \cite{halmos_1950}.  Integrating this result in  (\ref{eq_th_loss_it_bound_1}),  $r(v_{\tilde{Y}|\tilde{X}}(\cdot|\cdot),\mu_{X,Y}) = H(Y|X)$ is equivalent to $D(\mu_{Y|X}(\cdot|X) || v_{\tilde{Y}|\tilde{X}}(\cdot|X) |\mu_X)=0$ that is equivalent to the condition that  $v_{\tilde{Y}|\tilde{X}}(\cdot|X)= \mu_{Y|X}(\cdot|X)$ in total variation for $\mu_X$ almost every point\footnote{This means formally that: $\mu_X \left( \left\{x\in \mathcal{X},  v_{\tilde{Y}|\tilde{X}}(\cdot|x)= \mu_{Y|X}(\cdot|x) \right\} \right)=1$.}. This is  the statement in (\ref{eq_sec_mis_3}).
\end{proof}

\subsection{Proof of Lemma \ref{lemma_functional_expressiviness}}
\label{proof_lemma_functional_expressiviness}
\begin{proof}
	We know from Theorem \ref{th_representation} that for any model $\mu_{X,Y}\in \mathcal{P}_\eta(\mathcal{X}\times \mathcal{Y})$, there exists a measurable function $f:[0,1] \times \mathcal{U} \rightarrow \mathcal{Y}$  such that the conditional distribution $Y|X=x$ (for $\mu_X$-almost every point) follows from the following functional expression: 
	\begin{equation}\label{eq_proof_lemma_functional_expressiviness_1}
		Y=f(W,\eta(x)),
	\end{equation}
	where $W \sim Unif[0,1]$. Then,  (\ref{eq_proof_lemma_functional_expressiviness_1}) tells us that $\mu_X$-almost every point in $\mathcal{X}$
	\begin{equation}\label{eq_proof_lemma_functional_expressiviness_2}
		\mu_{Y|U}(B|\eta(x))= \mathbb{P}(Y\in B|U=\eta(x))= \mathbb{P}(f(W,\eta(x))\in B),\  \forall B\subset \mathcal{Y}.
	\end{equation}
	On the other hand, by the hypothesis,  $\exists \theta \in \Theta$ such that $\forall u\in \mathcal{U}$
	\begin{equation}\label{eq_proof_lemma_functional_expressiviness_3}
		f_\theta(W,u)= f(W,u), \text{ $\mu_W$-almost surely,}
	\end{equation}
	which means that  for all $u\in \mathcal{U}$ \cite{gray_2009}
	\begin{equation}\label{eq_proof_lemma_functional_expressiviness_4}
		\mathbb{P}(f_\theta(W,u)\in B) = \mathbb{P}(f(W,u)\in B).
	\end{equation}
	Integrating, we have that for $\mu_X$-almost every point $x\in \mathcal{X}$
	\begin{align}\label{eq_proof_lemma_functional_expressiviness_5}
		\mu_{Y|U}(B|\eta(x))	&= \mathbb{P}(f(W,\eta(x))\in B)\nonumber\\
						&= \mathbb{P}(f_\theta(W,\eta(x))\in B) \nonumber\\
						&= v^\theta_{\tilde{Y}|U}(B|\eta(x)), \text{ for all $B\subset \mathcal{Y}$},
	\end{align}
	from the observation that $\eta{(x)}\in \mathcal{U}$ and the functional construction of $v^\theta_{\tilde{Y}|U}(\cdot|\cdot)\in \Lambda_{\Theta,\eta}$. Then,  $\mu_{Y|U}(\cdot|\eta(X)) = v^\theta_{\tilde{Y}|U}(\cdot |\eta(X))$ in total variation, $\mu_X$-almost surely. 
\end{proof}

\subsection{Proof of Lemma \ref{lm:expresive_VQ}}
\label{proof_lm:expresive_VQ}
The proof of this result follows from Silva {et al.} \cite[Th.15]{silva_2021}, the construction presented by Liese et al. \cite{liese_2006} and the IP analogy in Theorem \ref{th_if_projection_aanlogy}. We begin introducing the new ingredients: 
\begin{theorem}(\cite[Th.15]{silva_2021})
\label{th_suf_IS_condition_for_embedded_VQ}
	Let $\left\{\eta_i(\cdot), i\geq 1 \right\}$ be a collection of finite-size VQs equipped with its induced finite-size measurable partitions $\left\{\pi_{i}, i\geq 1 \right\}$. 
	If the collection is embedded in the sense that\footnote{$\sigma(\eta)$ is the smallest sigma that makes $\eta(\cdot)$ a measurable function \cite{varadhan_2001}.} $\sigma(\eta_1) \subset \sigma(\eta_2) \subset \sigma(\eta_3) \ldots$ and $\sigma( \pi_{1} \cup \pi_{2} \ldots )= \mathcal{B}(\mathbb{R}^d)$ then for any distribution $\mu_{X,Y} \in \mathcal{P}(\mathcal{X} \times \mathcal{Y})$
	\begin{equation}\label{eq_proof_expresive_VQ_1}
		\lim_{i \rightarrow \infty} I(X;Y|\eta_i(X))=0.
	\end{equation} 
\end{theorem}

Liese et al. \cite{liese_2006} propose an embedded collection of measurable partitions of $\mathcal{X}=\mathbb{R}^d$. 
The indexed construction is the following:
\begin{equation}\label{eq_proof_expresive_VQ_2}
	\tilde{\pi}_m=  \left\{ B_{m,0} \right\} \cup \left\{B_{m,\bar{j}}, \bar{j}=(j_1,\ldots,j_d) \in \mathcal{J}_m \right\} \subset \mathcal{B}(\mathbb{R}^d),
\end{equation} 
where 
$\mathcal{J}_m = \left\{-m 2^m,\ldots,m2^m-1\right\}^d$ and 
\begin{align}\label{eq_proof_expresive_VQ_3}
	B_{m,0}&= \mathbb{R}^d \setminus [-m,m)^d, \\ 
	\label{eq_subsec_inf_divergence_17c2}
	B_{m,j_1,...,j_d} &= \bigotimes^d_{k=1}  \left[\frac{j_k}{2^m},\frac{j_{k}+1}{2^m} \right), \  \forall (j_1,\ldots,j_d)\in \mathcal{J}_m.
\end{align}
\begin{proof}
Liese et al.  \cite{liese_2006} prove that the collection of embedded 
 partitions $ \left\{ \tilde{\pi}_m, m\geq 1 \right\} $ in (\ref{eq_proof_expresive_VQ_2}) is universal for 
 $\mathcal{B}(\mathbb{R}^d)$, in the sense that any interval in $\mathcal{B}(\mathbb{R}^d)$ can be approximated (arbitrarily closely) by the union  of cells of $\tilde{\pi}_m$ as $m$ goes to infinity.  Consequently,  we have that $\sigma(\cup_{m\geq 1} \tilde{\pi}_m)=\mathcal{B}(\mathbb{R}^d)$ \cite{liese_2006}, which implies from Theorem \ref{th_suf_IS_condition_for_embedded_VQ} that 
	 \begin{equation}\label{eq_proof_expresive_VQ_4}
		\lim_{m \rightarrow \infty} I(X;Y|\eta_m(X))=0.
	\end{equation} 
	for any model $\mu_{X,Y}\in \mathcal{P}(\mathcal{X}\times \mathcal{Y})$.
	
	At this point the IP analogy, presented in Theorem \ref{th_if_projection_aanlogy},  tells us that for any finite-size partition $\pi$
	(and its induced VQ $\eta_\pi(\cdot)$) and for any model $\mu_{X,Y}$: 
	 \begin{equation}\label{eq_proof_expresive_VQ_5}
		I(X;Y|\eta_\pi(X)) = \min_{\tilde{\mu} \in \mathcal{P}_{\pi}(\mathcal{X} \times \mathcal{Y})} D(\mu_{X,Y} || \tilde{\mu}).
	\end{equation} 
	Then, (\ref{eq_proof_expresive_VQ_4}) and (\ref{eq_proof_expresive_VQ_5}) mean that for any $\epsilon>0$ and model $\mu_{X,Y}$, there is $m$ such that 
	\begin{equation}\label{eq_proof_expresive_VQ_5b}
	I(X;Y|\eta_m(X)) = \min_{\tilde{\mu} \in \mathcal{P}_{\pi_m}(\mathcal{X} \times \mathcal{Y})} D(\mu_{X,Y} || \tilde{\mu}))<\epsilon.
	\end{equation}  
	Considering that $\pi_m\in \mathcal{Q}(\mathcal{X})$ (the collection of finite-size measurable partitions), from (\ref{eq_proof_expresive_VQ_5b}) we have that for any $\epsilon>0$: 
	 \begin{equation}\label{eq_proof_expresive_VQ_6}
 		\min_{\tilde{\mu} \in \bigcup_{\pi \in \mathcal{Q}(\mathcal{X})} \mathcal{P}_{\pi}(\mathcal{X} \times \mathcal{Y})} D(\mu_{X,Y} || \tilde{\mu}) <\epsilon.
	\end{equation} 
	To conclude, the upper bound in (\ref{eq_proof_expresive_VQ_6}) is distribution-free and valid for an arbitrary small $\epsilon>0$, which proves the result in (\ref{eq_expresive_VQ_1}). 
\end{proof}

\subsection{Proof of Lemma \ref{lm:IB_expressive}}
\label{proof_lm:IB_expressive}
\begin{proof}
This result follows directly from the constructions presented in (\ref{eq_proof_expresive_VQ_2}) and (\ref{eq_proof_expresive_VQ_4}).
Let us consider the encoder induced by $\tilde{\pi}_m$: 
 \begin{equation}\label{eq_proof_IB_expressive_1}
	 	\tilde{\eta}_{\tilde{\pi}_m}(x) \equiv  f^m(m2^m,...,m2^m) \cdot {\bf 1}_{B_{m,0}}(x) + \sum_{\bar{j}  \in \mathcal{J}_m}   f^m(\bar{j}) \cdot {\bf 1}_{B_{m,\bar{j}}}(x) \in \mathbb{R}, 
\end{equation} 
where $f^m: \left\{ (m2^m,...,m2^m) \right\} \cup \mathcal{J}_m \rightarrow \left\{1,\ldots, (m2^{m+1})^d+1\right\} \subset \mathbb{R}$ is an injective scalar function.  As $\tilde{\eta}_{\tilde{\pi}_m}(\cdot)$ is a deterministic and finite-size (discrete) mapping, we have that: 
	 \begin{equation}\label{eq_proof_IB_expressive_1}
	 	I(X;\tilde{\eta}_{\tilde{\pi}_m}(X)) \leq H(\tilde{\eta}_{\tilde{\pi}_m}(X)) \leq  \underbrace{\log_2( (m2^{m+1})^d+1)}_{A_m \equiv}< \infty
	\end{equation}
	Therefore $ \left\{\tilde{\eta}_{\tilde{\pi}_m}(\cdot), m\geq 1 \right\}$ belongs to the class of finite-entropy mappings. Indeed, using the result in (\ref{eq_proof_expresive_VQ_5b}), it follows that for any $\epsilon>0$, $\exists m>0$ such that for any $B \geq A_m$: 
	 \begin{equation}\label{eq_proof_IB_expressive_2}
	 	I(X;Y|\eta^B(X)) \leq I(X;Y| \tilde{\eta}_m(X)) < \epsilon. 
	\end{equation}
	The first inequality in (\ref{eq_proof_IB_expressive_2}) comes from the definition of IB solution,  $\eta^B(\cdot)$, and the fact that $\tilde{\eta}_m(\cdot)$ meets the bandwidth constraint when  $B \geq A_m$. The second inequality in (\ref{eq_proof_IB_expressive_2}) comes from the result in  (\ref{eq_proof_expresive_VQ_5b}).  As $m$ in  (\ref{eq_proof_expresive_VQ_5b}) is function of $\epsilon$ and $\mu_{X,Y}$, we can consider $B(\epsilon, \mu_{X,Y}) \equiv A_{m}$, which concludes the argument.  
\end{proof}

\subsection{Proof Proposition \ref{pro_orbit}}
\label{proof_pro_orbit}
\begin{proof}	
	Let us consider an arbitrary $A\subset \mathcal{Y}$. 
	We want to show that $\mu_{Y|X}(A |  \left\{ x \right\})= \mu_{Y|X}(A\,|\,\text{orbit}(x))$ $\mu_X$-almost surely.
	 
	Without loss of generality,  let us assume that  $\mu_X(\text{orbit}(x))>0$. Then 
	\begin{equation}\label{eq_proof_pro_orbit_1}
		\mu_{Y|X}(A\,|\,\text{orbit}(x)) = \frac{\mu_{Y,X}(A \times \text{orbit}(x))}{\mu_{X}(\text{orbit}(x))}.
	\end{equation}
	For simplicity, let us assume that $X$ is equipped with a probability density function $f_X(\cdot)$.   
	It is simple to show that $\mu_{Y,X} ( \left\{ i \right\} \times \text{orbit}(x)) = \int_{\text{orbit}({x})} f_i(\tilde{x})\:\mathrm{d} \tilde{x} $ where $f_i(x)= \mu_{Y|X}(i|x) \cdot f_X(x),\forall x\in\mathcal{X}$. This implies that: 
	\begin{equation}\label{eq_proof_pro_orbit_2}
		\mu_{Y,X}(A \times \text{orbit}(x)) = \sum_{i\in A} \int_{\text{orbit}(x)} f_i(\tilde{x})\:\mathrm{d} \tilde{x} = \sum_{i\in A} \int_{\text{orbit}(x)} \mu_{Y|X}(i|\tilde{x}) \cdot f_X(\tilde{x})\:\mathrm{d} \tilde{x}.
	\end{equation}
	Here, we use the invariant assumption of $\mu_{X,Y}$. Eq.(\ref{eq_proof_lm_invariant_IS_1}) means that for any $\tilde{x}\in \text{orbit}(x)$, $\mu_{Y|X}(\cdot |\tilde{x})= \mu_{Y|X}(\cdot |x)$ for $\mu_X$-almost every point $x\in \mathcal{X}$. 
	Consequently, we have from (\ref{eq_proof_pro_orbit_2}) that for $\mu_X$-almost every $x\in \mathcal{X}$
	\begin{equation}\label{eq_proof_pro_orbit_3}
	\mu_{Y,X}(A \times \text{orbit}(x))= \mu_{Y|X}(A|x) \cdot \mu_{X}(\text{orbit}(x)).
	 \end{equation}
	Using this last expression in (\ref{eq_proof_pro_orbit_1}), we conclude that $\mu_{Y|X}(A\,|\,\text{orbit}(x)) = \mu_{Y|X}(A|x)$, $\mu_X$-almost surely.  
\end{proof}

\subsection{Proposition \ref{prop_invariant_property}}
\label{sec_proof_prop_invariant_property}
\begin{proposition}\label{prop_invariant_property}
	For any $B\in \mathcal{B}(\mathcal{X})$ and $\forall g\in \mathcal{G}$ we have that 
	\begin{equation} \label{eq_ip_1} 	
	 {\eta^*_\mathcal{G}}^{-1}( \eta^*_\mathcal{G}(B) ) =   {\eta^*_\mathcal{G}}^{-1}( \eta^*_\mathcal{G}(g^{-1}(B)) ),
	\end{equation}	 
\end{proposition}
where $\eta^*_\mathcal{G}(A) \equiv  \left\{\eta^*_\mathcal{G}(x), x\in A \right\}$ in (\ref{eq_ip_1}). 
\label{proof_prop_invariant_property}

\begin{proof}	
First, we use the fact that 
\begin{equation}
\text{orbit}(x)=\left\{g(x), g\in \mathcal{G} \right\}={\eta^*_\mathcal{G}}^{-1}( \left\{\eta^*_\mathcal{G}(x) \right\}).  
\end{equation}
Furthermore, for any $B\subset \mathcal{B}(\mathcal{X})$
\begin{equation}
{\eta^*_\mathcal{G}}^{-1}( \eta^*_\mathcal{G}(B) )=\bigcup_{x\in B}  \text{orbit}(x)=\bigcup_{g\in \mathcal{G}}g(B),
\end{equation}
where $g(B)=  \left\{g(x), x\in B \right\}$.  Finally using this last expression and noting that for any $g \in \mathcal{G}$,  $g^{-1} \in \mathcal{G}$ and $\mathcal{G}$ is close under composition (as $\mathcal{G}$ is a compact group) \cite{eaton_1889}, it follows directly that ${\eta^*_\mathcal{G}}^{-1}( \eta^*_\mathcal{G}(B))={\eta^*_\mathcal{G}}^{-1}( \eta^*_\mathcal{G}(g^{-1}(B)))$ for any $g\in \mathcal{G}$, which proves the result in (\ref{eq_ip_1}). 
\end{proof}

\section{More Examples of IS Classes}
\label{sec_IS_ML_algorithms}
Here, we complement Section \ref{sub_sec_is_in_learning} 
with two more examples. 
\subsection{Transform-Based Sparse Models: Linear Projector -- $\eta^U_{j_1,\ldots,j_q}(x)$} 
\label{sub_sec_T_sparse_model}
Let us consider an arbitrary orthonormal basis $U=(\bar{u}_1,...,\bar{u}_d)_{d\times d}$ of $\mathbb{R}^d$.\footnote{$U$ is a unitary matrix composed by a set of linearly independent (column) orthonormal vectors \cite{vetterli_1995,goyal_2001}. Consequently,  $U \cdot  U^{\mathsf{T}}=I_{d\times d}$, where $I_{d\times d}$ denotes the identity matrix. } In this case, we can consider the projection operator in the transform domain induced by the basis matrix $U$.  For that let us consider the coordinates $j_1<j_2<...<j_q\in \left\{1,\ldots,d \right\}$ (with $q<d$).  Then, the projection operator $\eta^U_{j_1,\ldots,j_q}: \mathbb{R}^d \rightarrow  \mathbb{R}^q$ is given by $\eta^U_{j_1,\ldots,j_q}(\bar{x})=(\bar{u}_{j_1}^{\mathsf{T}}\cdot \bar{x},\ldots,\bar{u}_{j_q}^{\mathsf{T}} \cdot \bar{x}) \in \mathbb{R}^q,\forall\bar{x}\in\mathbb{R}^d$. In other words, $\eta^U_{j_1,\ldots,j_q}(\cdot)$ is a linear operator given by $\eta^U_{j_1,\ldots,j_q}(\bar{x})=P_{j_1,\ldots,j_{q}} \cdot U^{\mathsf{T}}\cdot \bar{x}$. 
We recognize two parts in this linear processing; the part associated to $\bar{z}=U^{\mathsf{T}}\cdot \bar{x}$ that is a change of basis (rotation of the space), and the part associated to $P_{j_1,\ldots,j_{q}} \cdot \bar{z}$, which is a coordinate-wise projection in the transform domain of $U$ (introduced in Section \ref{sub_sec_sparse_canonical}).  
Emblematic cases of unitary matrices $U$ are the Discrete Fourier Transform (DFT), the Discrete Cosine Transform (DCT), the Discrete Wavelet transform (DWT), and many others \cite{vetterli_1995,proakis_1996}. Then, the family of $q$-sparse models in the components $j_1,\ldots,j_q\in \left\{1,\ldots,d \right\}$ of the transform domain $U$ is given by: 
\begin{equation} \label{eq_ps_14}
\mathcal{P}^U_{\eta_{j_1,\ldots,j_q}}(\mathcal{X} \times \mathcal{Y}) \equiv  \left\{\mu_{X,Y} \in  \mathcal{P}(\mathcal{X} \times \mathcal{Y})\text{, s.t., }I({X;Y}) = {I}(\eta^U_{j_1,\ldots,j_q}{(X);Y}) \right\}. 
\end{equation}  
Alternatively, if $\mu_{X,Y} \in \mathcal{P}^U_{\eta_{j_1,\ldots,j_q}}(\mathcal{X} \times \mathcal{Y})$ then we have that $I(X;Y)= I(Z_{j_1},Z_{j_2},...,Z_{j_q};Y)$, where $Z_j \equiv  \bar{u}_j^{\mathsf{T}}\cdot X$ for any $j\in \left\{1,\ldots,d \right\}$.  Consequently, projecting $X$ in the $q$-dimensional linear sub-space of $\mathcal{X}$ induced by the orthonormal vectors $\left\{ \bar{u}_{j_1},...,\bar{u}_{j_q} \right\}$ is IS for $\mu_{X,Y}$. 
From Theorem \ref{th_representation},
we have the following functional description: 
\begin{corollary}\label{cor_transform_sparse_class}
	$\mu_{X,Y}\in \mathcal{P}^U_{\eta_{j_1,\ldots,j_q}}(\mathcal{X} \times \mathcal{Y})$ if, and only if, the distribution of $Y$ given $X=x$ (for $\mu_X$-almost every point) can be obtained by 
	\begin{equation} \label{eq_ps_15}
		Y=f(W,\eta^U_{j_1,\ldots,j_q}(x)),
	\end{equation}  
	where $W\sim \emph{Unif}[0,1]$ and $f:[0,1] \times \mathbb{R}^q \rightarrow \mathcal{Y}$ is a measurable function. 
\end{corollary}

\begin{figure*}
  \centering
    \includegraphics[width=0.92\textwidth]{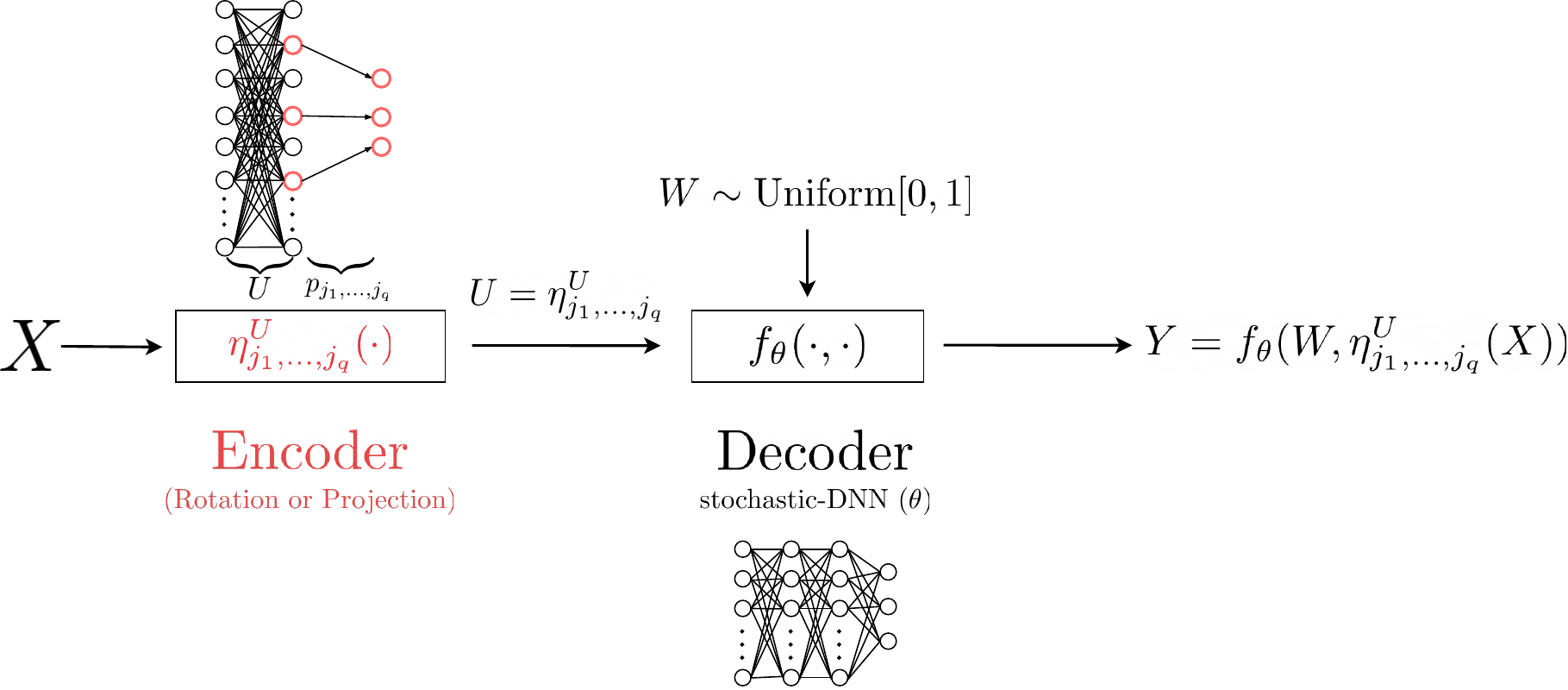}
    \caption{Transform Sparse Neural Network (TS-NN). 
    }
    \label{fig8}
 \end{figure*}

{\bf Transform Sparse Neural Networks (TS-NN):} 
If we want to learn $\mu_{X,Y}$ within $\mathcal{P}^U_{\eta_{j_1,\ldots,j_q}}(\mathcal{X} \times \mathcal{Y})$ (prior knowledge), 
then the first layer of a TS-NN should encode the linear projection operator $\eta^U_{j_1,\ldots,j_q} (\bar{x}) = P_{j_1,\ldots,j_{q}} \cdot U^{\mathsf{T}}\cdot \bar{x}$. This is a two-stage forward process that involves an orthonormal transformation $U^{\mathsf{T}}$ (a fully connected network $d-d$) and then a projection $P_{j_1,\ldots,j_{q}}$  (a point-to-point network $d-q$ that can be interpreted as a pooling operator). 
This encoder-decoder architecture is illustrated in Fig. \ref{fig8}. 
These two layers of pre-processing make a network expressive and fully consistent with the functional structure of $\mathcal{P}^U_{\eta_{j_1,\ldots,j_q}}(\mathcal{X} \times \mathcal{Y})$ in (\ref{eq_ps_15}). 

\subsection{Permutation-invariant models: The Empirical Distribution -- $\mathcal{M}(x)$}
\label{sub_sec_permutation_invariant}
An important example of predictive invariant models (see Def.~\ref{def_strong_invariances}) 
with a concrete  IS encoder is the class invariant to permutations in $\mathcal{X}=\mathbb{R}^d$ \cite{Bloem_2019}.\footnote{This class of models was systematically studied in \cite{Bloem_2019}.} In this case, the compact group $\mathcal{G}$ is denoted by $\mathbb{S}_d$ where for any $g\in \mathbb{S}_d$ there is a permutation of $[d]\equiv\left\{1,\ldots,d\right\}$, $p:[d] \rightarrow [d]$
such that $g({\bf x})=(x_{p(1)},x_{p(2)},\ldots,x_{p(d)})$,  $\forall {\bf x} \in \mathbb{R}^d$. Therefore, if a model $\mu_{X,Y}$ is predictive  
$\mathbb{S}_d$-invariant (see Def.~\ref{def_strong_invariances}), 
it means that its posterior distribution $\mu_{Y|X}(\cdot|\cdot)$ is invariant to the action of any permutation of ${\bf x}=(x_1,\ldots,x_d)$. Therefore,  $\mu_{Y|X}(\cdot|{\bf x}) \in \mathcal{P}(\mathcal{Y})$ depends on the set $\left\{x_1,\ldots,x_d\right\} \subset \mathbb{R}$ induced by ${\bf x}=(x_1,\ldots,x_d)$.\footnote{A complete characterization of this family of permutation invariant functions is presented in  
\cite{zaheer_2017} that is revisited and extended for a family of probabilistic models in \cite{Bloem_2019}.} 
For this class, i.e., $\mathcal{P}_{\mathbb{S}_d}(\mathcal{X}\times \mathcal{Y})$, it is well known that the {\em empirical distribution} (or the frequency counts over any measurable set) $\mathcal{M}:\mathbb{R}^d \rightarrow \mathcal{U}= \mathcal{P}(\mathbb{R})$ \footnote{$\mathcal{M}({\bf x}) = \frac{1}{d} \sum_{i=1}^d \delta_{x_i}(\cdot) \in \mathcal{P}(\mathbb{R})$ denotes the empirical distribution induced by ${\bf x}$ in $(\mathbb{R}, \mathcal{B}(\mathbb{R}))$.} is invariant to the actions of $\mathbb{S}_d$, but, more importantly (for the use of  Theorem  \ref{th_representation_invariant}) 
$\mathcal{M}(\cdot)$ is {\em maximal-invariant} for $\mathbb{S}_d$ \cite{Bloem_2019}. 
Then from Theorem  \ref{th_representation_invariant}, 
we have that:
\begin{corollary}\label{cor_functional_pi}
	$\mu_{X,Y}\in \mathcal{P}_{\mathbb{S}_d}(\mathcal{X}\times \mathcal{Y})$ if, and only if, the distribution of $Y$ given $X={\bf x}$ (for $\mu_X$-almost every point ${\bf x}$) can be obtained by the following functional relationship 
	\begin{equation} \label{eq_pi_3}
		Y=f\hspace{-1mm}\left(W, \mathcal{M}({\bf x})= \frac{1}{d} \sum_{i=1}^d \delta_{x_i}(\cdot)\right),
	\end{equation}  
	where $W\sim \emph{Unif}[0,1]$ and $f:[0,1] \times \mathcal{P}(\mathbb{R}) \rightarrow \mathcal{Y}$ is a measurable function. 
\end{corollary}
{\bf Invariant Neural Networks (I-NN)}:
I-NN were formally introduced with some clever architecture constraints in \cite{Bloem_2019} to be consistent with the structure of  
$\mathcal{P}_{\mathbb{S}_d}(\mathcal{X}\times \mathcal{Y})$. 
Many I-NN architectures  are presented  in \cite{Bloem_2019} and references therein.

\section{Empirical Study} 
\label{supp:emp-study}
Here, we present details of our controlled empirical study in Section \ref{sec_numerical}. In particular, we present the adopted models, their construction, information measures computations, ML architecture, and some training details. 

\subsection{Model Description} 
\label{supp:models}
We construct a class of models within the continuous-discrete setting, i.e., $\mathcal{X} = \mathbb{R}^d$ and $\mathcal{Y} = [M]$. Our design is driven by the intention to produce
$\mu_{X,Y}$ with closed-form expression for $I(\eta_{\mathbf{j}}(X);Y)$, where $\eta_{\mathbf{j}}(\cdot)\equiv\eta_{j_1,j_2,\ldots,j_q}(\cdot)$ denotes the feature selector where $\mathbf{j} = (j_\ell)_{\ell=1}^q\in [d]^q$ is a finite strictly-ordered sequence (f.s.o.~sequence).  We denote the set of all possible selections for $\mathbf{j}$ as $\mathcal{J}$. 

Our construction, driven by a histogram-based underlying structure, has the following steps:  
\begin{itemize}
\item The model is built upon a collection of cells 
that induces a measurable partition of $\mathbb{R}^d$. 
For each dimension, indexed by $k\in[d]$, we use an f.s.o.~sequence $\mathbf{a}_k\equiv(a_{k,\ell})_{\ell=0}^{n_k} \in \mathbb{R}^{n_k+1}$ such that $\{(-\infty,a_{k,0}),[a_{k,n_k},\infty)\}\cup\{[a_{k,\ell-1},a_{k,\ell})\}_{\ell=1}^{n_k}\subset \mathcal{B}(\mathbb{R})$ is a partition of $\mathbb{R}$. In this context, $A_{\mathbf{i}}\equiv\bigtimes_{k=1}^d[a_{k,i_k-1},a_{k,i_k})\subset \mathbb{R}^d$ is indexed by $\mathbf{i}\equiv(i_{k})_{k=1}^d\in\mathfrak{I}$ where $\mathfrak{I}\equiv\bigtimes_{k=1}^d[n_k]$ is an index set. 
Then,  $\{A_{\mathbf{i}}\}_{\mathbf{i}\in\mathfrak{I}}$ is an indexed partition of  $\bigtimes_{k=1}^d[a_{k,0},a_{k,n_k})\subset\mathbb{R}^d$, that will be the support of $X$. 
\item The next step is to define a discrete joint distribution on $\mathfrak{I}\times[M]$. In particular, we need to select $p_y,\forall y\in[M]$, such that $\sum_{y=1}^Mp_y=1$, and a conditional probability mass function $p_{\mathbf{i}|y},\forall (\mathbf{i},y)\in\mathfrak{I}\times[M]$, such that $\sum_{\mathbf{i}\in\mathfrak{I}}p_{\mathbf{i}|y}=1,\forall y\in[M]$. 
\item Using the indexed partition $\{A_{\mathbf{i}}\}_{\mathbf{i}\in\mathfrak{I}}$ and its discrete joint model $(p_y\cdot p_{\mathbf{i}|y})_{(\mathbf{i},y)\in \mathfrak{I}\times[M]}$ , we equip $X$ given the event $Y=y$ with a density.  We define a probability density function (pdf) for $X$ conditioned to $Y=y$ by 
\begin{equation}
    \label{eqn:pdf}
    f_{X|Y}(x|y) \equiv \sum_{\mathbf{i}\in\mathfrak{I}}\frac{p_{\mathbf{i}|y}\cdot 1_{A_{\mathbf{i}}}(x)}{\lambda(A_{\mathbf{i}})},\forall(x,y)\in\mathbb{R}^d\times[M],
\end{equation}
where $1_{A_{\mathbf{i}}}:\mathbb{R}^d\rightarrow\{0,1\}$ is the indicator function of $A_{\mathbf{i}}$ and $\lambda$ is the Lebesgue measure in $(\mathbb{R}^d,\mathcal{B}(\mathbb{R}^d))$. Equipped with this pdf, we have that $\forall(\mathbf{i},y)\in\mathfrak{I}\times[M]$
\begin{align}
    \mu_Y(y)  &= p_y,\label{eqn:study-dist-marginal}\\
    \mu_{X|Y}(A_{\mathbf{i}}|y)  &= p_{\mathbf{i}|y}.\label{eqn:study-dist-cond}
\end{align}
\item Finally, $(a_{k,\ell})_{\ell=0}^{n_k}$, $p_y$, and $p_{\mathbf{i}|y}$, for all $(k,y,\mathbf{i})\in[d]\times[M]\times\mathfrak{I}$, fully determine $\mu_{X,Y}$ as shown in (\ref{eqn:study-dist-marginal}) and (\ref{eqn:study-dist-cond}).
\end{itemize}

\subsection{Mutual Information}
\label{mi_computation_sim_model}
Given our model $\mu_{X,Y}$ (constructed from $(a_{k,\ell})_{\ell=0}^{n_k}$, $(p_y)_{y\in \mathcal{Y}}$, and $(p_{\mathbf{i}|y})_{(\mathbf{i},y)\in\mathfrak{I}\times\mathcal{Y}}$, where $\mathbf{i}= (i_{k})_{k=1}^d\in \mathfrak{I}=\bigtimes_{k=1}^d[n_k]$) and a f.s.o. sequence of coordinates $\mathbf{j}=(j_1,..,j_q)\in \mathcal{J}$, 
$I(\eta_{\mathbf{j}}(X);Y)=\mathcal{I}(\mu_{\eta_{\mathbf{j}}(X),Y})$ can be computed in closed-form. 
For this computation, it is convenient to use the following notation:  
$\mathbf{i}_{\mathbf{j}}\equiv(i_{j_{\ell}})_{\ell=1}^{q}$ and  $\mathbf{i}_{\mathbf{j}}^{\mathsf{c}}\equiv (i_{h_{\ell}})_{\ell=1}^{d-q}$, where $(h_{\ell})_{\ell=1}^{d-q}$ is the f.s.o.~sequence such that $\{h_{\ell}\}_{\ell=1}^{d-q}=[d]\setminus\{j_{\ell}\}_{\ell=1}^q$. In addition, we define $\mathfrak{I}_{\mathbf{j}}\equiv\bigtimes_{\ell=1}^q[n_{j_{\ell}}]$ and $\mathfrak{I}_{\mathbf{j}}^{\mathsf{c}}\equiv\bigtimes_{\ell=1}^{d-q}[n_{h_\ell}]$ as the set of possible values that $\mathbf{i}_{\mathbf{j}}$ and $\mathbf{i}_{\mathbf{j}}^{\mathsf{c}}$ can take, respectively. Let us note, that $\forall\mathbf{j}\in\mathcal{J}$ there exists a permutation mapping $\mathfrak{p}_{\mathbf{j}}:\mathfrak{I}_{\mathbf{j}}\times\mathfrak{I}_{\mathbf{j}}^{\mathsf{c}}\rightarrow\mathfrak{I}$ such that $\forall\mathbf{i}\in\mathfrak{I},\mathfrak{p}_{\mathbf{j}}(\mathbf{i}_{\mathbf{j}},\mathbf{i}_{\mathbf{j}}^{\mathsf{c}})=\mathbf{i}$. Finally, $\forall(\mathbf{j},y)\in\mathcal{J}\times[M]$ we define the following $\forall\mathbf{s}\in\mathfrak{I}_{\mathbf{j}}$:
\begin{equation}
    p_{(\mathbf{s}|y)_{\mathbf{j}}} \equiv \sum_{\mathbf{k}\in\mathfrak{I}_{\mathbf{j}}^{\mathsf{c}}}p_{(\mathfrak{p}_{\mathbf{j}}(\mathbf{s},\mathbf{k}))|y}
    \in \mathbb{R}.
\end{equation}
At this point, it is possible to verify that $\forall(\mathbf{i},\mathbf{j},y)\in\mathfrak{I}\times\mathcal{J}\times[M]$
\begin{equation}
     \mu_{\eta_{\mathbf{j}}(X)|Y}(\eta_{\mathbf{j}}(A_{\mathbf{i}})|y)=p_{(\mathbf{i}_{\mathbf{j}}|y)_{\mathbf{j}}},
\end{equation}
where $A_{\mathbf{i}}=\bigtimes_{k=1}^d[a_{k,i_k-1},a_{k,i_k})\subset\mathbb{R}^d$ and, consequently, it follows that $\forall\mathbf{j}\in\mathcal{J}$
\begin{equation}
    I(\eta_{\mathbf{j}}(X);Y) = \sum_{y=1}^m p_y \sum_{\mathbf{s}\in\mathfrak{I}_{\mathbf{j}}}p_{(\mathbf{s}|y)_{\mathbf{j}}} \log \frac{p_{(\mathbf{s}|y)_{\mathbf{j}}}}{\sum_{\ell=1}^m p_{\ell}\cdot p_{(\mathbf{s}|\ell)_{\mathbf{j}}}}.
\end{equation}

\subsection{Model Examples}
\label{sec_model_examples}
Here, we present some examples. The settings i) and ii) are presented for illustration, while setting iii) is presented to introduce setting iv), which is the construction used in our study.
\label{sec:model-examples}
\begin{figure*}[!t]
    \centering
    \subfloat[\label{sfig:singular-2d}2D-Singular]{\includegraphics[width=4.5cm]{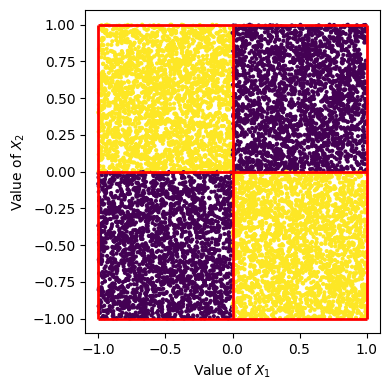}}
    \subfloat[\label{sfig:simple-3d}3D-Equiprobable]{\includegraphics[width=4.5cm]{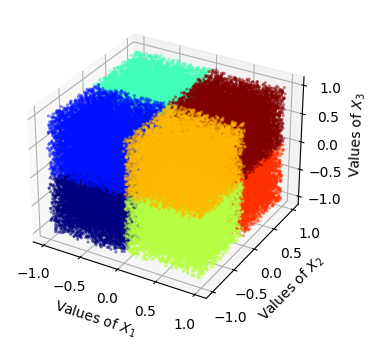}}
    \subfloat[\label{sfig:demo-2d}2D-Demonstration]{\includegraphics[width=4.5cm]{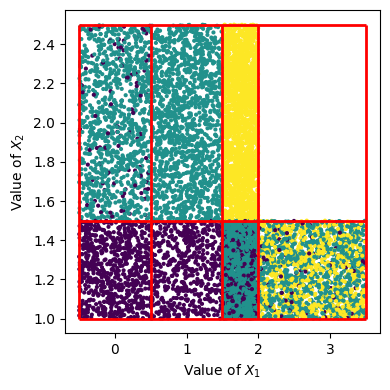}}
    \subfloat[\label{sfig:demo-3d}3D-Demonstration]{\includegraphics[width=4.5cm]{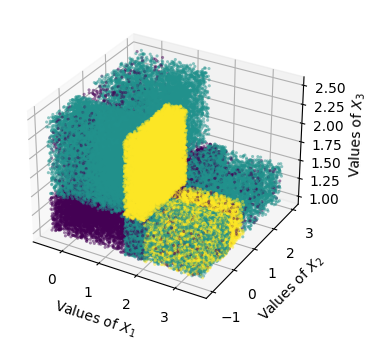}}
    \caption{Visualization of i.i.d. realizations for the four model examples  ($\mu_{X,Y}$) in Appendix \ref{sec_model_examples}. The color indicates the label identity of the sample ($y$).  The red lines in (\ref{sfig:singular-2d}) and (\ref{sfig:demo-2d}) represent the boundaries of the 2D cells ($A_{\mathbf{i}}$).}
    \label{fig:model-demo}
\end{figure*}

\begin{itemize}
    \item[i)] The model 2D-Singular (shown in Fig.~\ref{sfig:singular-2d}) considers $M=d=2$, 
    $p_1=p_2=1/2$, cell-boundary arrays $\mathbf{a}_1 = \mathbf{a}_2 = (-1,0,1)$, and conditional probability for the cell indexes  $p_{(0,0)|1} = p_{(1,1)|1} = p_{(0,1)|2} = p_{(1,0)|2} = 1/2$ and $p_{\mathbf{i}|y}=0$ for $(\mathbf{i},y)\in\mathfrak{I}\times[M]$ not already defined. In this example, $I(\eta_{(1)}(X);Y) = I(\eta_{(2)}(X);Y) = 0$, $I(\eta_{(1,2)}(X);Y) = I(X;Y) = 1\:\text{bit}$, and $H(Y)=1\:\text{bit}$.

    \item[ii)] The model 3D-Equiprobable (shown in Fig.~\ref{sfig:simple-3d}) considers $M=8$, $d=3$, 
    $p_y=1/8,\forall y\in\{1,2,\ldots,8\}$, cell-boundary arrays $\mathbf{a}_1=\mathbf{a}_2=\mathbf{a}_3=(-1,0,1)$, and conditional probability for the cell indexes
    \begin{equation}
        p_{(1,1,1)|1} = p_{(1,1,2)|2} = p_{(1,2,1)|3} = p_{(1,2,2)|4} = p_{(2,1,1)|5} = p_{(2,1,2)|6} = p_{(2,2,1)|7} = p_{(2,2,2)|8} = 1.
    \end{equation}
    $p_{\mathbf{i}|y}=0$ for all $(\mathbf{i},y)\in\mathfrak{I}\times[M]$ not already defined. In this example
    \begin{equation}
    \begin{aligned}
    I(X_1;Y) = I(X_2;Y) = I(X_3;Y) &= 1\:\text{bit},\\
    I(X_1,X_2;Y) = I(X_1,X_3;Y) = I(X_2,X_3;Y) &= 2\:\text{bits},\\
    I(X;Y) = H(Y) &= 3\:\text{bits}.
    \end{aligned}
    \end{equation}

    \item[iii)] The model 2D-Demonstration (shown in Fig.~\ref{sfig:demo-2d}) considers $M=3$; $d=2$; 
    $p_1=0.2$, $p_2=0.5$, and $p_3=0.3$; cell-boundary arrays $\mathbf{a}_1 = (-0.5,0.5,1.5,2.0,3.5)$ and $\mathbf{a}_2 = (1.0,1.5,2.5)$; and conditional probability for the cell indexes 
    \begin{equation}
    \arraycolsep=3mm
    \label{eqn:cond-prob-demo-2d}
        \begin{array}{lllll}
        p_{(1,1)|1} = 0.4, & p_{(1,2)|1} = 0.05, & p_{(2,1)|1} = 0.3, & p_{(3,1)|1} = 0.2, \hspace{5mm} & p_{(4,1)|1} = 0.05,\\
        p_{(1,2)|2} = 0.2, & p_{(2,2)|2} = 0.3, &  p_{(3,1)|2} = 0.3, & p_{(4,1)|2} = 0.2, & \\
        p_{(3,2)|3} = 0.7, & p_{(4,1)|3} = 0.3,
        \end{array}
    \end{equation}
    $p_{\mathbf{i}|y}=0$ for all $(\mathbf{i},y)\in\mathcal{I}\times[M]$ not already defined. 
    In this example, $I(X_1;Y) \approx 0.327\:\text{bits}$, $I(X_2;Y) \approx 0.175\:\text{bits}$, $I(X;Y) \approx 1.049\:\text{bits}$, and $H(Y) \approx 1.485\:\text{bits}$.

    \item[iv)] The model 3D-Demonstration (shown in Fig.~\ref{sfig:demo-3d}) extends from model 2D-demonstration. It considers $M=3$; $d=2$; 
    $p_1=0.2$, $p_2=0.5$, and $p_3=0.3$; cell-boundary arrays $\mathbf{a}_1 = (-0.5,0.5,1.5,2.0,3.5)$, $\mathbf{a}_2 = (-1.0,0.0,0.3,1.0,3.0)$, and $\mathbf{a}_3 = (1.0,1.5,2.5)$; and conditional probability of the cell indexes given by: 
    \begin{equation}
        p_{(i,\ell,j)|y} = \left(1-1_{\{(k+1,k)\}_{k=1}^3}(\ell, y)\right)\cdot\tilde{p}_{(i,j)|y}/3,\forall(i,\ell,j,y)\in\left(\bigtimes_{k=1}^3[n_k]\right)\times[M],
    \end{equation}
    where $\tilde{p}_{(i,j)|y}$ denotes the numerical value shown in (\ref{eqn:cond-prob-demo-2d}) of the conditional probabilities $p_{(i,j)|y}$ of the 2D-Demonstration model.\footnote{For example, $p_{(2,1,1)|1} = \tilde{p}_{(2,1)|2}/3$; from (\ref{eqn:cond-prob-demo-2d}), $\tilde{p}_{(2,1)|2}=0.3$, and consequently, $p_{(2,1,1)|1}=0.1$.} This means, $p_{(i,\ell,j)|y}$ equals $\tilde{p}_{(i,j)|y}/3$ for all possible values of $(i,\ell,j,y)$, except when $(\ell,y)$ takes values in $\{(2,1),(3,2),(4,3)\}$ where $p_{(i,\ell,j)|y}$ equals $0$.
    In this example $H(Y) \approx 1.485\:\text{bits}$, $I(X;Y) \approx 1.182\:\text{bits}$, and
    \begin{equation}
    \arraycolsep=3mm
    \label{eqn:mi-demo-expressions}
        \begin{array}{rr}
            I(X_1;Y) \approx 0.327\:\text{bits,} & I(X_2;Y) \approx 0.376\:\text{bits,}\\
            I(X_3;Y) \approx 0.176\:\text{bits}, & I(X_1,X_2;Y) \approx 0.663\:\text{bits,}\\ I(X_1,X_3;Y) \approx 1.049\:\text{bits,} & I(X_2,X_3;Y) \approx 0.532\:\text{bits}.
        \end{array}
    \end{equation}
\end{itemize}

\subsection{Sampling}
To produce i.i.d. realizations of $(X,Y)\sim \mu_{X,Y}$, 
it is useful to consider an auxiliary (hidden) discrete rv. $\mathbf{I}$ in $\mathfrak{I}$ and the following three-stage process:
\begin{itemize}
    \item[i)] Sample $Y$ from its marginal pmf, i.e., $Y\sim\mu_Y$.
    \item[ii)] Given $Y=y$ from the previous step, sample $\mathbf{I}$ from the discrete conditioned pmf; i.e., sample $\mathbf{I}|Y=y\sim (p_{\mathbf{i}|y})_{\mathbf{i}\in \mathfrak{I}}$. 
    \item[iii)] Given $\mathbf{I}=\mathbf{i}$ from the previous step, sample $X$ with a uniform pdf in the indexed cell $A_{\mathbf{i}}$, i.e., $X|\mathbf{I}=\mathbf{i}\sim\text{Unif}(A_{\mathbf{i}})$.
\end{itemize}

For illustration,  Fig.~\ref{fig:model-demo} shows i.i.d. realizations of the four presented models. 

\subsection{Sparse Models}
\label{sec:non-info-components}
For our numerical analysis, we begin with the model presented in Appendix \ref{supp:models}, and we add components on $X$ that are non-informative of $Y$ (i.e., statistically independent) to produce a sparse model in the sense presented in Eq.(\ref{eq_ps_13b}).  This can be done by considering a $\nu$-dimensional rv. $V\equiv(V_{\ell})_{\ell=1}^{\nu}$ with values in $(\mathbb{R}^{\nu},\mathcal{B}(\mathbb{R}^{\nu}))$ independent of $(X,Y)$
and concatenating the components of $X$ and $V$ in a new input vector $X'$ with values in $(\mathbb{R}^{d+\nu},\mathcal{B}(\mathbb{R}^{d+\nu}))$.

\subsection{Selected Model, Masking, and IS Pre-Encoder} 
\label{supp:masked-models}
The specific models used in our experiments come from the 3D-Demonstration model presented in Appendix~\ref{sec:model-examples}, point (iv), see Fig.~\ref{sfig:demo-3d}. 
In particular, $X$ is a sparsified version (see Appendix~\ref{sec:non-info-components}) of the 3D-Demonstration model. For this, we concatenate $X^{\text{3D}}\equiv(X_1^{\text{3D}},X_2^{\text{3D}},X_3^{\text{3D}})$ with a 12-dimensional rv $V=(V_{\ell})_{\ell=1}^{12}$ (independent of $(X^{\text{3D}},Y)$) to produce  a 15-dimensional rv. $X\equiv (X_1^{\text{3D}},V_1,X_2^{\text{3D}},V_2,X_3^{\text{3D}},V_3,V_4,\ldots,V_{12})$. From this  sparse model (denoted by $\mu_{X,Y}$), we derive two other models, $\mu_{\tilde{X},Y}$ and $\mu_{\bar{X},Y}$, by masking specific coordinates of $X$ as presented in Section \ref{sec_numerical_model}. Finally, using these masking operators (as a special case of a feature selector), we compute the following MI values (following Appendix \ref{mi_computation_sim_model})
\begin{equation}
    I(X;Y) \approx 1.182\:\text{bits}, \hspace{5mm} I(\tilde{X};Y) \approx 0.532\:\text{bits}, \hspace{5mm} I(\bar{X};Y) = 0\:\text{bits}.
\end{equation}

\subsection{Neural Network Architectures} 
\label{supp:nn-mlps}
Here, we provide complementary details of the machine-learning scheme, $\Xi=\left\{ \xi_n(\cdot), n\geq 1 \right\}$ used in our study. For all the schemes, we chose $f_{\phi}(\cdot)$ to be a multilayer perceptron (MLP). Specifically, each architecture is composed of an input layer, $h_0(\cdot)$, followed by a ReLU activation, a set of $l-1\in\mathbb{N}$ hidden layers, $h_{k}(\cdot)$ interleaved with ReLU activations, and an output layer $h_l(\cdot)$, with $k\in[l-1]$.\footnote{The choice of an MLP with ReLU activations is motivated by established results in the literature that prove the functional capacity of these architectures as universal approximators with arbitrary precision and given enough width or enough depth. 
}
We use three specific architectures, denoted as $\text{MLP32}$, $\text{MLP256}$, and $\text{MLP1024}$, to study a range of complexities (functional approximation capacities). As their names imply, these three architectures adjust their hidden layers' width to 32, 256, and 1024, respectively. Moreover, the two latter architectures have two hidden layers, whereas the former has only one. Therefore, for all the models discussed in Appendix~\ref{supp:masked-models}, our MLP architectures have a 15-input dimension, an output dimension of 3, and between one or two hidden layers of different widths.  We also explore the effect of training ML schemes when applying a low-dimensional IS projection $\eta_{\mathbf{5}}(\cdot)$. This was done by training MLPs with a reduced input dimensionality matching the input after applying $\eta_{\mathbf{5}}(\cdot)$, i.e., a 5-dimensional input. This experimental design allows us to explore the capacity of MLP architectures to learn predictive models 
of varying difficulty. 

\subsection{Training Process} 
\label{supp:training}
Regarding training and optimization, our ML scheme was trained for 30 epochs using Stochastic Gradient Descent (SGD) and the standard Cross-Entropy Loss.  We specifically mention epochs instead of training steps, as the former are agnostic to the length of our training set ($S_n$). From the detail presented in the Appendixes \ref{supp:models}, \ref{supp:masked-models}, and \ref{supp:nn-mlps}, we derive a total of five specific data lengths, three architecture complexities, six models ($\mu_{X,Y}$, $\mu_{\tilde{X},Y}$, and $\mu_{\bar{X},Y}$, and its pre-encoded versions) summing a total of 90 training experiments. Each experiment was run three times using different seeds. The results shown in Figure \ref{fig2} are the average of these runs. Table \ref{tab:hyp} presents the training details for each experiment. 

\begin{table}[ht!]
    \centering
    \caption{Training hyperparameter details for each data length}
    \begin{tabular}{l|c|c}
        \textbf{Optimizer} & \textbf{Batch size} & \textbf{Training Set data length}\\\hline\hline
        \multirow{5}{9em}{\shortstack[l]{SGD\\Learning rate: $10^{-2}$}\\Momentum: $0.97$} & \multirow{1}{4em}{44} & \multirow{1}{4em}{$2.8\cdot10^3$}\\
        & \multirow{1}{4em}{344} & \multirow{1}{4em}{$2.2\cdot10^4$}\\
        & \multirow{1}{4em}{512} & \multirow{1}{4em}{$6.0\cdot10^4$}\\
        & \multirow{1}{4em}{512} & \multirow{1}{4em}{$4.6\cdot10^5$}\\
        & \multirow{1}{4em}{1024} & \multirow{1}{4em}{$1.3\cdot10^6$}\\
    \end{tabular}
    \label{tab:hyp}
\end{table} 

We explored various hyperparameters for the learning rates and batch sizes for each data length (see Table \ref{tab:hyp}). As our training set grew, we realized that maintaining the fixed learning rate had a lesser impact than modifying the batch size. Increasing batch size for the SGD allows us to explore a wide range of training scenarios while also accelerating training time.   

\bibliographystyle{IEEEtran}				
\bibliography{main_jorge_silva}				

\end{document}